\documentclass[a4paper,12pt,times,numbered,print,index]{Classes/PhDThesisPSnPDF}

\input{Preamble/preamble}

\title{Addressing Explainability of Generative AI using SMILE}

\subtitle{Statistical Model-agnostic Interpretability with Local Explanations}

\author{Zeinab Dehghani}

\dept{School of Computer Science}

\university{University of Hull}
\crest{\includegraphics[width=0.5\textwidth]{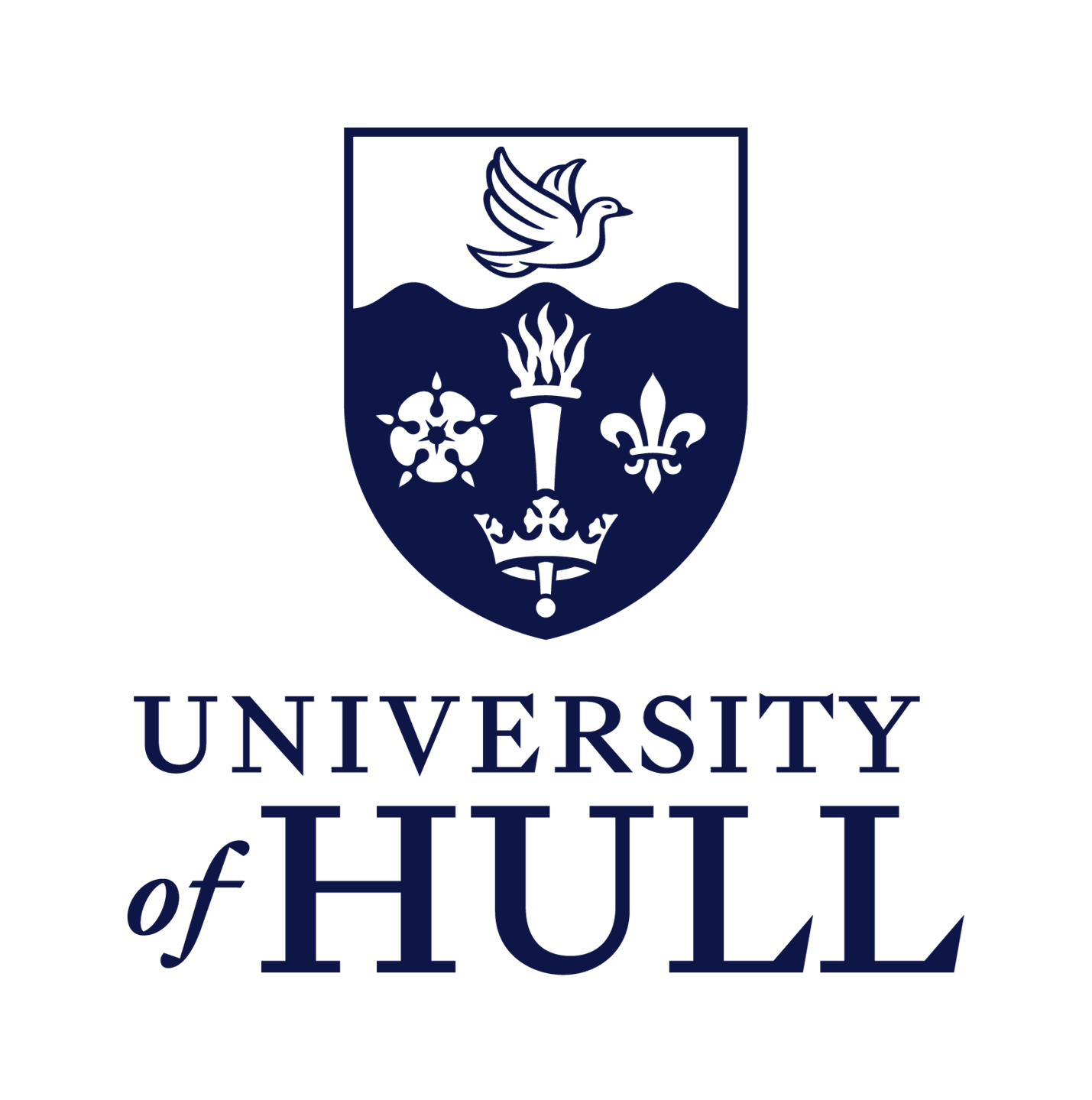}}

\degreetitle{Master of Science}

\degreedate{September 2025} 

\subject{LaTeX} \keywords{{LaTeX} {MSc Thesis} {Computer Science} {University of
Hull}}

\ifdefineAbstract
\fi

\ifdefineChapter
\fi
\usepackage{array, multirow, booktabs, longtable}
\usepackage{tikz}
\usepackage{adjustbox}
\usetikzlibrary{positioning}
\usepackage{tabularx}
\usepackage[x11names]{xcolor}  
\usepackage[linesnumbered,ruled,vlined]{algorithm2e}
\usepackage{float}

\usepackage{url}
\usepackage[english]{babel}
\usepackage{amsmath}
\microtypesetup{protrusion=true, expansion=false}

\usepackage{etoolbox}

\begin{document}

\frontmatter

\maketitle


\begin{dedication} 

\newpage

I dedicate this thesis with deepest love and gratitude to my kind and supportive husband, whose encouragement and companionship sustained me through every step of this challenging journey.

\end{dedication}


\begin{declaration}

I hereby declare that except where specific reference is made to the work of 
others, the contents of this dissertation are original and have not been 
submitted in whole or in part for consideration for any other degree or 
qualification in this, or any other university. This dissertation is my own 
work and contains nothing which is the outcome of work done in collaboration 
with others, except as specified in the text and Acknowledgements. This 
dissertation contains fewer than 21,000 words including appendices, 
bibliography, footnotes, tables and equations and has fewer than 25 figures.


\end{declaration}

\begin{acknowledgements}

I want to express my deepest gratitude to my supervisors, Dr.~Koorosh Aslansefat 
and Prof.~Adil Khan, for their invaluable guidance, encouragement, and continuous 
support throughout the course of this research. Their insightful feedback, 
patience and expertise were instrumental in shaping the direction and quality 
of this work.

I am also sincerely thankful to the academic staff and colleagues in the 
School of Computer Science at the University of Hull for their assistance, 
constructive discussions, and for creating a stimulating research environment.  

My heartfelt appreciation goes to my family for their endless love, patience, 
and moral support during this journey. In particular, I wish to thank my 
husband for his constant encouragement, understanding, and motivation, which 
have been my most significant source of strength throughout this thesis.  

Finally, I would like to acknowledge all the researchers and developers whose 
The work laid the foundation for this study and inspired many of the ideas explored 
in this dissertation.

\end{acknowledgements}


\begin{abstract}

The rapid advancement of generative artificial intelligence has enabled models capable of producing complex textual and visual outputs; however, their decision-making processes remain largely opaque, limiting trust and accountability in high-stakes applications. This thesis introduces gSMILE, a unified framework for the explainability of generative models, extending the Statistical Model-agnostic Interpretability with Local Explanations (SMILE) method to generative settings.
gSMILE employs controlled perturbations of textual input, Wasserstein distance metrics, and weighted surrogate modelling to quantify and visualise how specific components of a prompt or instruction influence model outputs. Applied to Large Language Models (LLMs), gSMILE provides fine-grained token-level attribution and generates intuitive heatmaps that highlight influential tokens and reasoning pathways. In instruction-based image editing models, the exact text-perturbation mechanism is employed, allowing for the analysis of how modifications to an editing instruction impact the resulting image. Combined with a scenario-based evaluation strategy grounded in the Operational Design Domain (ODD) framework, gSMILE allows systematic assessment of model behaviour across diverse semantic and environmental conditions.
To evaluate explanation quality, we define rigorous attribution metrics, including stability, fidelity, accuracy, consistency, and faithfulness, and apply them across multiple generative architectures. Extensive experiments demonstrate that gSMILE produces robust, human-aligned attributions and generalises effectively across state-of-the-art generative models. These findings highlight the potential of gSMILE to advance transparent, reliable, and responsible deployment of generative AI technologies.
\end{abstract}


\tableofcontents

\listoffigures

\listoftables



\mainmatter


\chapter*{Publications and Availability of Results}
\addcontentsline{toc}{chapter}{Publications and Availability of Results}

\begin{table}[H]
\centering
\renewcommand{\arraystretch}{1.3}
\begin{tabularx}{\textwidth}{
    p{0.03\linewidth} 
    >{\raggedright\arraybackslash}X 
    >{\raggedright\arraybackslash}p{0.35\linewidth} 
    p{0.18\linewidth}
}
\hline
\textbf{ID} & \textbf{Reference} & \textbf{Chapter \& Links}  \\
\hline

1 &
\textbf{Dehghani, Z.}, \& Aslansefat, K. (2024). 
\textit{Explaining Large Language Models with gSMILE}. 
Submitted to \textbf{IEEE Transactions on Artificial Intelligence}. &
\parbox[t]{\linewidth}{
Chapter 3\\
\href{https://arxiv.org/abs/2505.21657}{Paper Link}\\
\href{https://www.youtube.com/watch?v=0QTf2qasrgM&t=812s}{Video Link}
}  \\
\hline

2 &
\textbf{Dehghani, Z.}, Aslansefat, K., Mehmood Khan, A., Ramírez Rivera, A., George, F., \& Khalid, M. (2024). 
\textit{Mapping the Mind of an Instruction-based Image Editing using SMILE}. 
Submitted to \textbf{npj Nature Artificial Intelligence}. &
\parbox[t]{\linewidth}{
Chapter 4\\
\href{https://sciety.org/articles/activity/10.21203/rs.3.rs-5943708/v1}{Paper Link}\\
\href{https://www.youtube.com/watch?v=pJePjOb2Tj4}{Video Link}
} \\
\hline

\hline
3 &
Z. Zehtabi Sabeti Moghaddam, \textbf{Dehghani, Z.}, Rani, M., Aslansefat, K., 
Mishra, B., Kureshi, R., \& Thakker, D. (2025).  
\textit{Explainable Knowledge Graph Retrieval-Augmented Generation (KG-RAG) with KG-SMILE}.  
Submitted to \textbf{Springer}. &
\parbox[t]{\linewidth}{
\textbf{Other Collaboration}\\
\href{https://arxiv.org/abs/2509.03626}{Paper Link}
} \\
\hline

\end{tabularx}
\caption{Publications and associated resources resulting from this research.}
\label{tab:publications_availability}
\end{table}

\chapter*{List of Symbols}
\addcontentsline{toc}{chapter}{List of Symbols}

The following symbols and notations are used throughout this thesis:

\begin{table}[H]
\centering
\renewcommand{\arraystretch}{1.2}
\begin{tabular}{p{0.16\linewidth} p{0.70\linewidth}}
\hline
\textbf{Symbol} & \textbf{Description} \\
\hline
$ x $ & Original input prompt. \\
$ \hat{x}_j $ & $j$-th perturbed prompt generated from $x$. \\
$ \{ \hat{x}_j \}_{j=1}^{J} $ & Set of all $J$ perturbed prompts derived from $x$. \\
$ \pi^{(n)}(y \mid x) $ & Output distribution of the black-box model for input $x$. \\
$ \pi^{(n)}(y \mid \hat{x}_j) $ & Output distribution of the black-box model for perturbed input $\hat{x}_j$. \\
$ \delta_{x_j} $ & Input-level semantic distance (IWMD) between $x$ and $\hat{x}_j$. \\
$ \Delta(x, \hat{x}_j) $ & Output-level semantic shift (OWMD or Wasserstein distance) between outputs for $x$ and $\hat{x}_j$. \\
$ W(p, q) $ & Wasserstein distance between two distributions $p$ and $q$. \\
$ w_j $ & Gaussian kernel weight for perturbation $\hat{x}_j$, based on $\delta_{x_j}$. \\
$ \sigma $ & Kernel width parameter controlling weight decay in the Gaussian kernel. \\
$ z_j $ & Feature-vector representation of perturbation $\hat{x}_j$ (e.g., one-hot encoding). \\
$ h_{\theta}(z_j) $ & Surrogate model prediction for perturbation $\hat{x}_j$. \\
$ \theta, \theta_0 $ & Coefficients and bias term of the surrogate linear regression model. \\
$ n $ & Dimensionality of the embedding or feature space. \\
$ p $ & Norm order used in the Wasserstein-$p$ metric. \\
\hline
\end{tabular}
\caption{List of symbols and notations used throughout the thesis.}
\label{tab:symbols}
\end{table}

\chapter{Introduction}  
\label{chap:Introduction}

\ifpdf
    \graphicspath{{Chapter1/Figs/Raster/}{Chapter1/Figs/PDF/}{Chapter1/Figs/}}
\else
    \graphicspath{{Chapter1/Figs/Vector/}{Chapter1/Figs/}}
\fi


Artificial intelligence (AI) has made tremendous contributions recently, especially in deep learning (DL) and transformer-based models. Large language models (LLMs) and Instruction-guided image editing models represent two prominent yet distinct strands of generative AI. LLMs, built on transformer architectures ~\cite{vaswani2017attention, brown2020language}, have revolutionised natural language processing by generating fluent responses, following complex
instructions and supporting human-AI dialogue in various domains~\cite{achiam2023gpt}. In parallel, diffusion-based and transformer-based image editing models have enabled users to issue high-level textual commands such as ``make the weather look stormy'' or ``remove the person in the background'' and receive corresponding visual transformations~\cite{shuai2024survey, li2023blip}.

The fact that these models are so difficult to interpret is a significant concern, especially in high-stakes areas such as healthcare, self-driving cars, and forensic image analysis, where trust and reliability are crucial~\cite{samek2017explainable, rudin2019stop}. To address this issue, the field of Explainable Artificial
Intelligence (XAI) has emerged to make AI models more transparent and easier to understand~\cite{doshi2017towards, adadi2018peeking}. Although traditional XAI methods have been widely researched in tasks such as image classification and object detection~\cite{ribeiro2016should, selvaraju2017grad}, their use in multimodal generative models, such as LLMs and diffusion-based image editing, has not been explored as much~\cite{bommasani2021opportunities, achiam2023gpt, li2023blip}.

This thesis presents gSMILE (generative Statistical Model-agnostic Interpretability with Local Explanations), a new explainability method designed to make multimodal generative models more transparent. gSMILE is a model-agnostic approach, meaning it can be applied without access to internal model components such as gradients or model architecture, making it suitable for black-box and API-based systems. It builds upon the original SMILE framework~\cite{aslansefat2023explaining} and extends it to generative settings by incorporating statistical distance measures to provide more stable, consistent, and reliable interpretations. This research aims to bridge the gap between generative AI and explainability, enabling a deeper understanding of how specific input elements influence the outputs of text-to-text generation and instruction-based image editing models~\cite{brown2020language, shuai2024survey}.

\section{Generative AI models}

Generative AI is one of the fastest-evolving areas within artificial intelligence, focusing on producing entirely new content, such as text, images, audio, and video, by learning patterns from large datasets~\cite{goodfellow2014generative}. Unlike traditional AI systems that analyse or classify existing data, generative models can create outputs that closely resemble human-made work, whether writing a story, generating realistic visuals, or composing music~\cite{bommasani2021opportunities}.

Some of the most notable advancements in this field have come from large language models (LLMs) such as \textbf{OpenAI's ChatGPT}, \textbf{Google's Gemini}, \textbf{Anthropic's Claude}, \textbf{Meta's LLaMA}, and \textbf{DeepMind's Gemma}~\cite{achiam2023gpt, team2024gemini, claude2024, touvron2023llama}. These models have rapidly progressed beyond simple text generation. Today, tools like \textbf{ChatGPT-4o} and \textbf{Gemini 1.5} not only generate high-quality images from text prompts but also support image editing through conversational interfaces~\cite{hurst2024gpt, gemini2024image}.
This shift toward multimodal capabilities reflects a broader trend toward integrating vision and language in a seamless user experience.

Two of the most impactful applications of generative AI are \textbf{text-to-text generation} and \textbf{instruction-based image editing}. Text-to-text generation, powered by large language models (LLMs), has revolutionised natural language processing by enabling fluent text generation, summarisation, translation, and complex reasoning~\cite{brown2020language, achiam2023gpt}. Models like GPT-3~\cite{brown2020language} and GPT-4~\cite{achiam2023gpt} have set new benchmarks in this domain. Meanwhile, instruction-based image editing allows users to modify images through natural language commands, such as ``make the sky look stormy'' or ``remove the person in the background,'' leveraging diffusion-based and transformer-based models~\cite{shuai2024survey, li2023blip}. Recent advances have integrated these capabilities into versatile multimodal systems, enabling seamless interaction between text and images.

Instruction-based image editing enables users to modify visual content through natural language commands, for example, ``remove the person in the background'' or ``change the sky to a stormy scene'', without the need for manual tools. Diffusion-based models such as InstructPix2Pix~\cite{brooks2023instructpix2pix} and Imagen Editor~\cite{wang2023imagen}, along with the latest Gemini and ChatGPT updates, now support both the generation and direct modification of visual content via prompts~\cite{gemini2024image, hurst2024gpt}. These capabilities make such models attractive for creative, accessibility, and productivity-focused applications.

These tools have seen unprecedented adoption, reflecting their growing influence across both consumer and enterprise applications. For instance, \textbf{ChatGPT reached 100 million users in just two months} of its launch, making it the fastest-growing consumer application in history~\cite{bick2024rapid}. Other models, such as Gemini and Claude, are being integrated into search engines, office tools, mobile devices, and browsers, making generative AI more accessible than ever~\cite{mckinsey2023charts}.

\begin{figure}[H]
    \centering
    \includegraphics[width=0.95\textwidth]{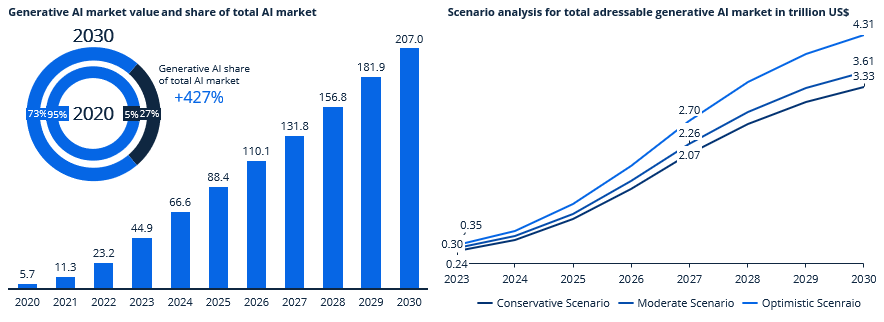}
    \caption{Projected growth of the generative AI market and its share of the total AI market (2020–2030). The left side shows year-by-year expansion, while the right panel presents scenario-based global projections reaching up to \$4.31 trillion by 2030. Source: Master of Code~\cite{bick2024rapid}.}
    \label{fig:genai-market-growth}
\end{figure}

Alongside this rapid adoption, the economic potential of generative AI has grown exponentially. Forecasts predict that the \textbf{generative AI market will exceed \$200 billion by 2030}, representing a 427\% increase in market share compared to 2020. Fig.~\ref{fig:genai-market-growth} illustrates this trend, highlighting both annual market expansion and scenario-based projections for total global value. As shown in Fig.~\ref{fig:genai-market-growth}, scenario-based estimates suggest that the worldwide value of generative AI could reach \$4.31 trillion by 2030~\cite{bick2024rapid}.

What sets generative models apart from traditional AI systems is their open nature. Unlike classification tasks that produce clear, bounded outputs (e.g. cat vs. dog), generative models generate creative, multimodal outputs such as complete sentences or altered images often influenced by abstract patterns and latent representations~\cite{bommasani2021opportunities, goodfellow2014generative}. This complexity makes it especially difficult to understand how a specific decision was made or why a particular output was generated over another~\cite{doshi2017towards}.

Despite these advancements, generative models remain largely opaque. Built on deep learning and transformer-based architectures, they often function as ``black boxes''~\cite{rudin2019stop, vaswani2017attention}, making their internal decision-making processes difficult to interpret or verify. This lack of transparency becomes particularly concerning in high-stakes environments like healthcare, law, and education, where understanding how a decision is as important as the decision itself~\cite{samek2017explainable, adadi2018peeking}.

\section{Understanding Explainability in AI} 

Explainability in artificial intelligence (AI) refers to the degree to which a human can understand the reasoning behind a model's decision or prediction~\cite{guidotti2018survey}. Unlike traditional ``black-box'' models, such as deep neural networks, which often lack interpretability, explainable models aim to make their internal logic more transparent. It is essential in critical domains where users must be able to trust, validate, and even contest AI-generated outcomes~\cite{rudin2019stop}. Explainability fosters trust in AI systems and supports accountability, fairness, and regulatory compliance~\cite{doshi2017towards, adadi2018peeking}.

In recent years, explainable artificial intelligence (XAI) has evolved from a research interest into an operational necessity across multiple sectors. As AI systems become increasingly embedded in high-stakes environments, such as healthcare, finance, national security, and industrial automation, the need for transparency and interpretability has increased significantly~\cite{adadi2018peeking,doshi2017towards, samek2017explainable}. Stakeholders are now demanding that AI not only perform well but also justify its decisions in a manner that is understandable to humans. This demand has sparked a surge in investment and development in the XAI market, particularly in the United States~\cite{mckinsey2023charts}. Analysts project substantial growth in the adoption of explainable AI tools in areas like fraud detection, drug diagnostics, and predictive maintenance, highlighting the market's response to regulatory pressures and the need for trust in AI-driven systems.

\begin{figure}[H]
    \centering
    \includegraphics[width=0.9\textwidth]{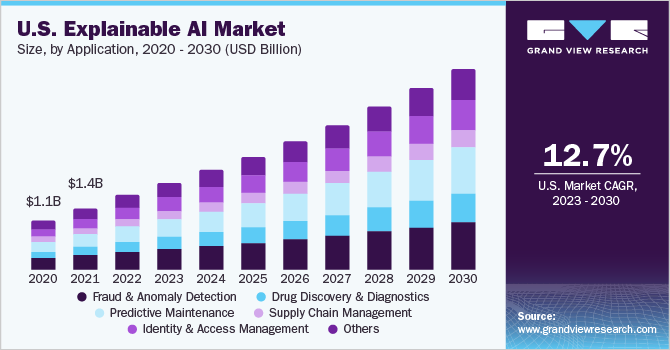}
    \caption{Projected growth of the U.S. Explainable AI market by application\\ (2020–2030)~\cite{grandview2023xai}.}
    \label{fig:xai-market-}
\end{figure}

As shown in Fig.~\ref{fig:xai-market-}, the U.S. explainable AI market is expected to experience significant growth across multiple application areas through 2030, reflecting the critical role of explainability in enabling responsible AI deployment across industries~\cite{grandview2023xai}.

\section{Research Questions and Contributions}

This research addresses key challenges in the interpretability of generative AI systems, specifically large language models (LLMs) and instruction-based image editing models. Despite recent advancements, understanding how these models process and respond to input data remains a significant obstacle to building trustworthy and transparent AI solutions. To address these challenges, this work is guided by the following research questions:

\begin{itemize}

    \item \label{rq1} \textbf{RQ1: What evaluation metrics can comprehensively assess interpretability methods' robustness, fidelity, and usability across different AI domains?}  
    To ensure the practical relevance of interpretability methods, rigorous evaluation is essential. This question aims to identify and apply a set of quantitative metrics, including stability, fidelity, accuracy, and consistency, to evaluate and benchmark the performance of the proposed methods in both language and image domains (\textbf{Chapter~\ref{chap:gSMILE Methodology}}).
    
\item \label{rq2} \textbf{RQ2: How can interpretability methods reveal the influence of individual input elements in large language models (LLMs)?}  
Although LLMs have demonstrated remarkable capabilities across various tasks, they often operate as black boxes. This question examines how interpretability techniques, particularly \textbf{gSMILE}, can trace and visualise the contributions of specific words or tokens to the model’s output, thereby offering more profound insight into the model’s internal reasoning processes (\textbf{Chapter~\ref{chap:llm_explainability}}).

    \item \label{rq3} \textbf{RQ3: How can these interpretability methods be adapted to multimodal systems, especially instruction-based image editing models?}  
    Instruction-based image editing presents unique challenges due to the complexity of translating textual commands into visual modifications. This question examines how interpretability methods can be applied to map textual inputs to specific changes in generated images, thereby enabling transparency and user trust in these systems (\textbf{Chapter~\ref{chap:image_editing_explainability}}).

\end{itemize}

\section{Research Hypothesis and Objectives}

Building upon the insights gained from the experimental validation of SMILE, this section outlines the core hypothesis driving the research and the primary objectives established to test and support this hypothesis.

The central hypothesis of this research is that \textbf{gSMILE} (generative Statistical Model-agnostic Interpretability with Local Explanations) can provide more stable, consistent, and robust explanations for complex AI models, particularly in multimodal generative tasks such as text-to-text generation and instruction-based image editing, compared to existing post-hoc explainability methods. It is further hypothesised that gSMILE’s use of statistical distance measures and distributional comparisons enhances its resilience to adversarial manipulations and improves alignment with human intuition.

The research aims to evaluate gSMILE’s performance in generating accurate and consistent explanations across different model types, including black-box and multimodal systems. The evaluation will focus on key criteria, including explanation accuracy (measured using metrics such as attention-based AUC), fidelity, consistency, and stability. The study will also compare the effects of different perturbation strategies on explanation quality and robustness, providing insights into how perturbation choices influence interpretability outcomes. Additionally, gSMILE’s resilience to adversarial attacks will be assessed by determining its ability to maintain explanation reliability and expose biased decision-making under adversarial conditions. The research will further analyse human alignment by comparing gSMILE’s explanations with expert-identified features and human intuition. Finally, the study will demonstrate gSMILE’s scalability and applicability in real-world, high-risk domains, including healthcare, autonomous driving, and forensic analysis.

\section{Thesis Structure}  

This thesis is organised into six chapters as follows:

\begin{itemize}
    \item \textbf{Chapter~\ref{chap:Introduction}: Introduction.}  
    This chapter introduces the background and motivation of the research, defines the research questions and objectives, outlines the hypothesis, and provides an overview of the overall structure of the thesis.

    \item \textbf{Chapter~\ref{chap:Background and Literature Review}: Background and Literature Review.}  
    This chapter provides a comprehensive review of the existing literature on generative AI, encompassing large language models (LLMs), instruction-based image editing, and explainable artificial intelligence (XAI). It also identifies the key challenges and limitations of current explainability techniques that motivate the proposed framework.

    \item \textbf{Chapter~\ref{chap:gSMILE Methodology}: gSMILE Methodology.}  
    This chapter describes the theoretical foundations and methodology of the proposed \textbf{gSMILE} framework, explaining how it extends LIME and SMILE through statistical distance measures, Gaussian weighting, and surrogate modelling. It also introduces the attribution evaluation metrics accuracy, fidelity, stability, and consistency used throughout the thesis.

    \item \textbf{Chapter~\ref{chap:llm_explainability}: Explainability of Large Language Models.}  
    This chapter applies the proposed \textbf{gSMILE} framework to large language models (LLMs) to address Research Question~RQ2. It analyses token-level attribution, evaluates interpretability using quantitative metrics, and visualises model reasoning paths.

    \item \textbf{Chapter~\ref{chap:image_editing_explainability}: Explainability of Instruction-based Image Editing.}  
    This chapter applies the \textbf{gSMILE} framework to multimodal generative models, with a focus on instruction-based image editing. It addresses Research Question~RQ3 by evaluating how textual instructions influence visual outputs, using both qualitative heatmaps and quantitative assessments.

    \item \textbf{Chapter~\ref{chap:Conclusion}: Conclusion and Future Works.}  
    This chapter summarises the main findings and contributions of the research, discusses identified limitations, and outlines potential directions for future work to enhance explainability in generative AI.
\end{itemize}

\label{section1.3}


\chapter{Background and Literature Review}
\label{chap:Background and Literature Review}

This chapter provides an overview of recent developments in generative artificial intelligence, including large language models, instruction-based image editing, and explainability techniques. The reviews explainability approaches applied specifically to large language and image editing models.

\section{Types of Generative Artificial Intelligence}

\begin{figure}[H]
    \centering
    \includegraphics[width=\textwidth]{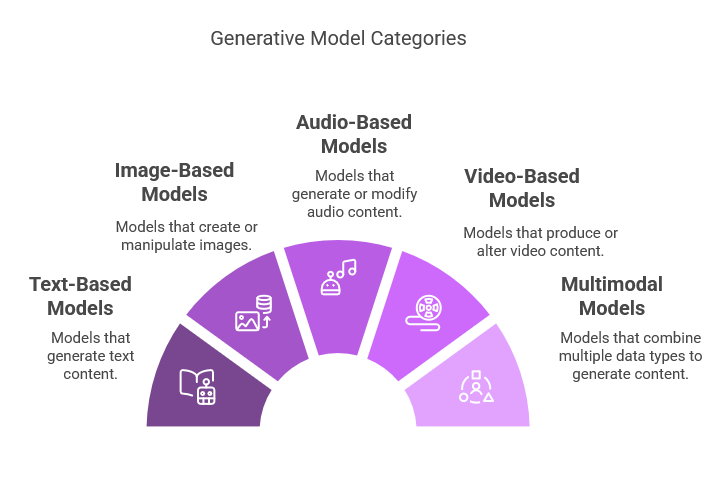}
    \caption{Types of generative AI, including text-based, image-based, audio-based, video-based, and multimodal systems.}
    \label{fig:types_generative_ai}
\end{figure}

Generative artificial intelligence (AI) models can be broadly categorised based on the modality of content they produce. As illustrated in Fig.~\ref{fig:types_generative_ai}, these categories include text-based, image-based, audio-based, video-based, and mixed or multimodal systems. Each category leverages different model architectures and techniques suited to the specific characteristics of the data modality.

Text-based generative models, such as large language models (LLMs), generate human-like text and have demonstrated capabilities in tasks including content creation, summarisation, translation, and dialogue~\cite{brown2020language, achiam2023gpt}. Image-based models, including those based on diffusion models and generative adversarial networks (GANs), can create and manipulate images, often guided by textual instructions~\cite{rombach2022high, li2023blip}. Audio-based models can generate speech, music, or other audio signals, with applications ranging from voice synthesis to music composition~\cite{van2016wavenet}. Video-based generative models extend these capabilities to dynamic visual content, enabling video generation or editing~\cite{villegas2017learning}. Finally, mixed or multimodal systems combine multiple data types, allowing for complex tasks such as generating images from text prompts or providing visual explanations for textual queries~\cite{tsimpoukelli2021multimodal, alayrac2022flamingo}.

\subsection{Large Language Models (LLMs)}

Large language models (LLMs) have become increasingly prevalent due to significant advancements in deep learning, improved access to data, and more powerful computing resources. These models are trained on vast amounts of text and can generate responses that sound natural, answer questions, and support a wide range of language-based tasks. Over time, several notable LLMs have helped push the field forward, each bringing new ideas and improvements.

GPT (OpenAI)~\cite{brown2020language} is one of the most well-known autoregressive transformer-based models, capable of generating coherent and contextually rich text. Google's T5~\cite{raffel2020exploring} introduced the text-to-text framework, where all NLP tasks are framed as text generation problems, improving transfer learning and efficiency. Similarly, PaLM (Google)~\cite{chowdhery2023palm} leverages the Pathways architecture for large-scale language modelling, optimising multitask learning across different NLP tasks.

Anthropic's Claude~\cite{bai2022training} focuses on ethical AI and safe interactions designed to minimise harmful outputs while maintaining interpretability. Google's Gemma 2~\cite{team2024gemma} is optimised for logical reasoning and structured data interpretation, making it a strong candidate for academic and enterprise use cases. Meta's LLaMA~\cite{touvron2023llama} is an efficient, open-source model designed for various NLP applications, providing researchers with an accessible yet powerful alternative to proprietary models. 

EleutherAI's GPT-NeoX~\cite{black2022gpt} is an open-source alternative to GPT, providing robust text generation capabilities while allowing customisation. AI21 Labs' Jurassic-2~\cite{lieber2021jurassic} is designed for large-scale enterprise applications, emphasising contextual understanding and user intent. Cohere's Command R+~\cite{gao2023retrieval} integrates retrieval-augmented generation, improving fact-based responses and minimising hallucinations. 

Furthermore, models like BERT (Google)~\cite{devlin2019bert} have NLP through bidirectional context understanding, significantly improving sentiment analysis and question-answering tasks. Microsoft's Phi-4 focuses on efficiency and cost-effectiveness, delivering competitive performance with fewer parameters. Mistral and Alibaba's Qwen 2.5 push the boundaries of model adaptability and fine-tuning capabilities, making them suitable for multilingual and domain-specific applications.  

DeepSeek-LLM~\cite{bi2024deepseek} is another promising open-source model, competing with top-tier models in terms of benchmark performance. Models such as Mixtral and StableLM have been developed with efficiency and specific NLP task performance in mind, ensuring a balance between accuracy and computational demands. 

Recent developments have introduced models such as DeepSeek-R1, which focused on advanced reasoning capabilities, and Qwen2.5-Max, which explores large-scale Mixture-of-Experts (MoE) models to enhance efficiency and performance. OpenAI's o3-mini pushes the frontier of cost-effective reasoning, balancing performance with computational efficiency. DeepSeek-V3, the first open-sourced GPT-4o-level model, represents a significant step forward in open-source LLM development.

The following models in Fig.~\ref{fig:llm-full-table} are explicitly designed for text-to-text tasks, meaning they take text input and generate text output across various NLP applications:

\begin{figure}[H]
\centering
\resizebox{\textwidth}{!}{%
\begin{tikzpicture}[
    font=\sffamily\scriptsize,
    box/.style={minimum height=1.3cm, align=center, rounded corners, draw=black, fill=white, text=black},
    cat/.style={box, text width=3.8cm},
    model/.style={box, text width=6.2cm},
    desc/.style={box, text width=4.5cm},
    header/.style={box, fill=blue!40, font=\bfseries}
]

\node[header, text width=3.8cm] (h1) at (0,0) {Category};
\node[header, text width=6.2cm] (h2) at (5.45,0) {Models};
\node[header, text width=4.5cm] (h3) at (11.2,0) {Description};

\foreach \y/\cat/\models/\desc in {
  -1.5/{General-Purpose}/{GPT-4~\cite{achiam2023gpt}, GPT-4o~\cite{hurst2024gpt},\\ Claude 3~\cite{anthropic2024commandrplus}, Gemini 1.5~\cite{team2024gemini},\\ Command R+~\cite{gao2023retrieval}, Mistral~\cite{jiang2023mistral7b}}/{Versatile for wide\\ range of NLP tasks},
  -3.0/{Open-Source}/{LLaMA 3~\cite{grattafiori2024llama}, Mistral~\cite{jiang2023mistral7b},\\ Mixtral~\cite{jiang2024mixtral}, Falcon 180B~\cite{almazrouei2023falcon},\\ DeepSeek-V3~\cite{liu2024deepseek}, OpenChat~\cite{wang2023openchat}}/{Customizable models for\\ research applications},
  -4.5/{Retrieval-Augmented}/{Command R+~\cite{gao2023retrieval},\\ GPT-4 Bing~\cite{achiam2023gpt}, Claude 3 RAG~\cite{anthropic2024claude3},\\ Gemini Search~\cite{team2024gemini}}/{Enhance information\\ retrieval accuracy,\\ reduce misinformation},
  -6.0/{Long-Text Processing}/{Claude 3~\cite{anthropic2024claude3}, Gemini 1.5~\cite{team2024gemini},\\ GPT-4o~\cite{hurst2024gpt}, Grok-1.5~\cite{xai2024grok}}/{Specialized in\\ processing long-form text},
  -7.5/{Advanced Reasoning}/{GPT-4~\cite{achiam2023gpt}, Claude 3 Opus~\cite{anthropic2024claude3},\\ Gemini 1.5~\cite{team2024gemini}, Mixtral~\cite{jiang2024mixtral},\\ Grok-1.5~\cite{xai2024grok}}/{Logical reasoning,\\ multi-task learning,\\ problem-solving},
  -9.0/{Ethically-Aligned}/{Claude 3~\cite{anthropic2024claude3}, Gemini~\cite{team2024gemini},\\ GPT-4~\cite{achiam2023gpt}, Grok~\cite{xai2024grok}}/{Emphasize safety,\\ reducing bias,\\ ethical AI},
  -10.5/{Mixture-of-Experts (MoE)}/{Mixtral~\cite{jiang2024mixtral}, Qwen2.5-Max~\cite{hui2024qwen2},\\ GShard~\cite{lepikhin2020scaling}, Switch Transformer~\cite{fedus2022switch}}/{Improved efficiency\\ and performance via MoE},
  -12.0/{Multimodal}/{GPT-4o~\cite{hurst2024gpt}, Gemini 1.5~\cite{team2024gemini},\\ Claude 3~\cite{anthropic2024claude3}, Kosmos-2~\cite{peng2023kosmos},\\ Grok-1.5V~\cite{xai2024grok}}/{Capable of processing\\ both text and visual inputs},
  -13.5/{Multilingual}/{BLOOM~\cite{workshop2022bloom}, XGLM~\cite{chi2021xlm},\\ mBERT~\cite{devlin2019bert}, XLM-R~\cite{conneau2019unsupervised},\\ ByT5~\cite{xue2022byt5}}/{Optimized for cross-lingual\\ understanding and translation},
  -15.0/{Domain-Specific}/{Med-PaLM 2~\cite{singhal2025toward}, Galactica~\cite{taylor2022galactica},\\ Legal-BERT~\cite{chalkidis2020legal}, StarCoder2~\cite{lozhkov2024starcoder},\\ Code LLaMA~\cite{roziere2023code}}/{Tailored for specific tasks\\ such as medical, legal, or code},
  -16.5/{Instruction-Tuned}/{Alpaca~\cite{taori2023alpaca}, Vicuna~\cite{chiang2023vicuna},\\ ChatGLM~\cite{du2021glm}, Baize~\cite{xu2023baize},\\ OpenChat~\cite{wang2023openchat}}/{Tuned for better instruction-following\\ and chat capabilities},
  -18.0/{Lightweight / Edge}/{Phi-2~\cite{li2023textbooks}, TinyLLaMA~\cite{zhang2024tinyllama},\\ DistilGPT2~\cite{sanh2019distilbert}, GPT-2~\cite{radford2019gpt2},\\ LLaMA 2 7B~\cite{touvron2023llama}}/{Designed for fast inference\\ and deployment on edge devices}
}{
  \node[cat]   at (0,\y) {\cat};
  \node[model] at (5.45,\y) {\models};
  \node[desc]  at (11.2,\y) {\desc};
}

\end{tikzpicture}%
}
\caption{Extended LLM Categories, Models, and Descriptions}
\label{fig:llm-full-table}
\end{figure}

\subsection{Instruction-based Image Editing}

Instruction-based image editing modifies an input image according to specific user instructions~\cite{huang2024smartedit}. This task involves providing an instruction that guides the transformation of the original image into a new design that aligns with the given directive~\cite{brooks2023instructpix2pix}. Text-guided image editing enhances the controllability and accessibility of visual manipulation by allowing users to make changes through natural language commands~\cite{fu2023guiding}. Advanced models use large-scale training to perform precise edits efficiently~\cite{kawar2023imagic, huang2024smartedit}. 

These models have revolutionised the creative industry, providing intuitive tools for designers, artists, and content creators to produce high-quality visual content with minimal effort. Moreover, fine-tuning edits through iterative instructions allows for greater flexibility and user satisfaction, enabling personalised outcomes catering to diverse creative needs.

For example, Instruct-Pix2Pix~\cite{brooks2023instructpix2pix} uses GANs and diffusion models to make detailed edits based on user input, making it a versatile tool for flexible and specific changes. Similarly, Adobe Firefly~\cite{adobe_firefly} expands this capability by working with images, videos, and text to create visually enhanced content, often targeting professional use cases. On the other hand, tools like MagicBrush~\cite{zhang2023magicbrush} focus on applying artistic styles and creative effects to images, guided by simple text instructions. Recent studies have also explored the integration of multimodal inputs, such as combining text with sketches or reference images, to achieve more precise and context-aware modifications.

For example, SmartEdit~\cite{huang2024smartedit} utilises bidirectional interaction to manage complex edits, while MGIE~\cite{shuai2024survey} is specifically designed for large-scale, high-resolution input. Img2Img-Turbo~\cite{parmar2024one} addresses significant challenges traditional diffusion-based models face, such as slow inference times and dependence on paired datasets. It employs a single-step diffusion process and a streamlined generator network, resulting in faster inference, reduced over-fitting, and improved preservation of the structures in the input images. The model excels in unpaired tasks, such as scene translation (e.g., day-to-night and weather effects), and paired tasks, including Sketch2Photo and Edge2Image. Its efficiency and versatility make it a competitive and reliable framework for diverse GAN-based applications. Diffusers-Inpaint~\cite{rombach2022high} specialises in precise inpainting, maintaining seamlessness in modified regions.

Frameworks like MasaCtrl~\cite{cao2023masactrl} and Imagic~\cite{kawar2023imagic} ensure high-quality, realistic results. MasaCtrl excels in non-rigid transformations, while Imagic enables detailed, text-driven edits. MOECONTROLLER~\cite{li2023moecontroller} integrates global and local edits, and Learnable Regions~\cite{lin2024text} uses bounding boxes for localised adjustments without manual masking. DALL-E~\cite{ramesh2022hierarchical} and OmniGen~\cite{xiao2025omnigen} focus on dynamic, layout-preserving transformations and global/local changes, respectively.

Recent advancements improve efficiency and usability. Innovations reduce the computational overhead of diffusion models, improving practical applications~\cite{ulhaq2022efficient}. FoI (Focus on Your Instruction) isolates relevant regions for precise multi-instruction editing~\cite{guo2024focus}. At the same time, Human-Centred Generative AI (HGAI) aligns generative models with human intent and ethical standards~\cite{chen2023next}, promoting accessible and creative solutions. Instruction-based image editing will likely focus on making models more adaptable and better at handling unclear or vague user instructions. As these technologies become more common in different industries, ethical concerns such as reducing bias and ensuring the authenticity of edited content will also become increasingly important.

\section{Explainable Artificial Intelligence}

As artificial intelligence (AI) systems grow increasingly complex, interpretability becomes essential, especially in high-stakes domains such as healthcare, finance, and autonomous driving, where understanding and justifying decisions is critical~\cite{samek2017explainable, doshi2017towards, rudin2019stop}. Explainable Artificial Intelligence (XAI) aims to address this need by making the processes and outputs of AI models understandable and meaningful to humans.

Broadly, XAI methods can be categorised into two main categories~\cite{guidotti2018survey}. Intrinsic methods use inherently interpretable models, such as decision trees or linear models. After training, Post hoc methods analyse complex ``black-box'' models, providing explanations without altering the original model's structure.

Explainability answers the fundamental question: Why did the model make this decision? It provides insights into model behaviour, enhances user trust, supports regulatory compliance, and facilitates accountability. In traditional machine learning tasks, such as classification or regression, explainability often involves identifying which features most significantly influence a prediction. However, in generative AI, where outputs may be complete sentences, complex images, or even edited media, the reasoning behind decisions becomes much more challenging to trace~\cite{bommasani2021opportunities, samek2017explainable}.

For instance, it may be unclear why a model chooses certain words or sentence structures in natural language generation. In computer vision, an image captioning model might describe a person as ``smiling'' when the facial expression is ambiguous~\cite{xu2015show}. Similarly, in instruction-based image editing, a model might change the background colour instead of adding a new object~\cite{park2019semantic}.

As generative models are increasingly deployed in sensitive and creative domains, explainability becomes more than a desirable feature; it is necessary for transparency, safety, and trust. Understanding why an AI model generates a specific caption, translation, or edited image reveals how particular input elements are emphasised or transformed. It also helps identify the key features influencing decisions, detect biases or inconsistencies, and optimise performance~\cite{ribeiro2016should, selvaraju2017grad}.

\begin{figure}[H]
    \centering
    \includegraphics[width=\textwidth]{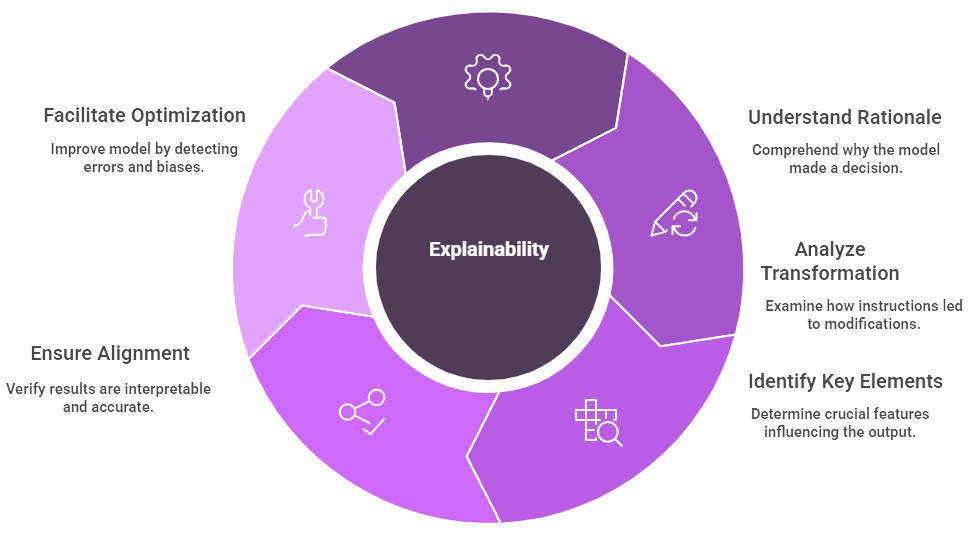}
    \caption{Key functions of explainability in AI: understanding rationale, analysing transformations, identifying key elements, facilitating optimisation, and ensuring alignment.}
    \label{fig:xai-imp}
\end{figure}

Fig.~\ref{fig:xai-imp} summarises the key purposes of explainability in AI systems: understanding the model's rationale, analysing transformations, identifying key elements, facilitating optimisation, and ensuring alignment with human expectations~\cite{guidotti2018survey}.

By offering transparency in AI decision-making, XAI enhances trust and facilitates debugging, regulatory compliance, and the responsible deployment of AI systems across various fields.

\subsection{The Role of Explainability for Diverse Stakeholders}

\begin{figure}[H]
    \centering
    \includegraphics[width=\textwidth]{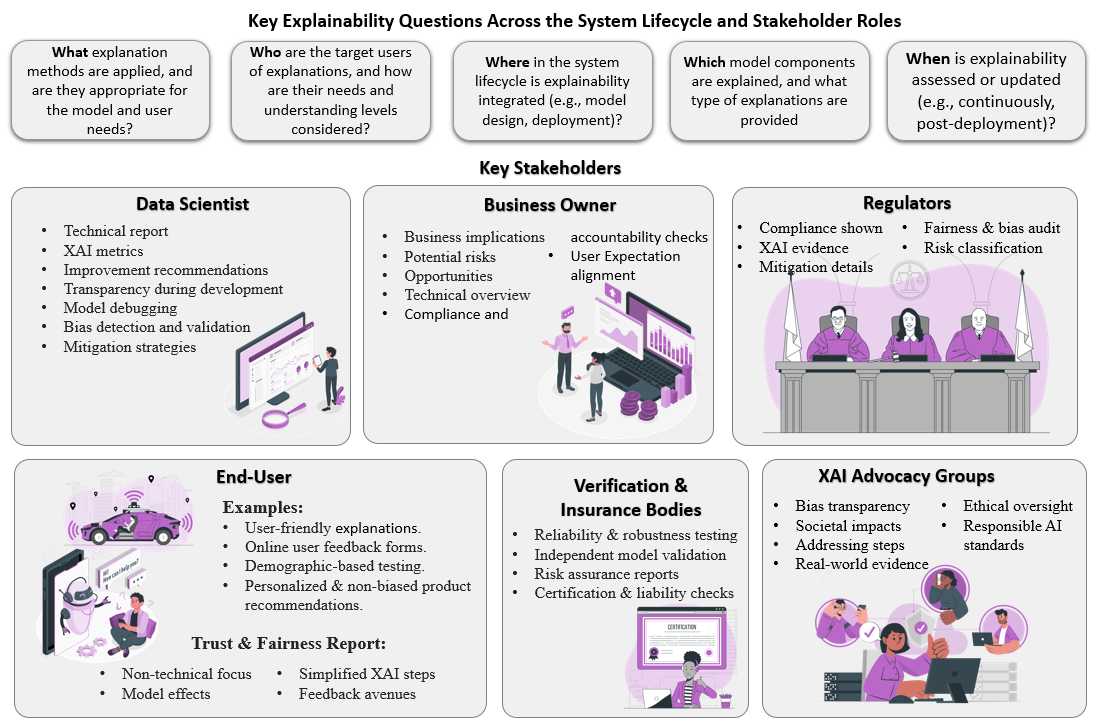}
    \caption{Stakeholder-specific explainability requirements. Explainability in AI systems must address the diverse needs of data scientists, business owners, regulators, end-users, verification bodies, and advocacy groups to ensure transparency, accountability, and the trustworthy deployment of AI systems.}
    \label{fig:stakeholders}
\end{figure}

As explainable artificial intelligence (XAI) becomes increasingly embedded in real-world applications, it is essential to recognise the diverse needs of stakeholders involved in the development, deployment, and oversight of these systems~\cite{doshi2017towards, samek2017explainable, rudin2019stop, arrieta2020explainable, lipton2018mythos, miller2019explanation}. These stakeholders include data scientists, business owners, regulators, end-users, verification and insurance bodies, and advocacy groups. Each group brings distinct expectations and requirements concerning model transparency, reliability, fairness, and accountability. Fig.~\ref{fig:stakeholders} summarises these stakeholder categories and their explainability priorities across the AI system lifecycle.

Data scientists play a pivotal role in developing, analysing, and improving AI models. Their need for explainability is twofold: first, to understand and optimise model behaviour during development~\cite{guidotti2018survey, molnar2020interpretable}; and second, to ensure that the deployed model operates reliably and transparently in real-world contexts~\cite{arrieta2020explainable, murdoch2019interpretable}. Explainability enables them to diagnose model failures, identify and validate biases, evaluate mitigation strategies, and ensure transparency throughout the model lifecycle~\cite{samek2017explainable, ribeiro2016should, gilpin2018explaining}. This is particularly critical for complex architectures such as large language models (LLMs) and image generation systems, whose decision-making processes are often opaque and complex to interpret~\cite{bommasani2021opportunities, wei2022emergent}.

Business owners, including AI product managers and enterprise clients, depend on explainability to determine whether a model aligns with user expectations, complies with industry regulations, and delivers tangible business value~\cite{adadi2018peeking, vinuesa2020role}. It supports strategic decision-making by revealing potential risks, clarifying model behaviour, and ensuring accountability and compliance in AI-driven processes. Explainability also contributes to organisational trust, improving communication between technical developers and non-technical stakeholders~\cite{arrieta2020explainable, miller2019explanation}.

For regulators, explainability forms the foundation for verifying fairness, robustness, and adherence to ethical and legal standards~\cite{goodman2017european, euai2024act, marchetti2024artificial}. It provides the transparency required to perform fairness and bias audits, assess mitigation measures, and classify risk levels, particularly in high-stakes domains such as healthcare, finance, and autonomous systems~\cite{vilone2021notions, floridi2018ai4people}. Regulators rely on well-documented XAI evidence and audit trails to ensure responsible deployment and to sustain public trust in automated decision-making~\cite{mohseni2021multidisciplinary}.

End-users, ranging from chatbot users and creative tool designers to professionals in healthcare and autonomous driving, require understandable explanations, trustworthy, and actionable~\cite{rudin2019stop, doshi2017towards, miller2019explanation}. User-oriented explainability enhances confidence in AI outputs by providing interpretable reports, intuitive visual summaries, and mechanisms for user feedback or dispute resolution~\cite{rosenfeld2019explainability, hoffman2018metrics}. This user-facing transparency is essential for maintaining fairness, accessibility, and trust across diverse usage contexts.

Verification and insurance bodies act as intermediaries between developers and regulators. Their role centres on evaluating model reliability and robustness, conducting independent validation, and issuing certification or liability assessments that confirm the trustworthy performance of the model across varied operational conditions~\cite{vinuesa2020role, vilone2021notions}. Explainability supports these efforts by providing traceable evidence for verification and compliance audits, enabling risk mitigation and certification in safety-critical applications.

Advocacy groups, including ethical research communities and non-governmental organisations, advance explainability at the societal level~\cite{floridi2018ai4people, jobin2019global}. They promote transparency and accountability by identifying bias, assessing societal impacts, and encouraging the adoption of responsible AI standards~\cite{cath2018governing}. Through public engagement, independent monitoring, and ethical oversight, these groups help bridge the gap between technological innovation and social responsibility, ensuring that explainability contributes to broader societal trust.

Addressing the diverse needs of these stakeholders represents not only a technical challenge but also a regulatory and ethical imperative. The emergence of frameworks such as the European Union’s AI Act underscores the growing importance of explainability, transparency, and accountability throughout the AI system lifecycle~\cite{euai2024act, marchetti2024artificial}. Moreover, as AI technologies evolve toward increasingly complex paradigms, such as large language models and image-editing or generation systems, explainability becomes vital for ensuring interpretability, safety, and trustworthiness~\cite{bommasani2021opportunities, rombach2022high}. These models, with their intricate architectures and generative capacities, highlight the need for stakeholder-centred transparency that extends beyond performance metrics to encompass fairness, interpretability, and governance. In particular, high-risk applications such as healthcare and autonomous driving demand rigorous explainability standards to safeguard users, promote fairness, and ensure compliance across both technical and societal dimensions.

\subsection{Types of Explainability Methods}

Explainability techniques are broadly categorised into two main types: \textit{intrinsic} and \textit{post-hoc} methods~\cite{adadi2018peeking}. The distinction between them lies in \textit{when} and \textit{how} interpretability is achieved, either as an inherent property of the model or as an analytical process applied after model training.

\textbf{Intrinsic explainability} refers to models that are inherently interpretable due to their simple and transparent structure. Typical examples include linear regression, decision trees, and rule-based models, where the relationship between input features and the resulting prediction can be explicitly inspected. Each component of these models, such as a regression coefficient or a decision split, directly illustrates how input variables contribute to the output. Such models are advantageous when interpretability and transparency are prioritised over predictive performance. However, their simplicity limits their ability to represent complex, nonlinear relationships or high-dimensional data distributions. Consequently, they are often unsuitable for modern generative and multimodal models that rely on deep hierarchical representations~\cite{rudin2019stop}.

\textbf{Post-hoc explainability}, in contrast, is applied to complex “black-box” models such as deep neural networks and transformer-based architectures. These models achieve state-of-the-art performance but are inherently opaque, making it difficult to understand the reasoning behind their outputs. Post-hoc methods aim to interpret such models without modifying their internal structure by analysing their behaviour after training. For example, LIME (Local Interpretable Model-Agnostic Explanations) constructs simple surrogate models around specific predictions to approximate local decision boundaries~\cite{ribeiro2016should}. SHAP (SHapley Additive exPlanations) uses cooperative game theory to assign fair contribution values to input features~\cite{slack2020fooling}. Grad-CAM (Gradient-weighted Class Activation Mapping) visualises class-specific gradients to highlight salient regions in input images that influence a model’s decision~\cite{selvaraju2017grad}. Attention visualisation techniques, commonly used in transformer models, display learned attention weights that reveal which tokens or image patches the model focuses on during inference~\cite{vaswani2017attention}.

These post-hoc approaches are particularly valuable for analysing large language models (LLMs) and multimodal systems, such as instruction-based image editing models, where explanations must integrate textual and visual reasoning. While they do not guarantee complete transparency, they provide meaningful insight into model decision-making and contribute to the growing interpretability of high-capacity generative systems.




\subsection{Challenges of Current Explainability Methods} 

Despite advances in explainability research, most existing methods exhibit notable limitations. While techniques like LIME, SHAP, Grad-CAM, and attention visualisations have improved our ability to interpret complex AI models, they are far from perfect. Many methods offer only partial explanations, lack consistency across different inputs, or yield results that are difficult for non-experts to understand. Furthermore, as models grow larger and more multimodal, combining text, images, and audio, the challenges of generating meaningful and reliable explanations become even greater~\cite{rudin2019stop, bommasani2021opportunities}.

Current explainability techniques often struggle with providing faithful explanations for black-box models, balancing interpretability with fidelity, and addressing domain-specific data challenges. Inconsistencies between explanation outputs across different runs or inputs undermine user trust and hinder the practical deployment of explainable AI in critical domains such as healthcare, finance, and autonomous driving~\cite{doshi2017towards, rudin2019stop}.

These challenges highlight the urgent need for developing more robust, generalisable, and domain-adaptive explanation methods that can scale to the complexity of modern AI systems~\cite{samek2017explainable, bommasani2021opportunities}.
\subsection{Limitations of Existing Explainability Techniques}

LIME (Local Interpretable Model-Agnostic Explanations) constructs surrogate models to approximate feature importance~\cite{ribeiro2016should}; however, a key limitation is that its explanations can be inconsistent across different runs due to random perturbations. SHAP (Shapley Additive Explanations) applies game theory, specifically Shapley values, to quantify feature contributions~\cite{lundberg2017unified}, but it suffers from high computational expense, making it impractical for real-time applications.

Several extensions to LIME have been proposed to address its weaknesses. BayLIME incorporates Bayesian priors to reduce instability~\cite{zhao2021baylime}, yet it remains susceptible to adversarial perturbations and depends on prior assumptions. S-LIME (Stabilised LIME) attempts to improve consistency by determining the required number of perturbation samples~\cite{zhou2021s}, but this increases computational complexity.

Broadly, post-hoc explainability methods face several general challenges. \textbf{Perturbation-based instability} arises because methods relying on random samplings, such as LIME, may produce inconsistent explanations across different runs~\cite{ribeiro2016should, zhou2021s}. There is also a \textbf{difficulty in capturing multimodal dependencies}, as conventional techniques struggle to interpret interactions between modalities like vision and text~\cite{bommasani2021opportunities, li2023blip}. Additionally, \textbf{high computational costs} are common, particularly in backpropagation-based methods like SHAP and Layer-wise Relevance Propagation (LRP)~\cite{lundberg2017unified, bach2015pixel}. \textbf{Limited local fidelity} is another issue since surrogate models used in post-hoc techniques may oversimplify complex decision boundaries~\cite{ribeiro2016should}.

\textbf{Lack of stability and generalisation} also remains a critical problem. Explanations often depend heavily on the specific dataset and model architecture, reducing their broader applicability~\cite{doshi2017towards, zhou2021s}. Furthermore, most explainability methods struggle when applied to \textbf{non-open-source models} or black-box commercial systems, where internal weights, architectures, or data pipelines are inaccessible~\cite{rudin2019stop, hohman2019gamut}. This limits the effectiveness of perturbation-based or gradient-based techniques, which typically require detailed access to model internals or controlled prediction environments.

\begin{figure}[H]
    \centering
    \includegraphics[width=0.95\textwidth]{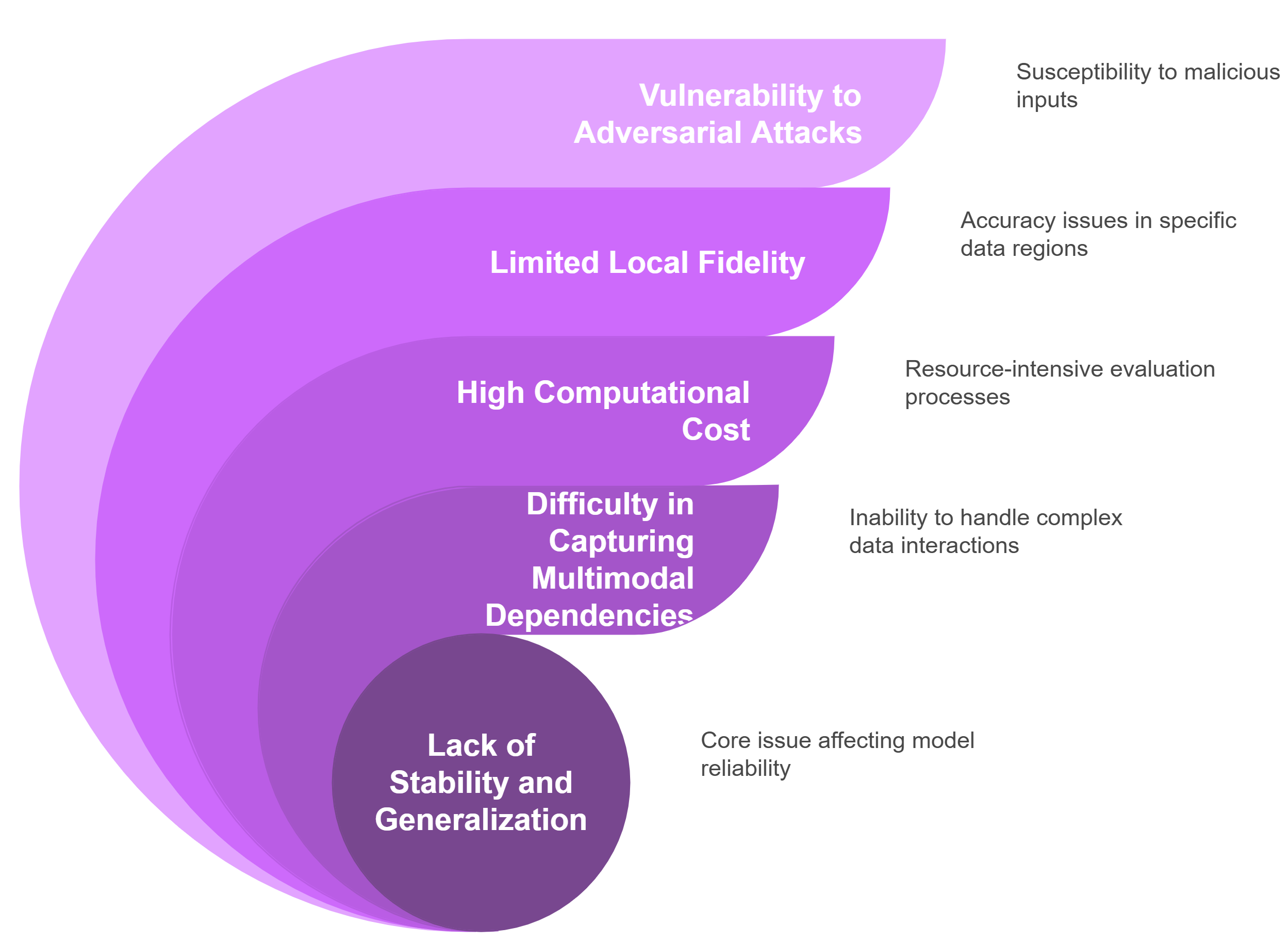} 
    \caption{Key challenges in AI model evaluation and explainability.}
    \label{fig:explainability-challenges}
\end{figure}

Finally, many post-hoc methods are vulnerable to \textbf{adversarial attacks}. Attackers can manipulate explanations without altering the model's predictions or performance, potentially leading to misleading or deceptive explanations~\cite{slack2020fooling, zhang2020interpretable}. This raises serious concerns about the robustness and trustworthiness of existing XAI tools, particularly in sensitive applications such as healthcare, finance, and autonomous systems. These challenges are summarised in Fig.~\ref{fig:explainability-challenges}, which illustrates the significant limitations of current explainability techniques.

\subsection{Adversarial Attacks on Explainability}
Explainability techniques are also vulnerable to adversarial manipulations. Malicious actors can modify input data to mislead explainability models, generating false or misleading explanations~\cite{slack2020fooling, zhang2020interpretable, rjoub2023survey}. This is particularly concerning in applications where AI-generated captions or image edits influence medical diagnoses~\cite{holzinger2017we}, forensic investigations~\cite{vakhshiteh2021adversarial}, or autonomous decision-making~\cite{finlayson2018adversarial}.

SMILE enhances robustness against adversarial attacks by using statistical distance measures, making it more resistant to manipulated inputs. Unlike conventional methods, which may be fooled by minor perturbations, SMILE compares input-output relationships at a distributional level, ensuring that explanations remain consistent and resilient to adversarial distortions. By incorporating ECDF-based measures~\cite{li2022ecod}, SMILE mitigates adversarial bias and strengthens interpretability in high-risk applications.

\subsection{Accuracy and Faithfulness of Explanations}

A critical challenge for current explainability methods is ensuring the accuracy and faithfulness of the generated explanations. Many popular techniques, such as LIME, SHAP, and attention-based methods, approximate model behaviour rather than reflecting it precisely. As a result, the explanations provided may not always align with the actual decision-making processes of the underlying AI models~\cite{rudin2019stop, slack2020fooling}.

This lack of faithfulness can lead to misleading interpretations. For example, surrogate models used in methods like LIME may oversimplify complex decision boundaries, providing explanations that are easy to understand but not fully representative of the model's actual logic~\cite{ribeiro2016should}. Similarly, gradient-based techniques such as Grad-CAM may highlight salient features that correlate with outputs but do not necessarily drive the model’s decisions~\cite{selvaraju2017grad}.

Ensuring explanation fidelity is particularly challenging in large, multimodal models that process diverse data types, including text, images, and audio. In such models, the internal representations are distributed and dynamic, making it challenging to trace clear causal links between inputs and outputs~\cite{bommasani2021opportunities}.

Improving the accuracy of explanations is crucial for fostering user trust, facilitating informed decision-making, and meeting regulatory requirements, particularly in sensitive domains such as healthcare and autonomous driving~\cite{holzinger2017we, euai2024act}. Researchers are actively developing techniques that enhance the fidelity of explainability methods without sacrificing interpretability or computational efficiency.

\subsection{Scalability and Computational Cost}

Another significant challenge facing explainability techniques is scalability. Many explanation methods, particularly perturbation-based approaches such as SHAP, require extensive computational resources to generate accurate attributions. As models grow more complex, such as large language models (LLMs) and multimodal transformers, the time and resources needed to produce reliable explanations increase substantially~\cite{lundberg2017unified, bommasani2021opportunities}.

This computational burden can limit the practical deployment of explainable AI in time-sensitive or resource-constrained environments. Balancing explanation accuracy with computational efficiency remains an open research challenge~\cite{samek2017explainable}.

\section{LIME and Its Evolving Adaptations} 

Among post-hoc explainability methods, Local Interpretable Model-Agnostic Explanations (LIME) has been widely adopted due to its flexibility and model-agnostic design~\cite{ribeiro2016should}. LIME is designed to explain individual predictions by fitting a simple, interpretable model around a prediction of interest, thereby approximating the complex model's behaviour within a local region.

\begin{figure}[H]
    \centering
      \begin{adjustbox}{center, max width=\textwidth}

    \begin{tikzpicture}[
        node distance=2cm,
        every node/.style={rectangle, draw=black, rounded corners, font=\small, align=left},
        main/.style={fill=blue!40, minimum width=10cm, text width=10cm, text centered, font=\Large},
        branch/.style={fill=blue!20, minimum width=8cm, text width=8cm, text centered, font=\large},
        subbranch/.style={fill=white, minimum width=3cm, text width=3cm, text centered, font=\small},
        description/.style={fill=white, text width=12cm, font=\small, align=justify}
    ]

    \node[main] (root) {Changes in LIME};

    \node[branch, below=0.4cm of root, xshift=-0.75cm] (surrogate) {Use Different Surrogate Model};

    \node[subbranch, below=0.4 of surrogate, xshift=-2cm] (qlime) {Q-LIME};
    \node[description, right=0.5cm of qlime] {Uses a quadratic surrogate model instead of a linear one to better capture non-linear relationships~\cite{bramhall2020qlime}};
    \node[subbranch, below=0.8cm of qlime] (slime) {S-LIME};
    \node[description, right=0.5cm of slime] {Focuses on creating sparse, interpretable explanations by selecting the most relevant features~\cite{zhou2021s}};
    \node[subbranch, below=0.4cm of slime] (bayLime2) {Bay-LIME};
    \node[description, right=0.5cm of bayLime2] {The surrogate model is extended to be a Bayesian linear model~\cite{zhao2021baylime}};
    \node[subbranch, below=0.4cm of bayLime2] (alime) {ALIME};
    \node[description, right=0.5cm of alime] {Employs autoencoders to learn a compressed representation of the input data~\cite{shankaranarayana2019alime}};

    \node[branch, below=0.5cm of alime, xshift=2cm] (distance) {Change Distance Parameter};

    \node[subbranch, below=0.5cm of distance,xshift=-2cm ] (smile) {SMILE};
    \node[description, right=0.5cm of smile] {Uses statistical distance measures (Wasserstein distance instead of Cosine distance)~\cite{aslansefat2023explaining}};

    \node[branch, below=0.5cm of smile, xshift=2cm] (sampling) {Change Sampling Technique};

    \node[subbranch, below=0.5cm of sampling, xshift=-2cm] (slime2) {S-LIME};
    \node[description, right=0.5cm of slime2] {Uses hypothesis testing and the Central Limit Theorem to determine the needed perturbation points~\cite{upadhyay2021extending}};
    \node[subbranch, below=0.7cm of slime2] (anchor) {Anchor};
    \node[description, right=0.5cm of anchor] {Identifies key features through coefficients matching the function's gradient~\cite{garreau2020explaining}};
    \node[subbranch, below=0.7cm of anchor] (uslime) {US-LIME};
    \node[description, right=0.5cm of uslime] {Chooses data samples close to the decision boundary and near the original data point~\cite{saadatfar2024us}};
    \node[subbranch, below=0.5cm of uslime] (guidedlime) {Guided-LIME};
    \node[description, right=0.5cm of guidedlime] {Uses FCA for structured sampling of instances~\cite{sangroya2020guided}};
    \node[subbranch, below=0.4cm of guidedlime] (dlime) {DLIME};
    \node[description, right=0.5cm of dlime] {Employs AHC and KNN to identify the relevant cluster~\cite{zafar2021deterministic}};
    \node[subbranch, below=0.6cm of dlime] (lslime) {LS-LIME};
    \node[description, right=0.5cm of lslime] {Focuses sampling on relevant parts of the decision boundary rather than the prediction itself~\cite{laugel2018defining}};
    \node[subbranch, below=0.8cm of lslime] (melime) {MeLIME};
    \node[description, right=0.5cm of melime] {Allows flexible perturbation sampling and different interpretable models to generate robust explanations~\cite{botari2020melime}};
    \node[subbranch, below=0.5cm of melime] (rllime) {RL-LIME};
    \node[description, right=0.5cm of rllime] {Uses reinforcement learning to optimize local interpretability~\cite{yoon2019rl}};
    \node[subbranch, below=0.3cm of rllime] (rlime) {R-LIME};
    \node[description, right=0.5cm of rlime] {optimizes the sampling region by expanding a rectangular region~\cite{ohara2024r}};

    \node[branch, below=0.3cm of rlime, xshift=2cm] (optimize) {Optimize LIME};

    \node[subbranch, below=0.5cm of optimize, xshift=-2cm] (optlime) {OptiLIME};
    \node[description, right=0.5cm of optlime] {Balances explanation stability and model fidelity using mathematical methods~\cite{visani2020optilime}};
    \node[subbranch, below=0.8cm of optlime] (glime) {G-LIME};
    \node[description, right=0.5cm of glime] {Incorporates global context and advanced techniques such as ElasticNet and LARS~\cite{li2023g}};

    \end{tikzpicture}
        \end{adjustbox}

   \caption{Overview of modifications and extensions in the LIME framework for model interpretability.}

    \label{fig:lime_changes}
\end{figure}

LIME's flexibility makes it particularly useful for applications where interpretability is critical, such as medical diagnosis, credit scoring, and fraud detection. Its model-agnostic nature allows it to be applied to diverse AI models without requiring knowledge of their inner workings.

Recently, researchers have advanced LIME by adapting it to specific data types and domains. Variants include PointNet LIME for 3D point cloud data~\cite{levi2024fast}, TS-MULE for time series~\cite{schlegel2021ts}, Graph LIME for graph-structured data~\cite{huang2022graphlime}, Sound LIME for audio~\cite{mishra2017local}, Sig-LIME for signal data~\cite{abdullah2024sig}, DSEG-LIME for image segmentation~\cite{knab2024dseg}, and B-LIME for ECG signals~\cite{abdullah2023b}. Each variant tailors LIME to address domain-specific challenges, improving relevance and interpretability.

Additionally, LIME-Aleph incorporates inductive logic programming to generate rule-based explanations, enhancing regulatory compliance and decision support~\cite{rabold2019enriching}. Beyond these adaptations, researchers have refined LIME’s surrogate models, distance metrics, sampling strategies, and parameters to improve explanation fidelity and computational efficiency. Fig.~\ref{fig:lime_changes} summarises some of the most notable LIME-based methods.
\subsection{How LIME Works}

Local Interpretable Model-Agnostic Explanations (LIME)~\cite{ribeiro2016should} is a widely used method for explaining the predictions of complex black-box models. The key idea behind LIME is to approximate the behaviour of a model in the vicinity of a specific input using a simpler, interpretable surrogate model, typically linear regression.

LIME generates a set of perturbed samples around the input instance and feeds them into the black-box model to obtain predictions. Each perturbed instance is assigned a weight based on its proximity to the original input, typically using a kernel function (e.g., an exponential kernel based on Euclidean distance). These samples, their corresponding model outputs, and their proximity-based weights are then used to fit the surrogate model. The coefficients of the surrogate model provide interpretable feature attributions that explain the model's local behaviour.

\begin{figure}[H]
    \centering
    \includegraphics[width=1\textwidth]{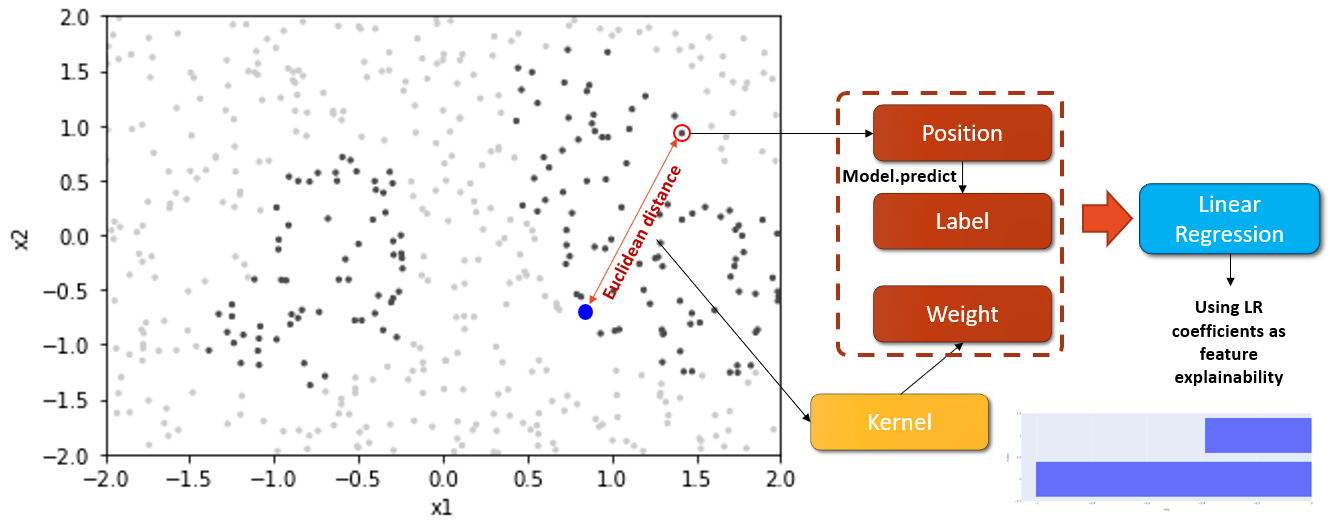}
    \caption{Illustration of how LIME approximates a complex model locally using weighted samples and linear regression~\cite{ribeiro2016should}.}
    \label{fig:LIME_work}
\end{figure}

Fig.~\ref{fig:LIME_work} illustrates this process using a 2D example, where local linear regression is fitted to the perturbed samples around a query point, and the distance weighting ensures locality.

\section{SMILE}  

As machine learning continues to evolve, the models we build are becoming more powerful, but also more complex and challenging to interpret, which is especially true in high-stakes fields like healthcare, autonomous driving, and instruction-based image editing, where understanding \textit{why} an AI system made a specific decision is just as important as the decision itself. These models are often seen as ``black boxes,'' making it hard for users to trust their outputs. To address this challenge, we chose SMILE (Statistical Model-agnostic Interpretability with Local Explanations) as the foundation for our research on explainable AI. This foundation lays the groundwork for the subsequent chapters, where SMILE's development, application, and evaluation are examined in detail.

\subsection{How SMILE Works}

SMILE interprets the behaviour of complex AI models by building a simple, interpretable approximation around a specific input, offering a local explanation. The process begins by applying small perturbations to the original input, such as removing or masking individual words in a sentence or prompt. For each perturbed input, the output of the target model is recorded and compared to the original production.

To measure the influence of each input element, SMILE utilises statistical distance metrics, such as the Wasserstein distance, to quantify shifts in the output distribution relative to the original, unperturbed input. These distributional changes help estimate the model's sensitivity to each input feature.

A local surrogate model, typically a linear regression, is then trained using the perturbation data and their corresponding outputs. This surrogate assigns attribution scores to each input element, indicating its relative impact on the model’s prediction.

Unlike gradient-based or attention-based methods that require access to internal model parameters, SMILE operates in a model-agnostic fashion. It treats the target model as a black box, making it compatible with both text- and image-based generative models. The resulting attribution scores are visualised as heatmaps, allowing users to identify the most influential parts of the input intuitively. Due to its flexibility and scalability, SMILE is well-suited for a wide range of generative AI applications.
\begin{figure}[H]
    \centering
    \includegraphics[width=1\textwidth]{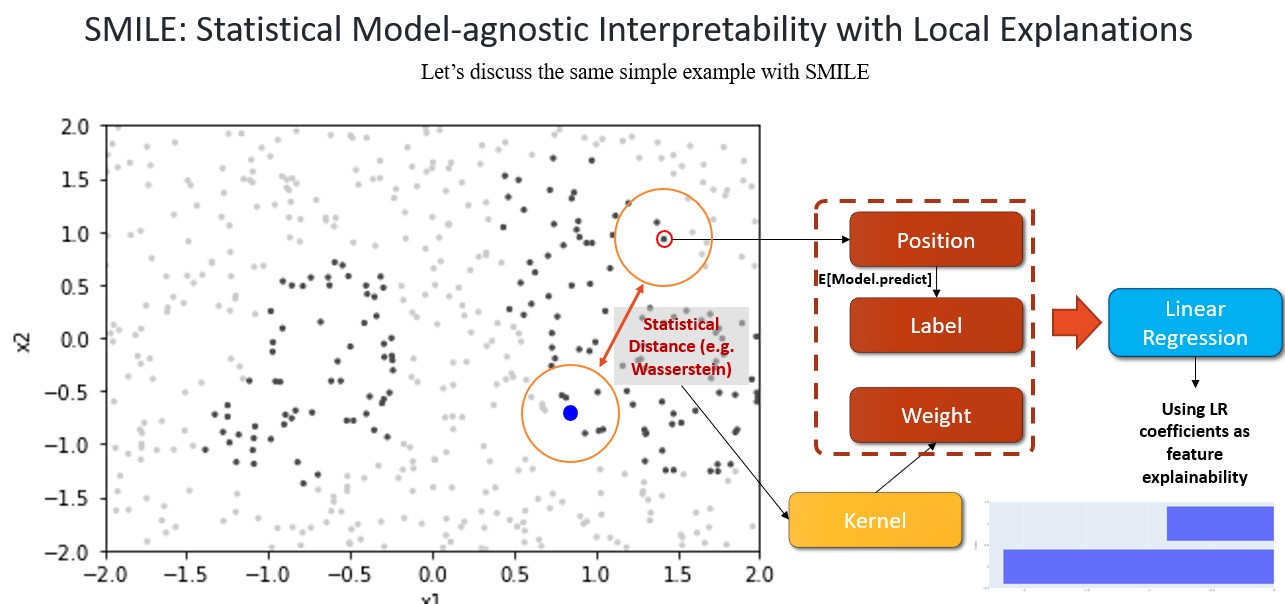}
    \caption{SMILE extends the LIME framework by using statistical distance metrics (e.g., Wasserstein) instead of direct Euclidean distances and builds local surrogate models on distributional shifts~\cite{aslansefat2023explaining}.}
    \label{fig:SMILE_work}
\end{figure}

\subsection{Advantages of SMILE}

What makes SMILE stand out is its \textbf{model-agnostic nature}. Unlike methods tied to a specific type of model or architecture, SMILE can be applied to any machine learning system without needing to access its internal workings~\cite{aslansefat2023explaining}. This flexibility is incredibly valuable in real-world applications, where models are often proprietary or too complex to be easily interpreted from the inside.

Another significant advantage is \textbf{robustness to adversarial attacks}. Some popular explainability methods, such as LIME, are easily fooled, meaning they can provide misleading explanations if the model is manipulated~\cite{slack2020fooling}. SMILE was explicitly designed to resist this kind of manipulation. It uses statistical distance measures, such as Wasserstein distance, to generate more stable and trustworthy explanations, even when the model itself has been tampered with or biased.

Finally, SMILE offers \textbf{more stable and consistent explanations} by moving beyond basic similarity measures like cosine or Euclidean distance. Instead, it uses advanced statistical comparisons between distributions of input data to capture the nuanced ways features affect model decisions~\cite{aslansefat2023explaining}, which is especially helpful in cases like image editing, where SMILE can produce heatmaps that clearly show which parts of a sentence or image influenced the model's output~\cite{dehghani2024mapping}. These visual cues make it easier for humans to understand and trust the model's behaviour.

In short, SMILE combines flexibility, robustness, and clarity in a way that few other explainability methods do. That is why it is a strong fit for our goal of creating AI systems that people can understand and rely on.

\section{Explainability in Generative AI}
As generative AI continues to evolve, the ability to understand and explain how these models produce their outputs has become increasingly important~\cite{bommasani2021opportunities, rudin2019stop}. Unlike traditional discriminative models, which perform well-defined tasks such as classification or regression, generative models operate in high-dimensional output spaces and generate open-ended content, including text, images, or audio. This complexity introduces substantial challenges for interpretability~\cite{gilpin2018explaining}.

Explainability in generative AI refers to understanding how specific inputs, such as words in a prompt or elements of an instruction, influence the resulting outputs~\cite{doshi2017towards, adadi2018peeking}. This is particularly relevant in large language models (LLMs) and instruction-based image editing systems, where subtle variations in input can lead to significantly different generations~\cite{zhao2024explainability, tang2022daam}. In high-stakes applications such as healthcare, autonomous driving, and legal reasoning, the inability to interpret such outputs raises concerns about trust, fairness, and accountability~\cite{bender2021dangers, samek2017explainable}.

This section reviews the state-of-the-art explainability techniques for generative AI, with a focus on two major domains: large language models and instruction-based image editing. The goal is to identify the limitations of existing approaches and motivate the need for a unified, robust, and model-agnostic framework, such as SMILE, that can provide consistent and intuitive explanations across modalities~\cite{aslansefat2023explaining, ribeiro2016should}.
\subsection{Explainability in Large Language Models}

Large Language Models (LLMs) have revolutionised natural language processing, enabling tasks such as text generation, translation, and question answering. However, their opaque ``black-box'' nature raises concerns regarding interpretability, fairness, and trustworthiness~\cite{zhao2024explainability, bender2021dangers}. Explainability in LLMs is crucial for ensuring accountability, enhancing user trust, and mitigating biases. This literature review explores existing research on explainability methods, applications, and challenges in LLMs.

\begin{figure}[H]
    \centering
    \includegraphics[width=0.9\textwidth]{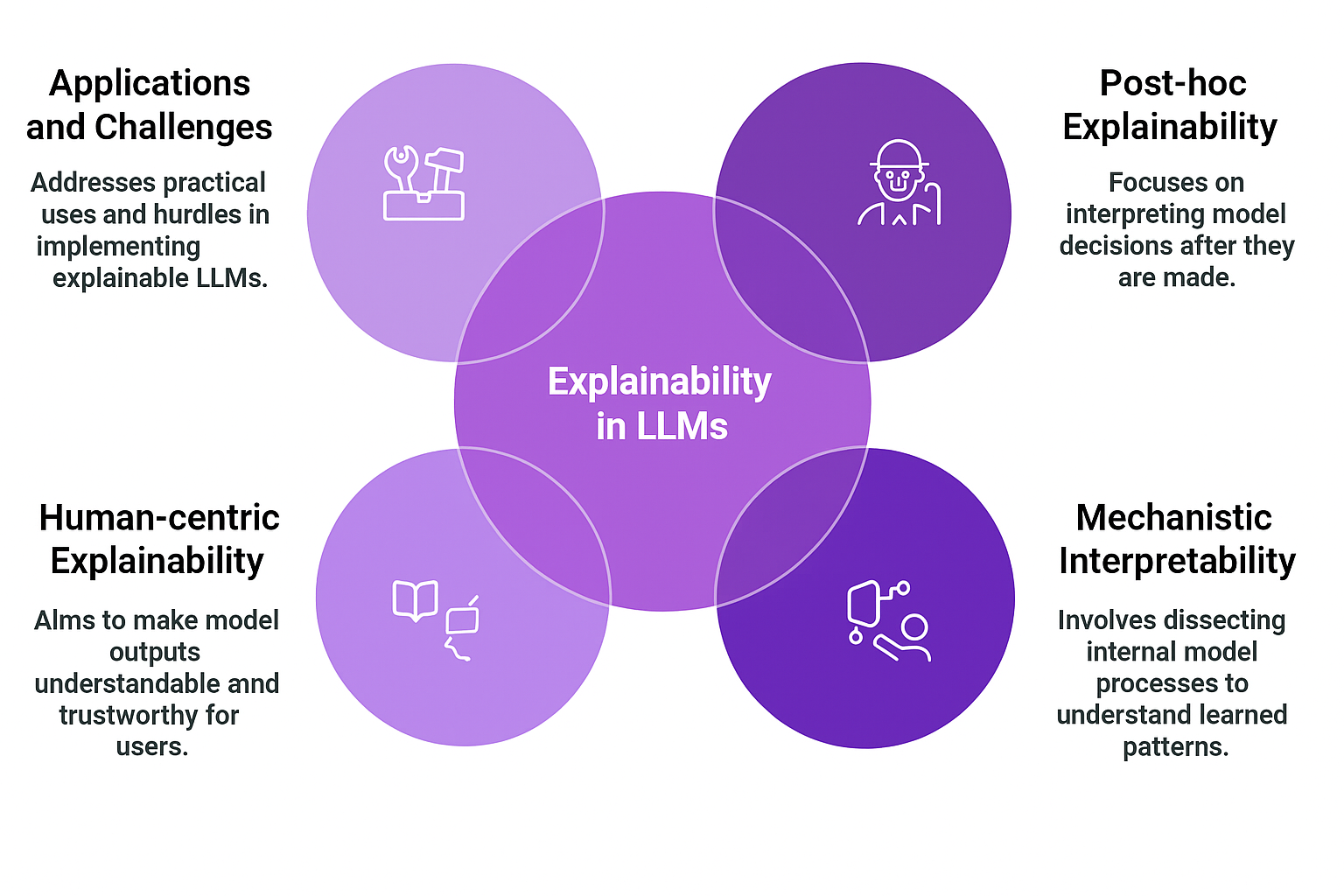}
    \caption{Categories and relationships of explainability techniques in Large Language Models.}
    \label{fig:llm_explainability_framework}
\end{figure}

An overview of the conceptual framework for explainability in LLMs is illustrated in Fig.~\ref{fig:llm_explainability_framework}, which highlights the four key dimensions: Post-hoc Explainability, Mechanistic Interpretability, Human-centric Explainability, and Applications and Challenges. These categories reflect the main approaches and practical considerations in enhancing the transparency and trustworthiness of LLMs.

The conceptual framework for explainability in LLMs includes four key dimensions: Post-hoc Explainability, Mechanistic Interpretability, Human-centric Explainability, and Applications and Challenges. These categories reflect the main approaches and practical considerations in enhancing the transparency and trustworthiness of LLMs.

After training, post-hoc explainability methods aim to interpret LLMs, providing insights into their decision-making processes~\cite{zhao2024explainability, madsen2022post}. Zhao et al.~\cite{zhao2024explainability} present a taxonomy of explainability techniques for LLMs, categorising approaches into local and global explanations, and discussing limitations and future research directions. Similarly, Bender et al.~\cite{bender2021dangers} highlight the risks of opaque LLMs, including biases and misinformation, and call for increased transparency and accountability. Recent surveys further expand this field by reviewing how LLMs contribute to explainable AI at large~\cite{mumuni2025explainable, bilal2025llms}. Amara et al.~\cite{amara2025concept} recently introduced ConceptX, a concept-level attribution method highlighting semantically meaningful input phrases through a coalition-based Shapley framework, offering robust and human-aligned explanations.

Mechanistic interpretability seeks to uncover how LLMs process information internally~\cite{bereska2024mechanistic, nanda2023progress, anthropic2024claude3}. 
Olah et al.~\cite{bereska2024mechanistic} propose reverse-engineering techniques to dissect internal representations. 
Nanda et al.~\cite{nanda2023progress} examine how LLMs develop abstract representations and generalisation, introducing progress measures for interpretability research. Recent work by Anthropic~\cite{anthropic2024claude3} demonstrates the role of mechanistic interpretability in understanding and controlling model behaviour. 
Additionally, the SEER framework~\cite{chen2025seer} introduces self-explainability mechanisms to enhance the interpretability of internal representations in LLMs. 
The CELL framework~\cite{luss2024cell} further advances mechanistic interpretability by integrating concept-based explanations directly into the training and representation learning of language models.

Recent advancements from the Transformer Circuits framework have further enriched mechanistic interpretability. Using causal scrubbing, Ameisen et al.~\cite{ameisen2025circuit} introduced Attribution Graphs to uncover emergent structures within neural networks. In contrast, Nanda et al.~\cite{nanda2024monosemanticity} investigated the scaling of monosemanticity, extracting interpretable features from state-of-the-art models. Additional recent contributions explore specific tasks such as relevance estimation~\cite{elhage1others}, emotion inference~\cite{tak2025mechanistic}, and geospatial reasoning~\cite{de2025geospatial}. These works employ techniques such as activation patching and sparse autoencoders to reveal the internals of LLMs. Sparse interpretability methods are further reviewed comprehensively in~\cite{elhage1others}. Neuroscientific perspectives have also been applied, using dynamical systems theory to model token-level trajectories within transformers~\cite{fernando2025transformer}.

Human-centric explainability focuses on making LLM outputs understandable to non-experts~\cite{ji2023survey, mondorf2024beyond, goethals2025if}. Ji et al.~\cite{ji2023survey} investigate hallucination in LLM outputs and propose techniques for detecting and mitigating incorrect responses. Krause et al.~\cite{mondorf2024beyond} assess reasoning abilities compared to human performance, and Martens et al.~\cite{goethals2025if} explore counterfactual explanations to improve human understanding. Recent works extend this perspective by enabling interactive explanations of black-box models~\cite{goethals2025if} and by benchmarking the quality of explanations in clinical domains~\cite{chen2025benchmarking}. These approaches improve the accessibility and transparency of LLM outputs in real-world decision-making.

Recent research has explored cost-efficient, model-agnostic explanation methods for LLMs~\cite{liu2025towards}, which align well with the objectives of gSMILE; their approach leverages sampling and scoring strategies to generate faithful explanations without requiring access to model internals.

Explainability enhances debugging, bias detection, regulatory compliance, and user trust in LLMs~\cite{adadi2018peeking}. Challenges remain, including defining ``meaning'' in LLM-generated outputs~\cite{adadi2018peeking}, managing the ethical and societal implications of black-box models, and striking a balance between performance and interpretability. Recent work has demonstrated SMILE's applicability beyond language, such as in instruction-based image editing~\cite{dehghani2024mapping}, highlighting its modality-agnostic nature.

\subsection{Explainability in Instruction-based Image Editing}

Instruction Image Editing models present unique challenges when it comes to explainability. These models convert textual descriptions into images involving complex, high-dimensional transformations. Consequently, the internal workings of these models could be more inherently interpretable. For instance, identifying which parts of the input text influence specific regions of the generated image remains challenging~\cite{lee2024diffusion, evirgen2024text, tang2022daam, dang2024explainable}.

Previous studies have employed local XAI methods, such as heatmaps, to map text features to corresponding image regions visually. DAAM~\cite{tang2022daam} introduces an innovative enhancement for capturing self-attention within images, and the text-image cross-attention is initially addressed. By guiding self-attention, DAAM-I2I~\cite{chowdhury2022daam} significantly improves the quality of heatmaps, leading to more precise localisation in segmentation and object detection tasks.

Recent research on robustness, fairness, security, privacy, factuality, and explainability addresses ethical, harmful, and social concerns~\cite{zhang2024trustworthy, hao2024harm}. 

Evirgen et al.~\cite{evirgen2024text} addressed user interaction challenges with instruction-based image editing models by emphasising the use of example-based explanations and curating datasets to better tailor explanations for beginners. Similarly, PromptCharm~\cite{wang2024promptcharm} introduced model explanations by visualising attention values, allowing users to refine their prompts and produce higher-quality outputs interactively.

Patcher~\cite{chang2024repairing} introduced a method for mitigating catastrophic neglect in instruction-based image editing models through attention-guided feature enhancement. This approach improves the alignment between prompts and generated content, directly addressing semantic inconsistencies. In addition to text-to-image generation, diffusion-based models have been applied to tasks such as open-vocabulary segmentation~\cite{karazija2023diffusion, karazija2024diffusion} and person detection dataset generation~\cite{rodriguez2024exploring}.

OVDiff~\cite{karazija2023diffusion} leverages pre-trained diffusion models for zero-shot segmentation without additional training. FOSSIL~\cite{barsellotti2024fossil} integrates text-conditioned diffusion models and self-supervised features for unsupervised segmentation, capturing semantic variability while improving explainability.

In video generation, Vico~\cite{yang2024compositional} analyses the token influence and balances latent updates, enhancing compositional accuracy in video generation. Similarly, MultiLate~\cite{vetagiri2024multilate} utilised attention attribution maps to enhance multimodal hate speech detection, demonstrating the impact of explainability on improving classification performance.

Methods for generating controllable RGBA illustrations~\cite{quattrini2024alfie} and fine-grained visuo-spatial representations for embodied AI tasks~\cite{gupta2024pre} showcase diffusion models' potential in creative workflows. Finally, Tankelevitch et al.~\cite{tankelevitch2024metacognitive} proposed integrating explainability to reduce cognitive load in generative workflows.

These contributions underscore the significance of explainability in enhancing the usability, trustworthiness, and transparency of diffusion-based generative AI systems across various applications.

In this work, we introduce SMILE (Statistical Model-agnostic Interpretability with Local Explanations) to create heatmaps that show how specific text elements influence image edits in instruction-based editing~\cite{aslansefat2023explaining}. SMILE, which is model-agnostic, is chosen for its compatibility with various models~\cite{ribeiro2016should}, drawing from methods similar to LIME but with improved robustness~\cite{gilpin2018explaining}. Unlike LIME, which can be vulnerable to adversarial attacks that distort interpretability, SMILE is more resistant to manipulation~\cite{aslansefat2023explaining}. These heatmaps make interactions with editing models more understandable and reliable, helping users trust the edits by offering insights that are less prone to distortion.

\subsection{Explainability in Multimodal AI}

With the rise of multimodal large language models (MLLMs) and vision language models (VLMs), recent surveys have emphasised that unimodal explanation techniques are insufficient to capture cross-modal dependencies. Comprehensive reviews of MLLMs and multimodal XAI propose taxonomies that integrate data, model, and inference-level explainability, ranging from token and embedding-level attribution to architectural and inference-based transparency~\cite{dang2024explainable,sun2024review}. 
In the vision domain, a recent survey on foundation models highlights the combination of post-hoc approaches (perturbation, counterfactual, and neuronal) with large-scale models, raising both opportunities and challenges for quantitative evaluation of explanations~\cite{kazmierczak2025explainability}. 
Similarly, surveys of VLMs emphasise issues of alignment, benchmarking, hallucination, fairness, and safety, underscoring the importance of trustworthy explanations in high-stakes applications~\cite{li2025survey}. This body of work motivates methods that quantify cross-modal influence while ensuring stability, fidelity, and standardisable evaluation, directions directly addressed in this thesis through the SMILE framework.


\section{Conclusion}

This chapter's review examines the rapid advancements in generative artificial intelligence (AI), with a particular focus on large language models (LLMs), instruction-based image editing, and explainability techniques. It examines how innovations in model architectures, such as transformers and diffusion models, have enhanced the capabilities of generative AI across various tasks and modalities.

For LLMs, significant efforts have been made to address the interpretability challenges inherent in large-scale models. Research has introduced diverse explainability methods, including post-hoc techniques, mechanistic interpretability, and human-centric approaches, each contributing to a deeper understanding of how these models generate and process information. Despite this progress, achieving a balance between high performance and transparency remains an ongoing challenge.

Advancements in instruction-based image editing have focused on improving the alignment between textual prompts and generated images. Studies have proposed innovative attention-based methods, heatmap visualisations, and interactive explanation tools to enhance model interpretability and user trust. However, the high dimensionality and complexity of these multimodal systems continue to pose significant hurdles for explainability.

Overall, integrating explainability methods into generative AI research has enhanced transparency and usability, contributing to the development of ethical and accountable AI practices. Continued interdisciplinary collaboration and methodological innovation will be essential to overcoming remaining challenges and advancing the responsible development of generative AI technologies.


\chapter{gSMILE Methodology}
\label{chap:gSMILE Methodology}

\section{Overview of the gSMILE Framework}

The proposed methodology employs the gMILE framework to enhance interpretability across text generation and image editing tasks. By analysing how specific input components influence model outputs, the method improves transparency and predictability in model behavior~\cite{bommasani2021opportunities, molnar2020interpretable}.

\begin{figure}[H]
    \centering
    \includegraphics[width=0.8\linewidth]{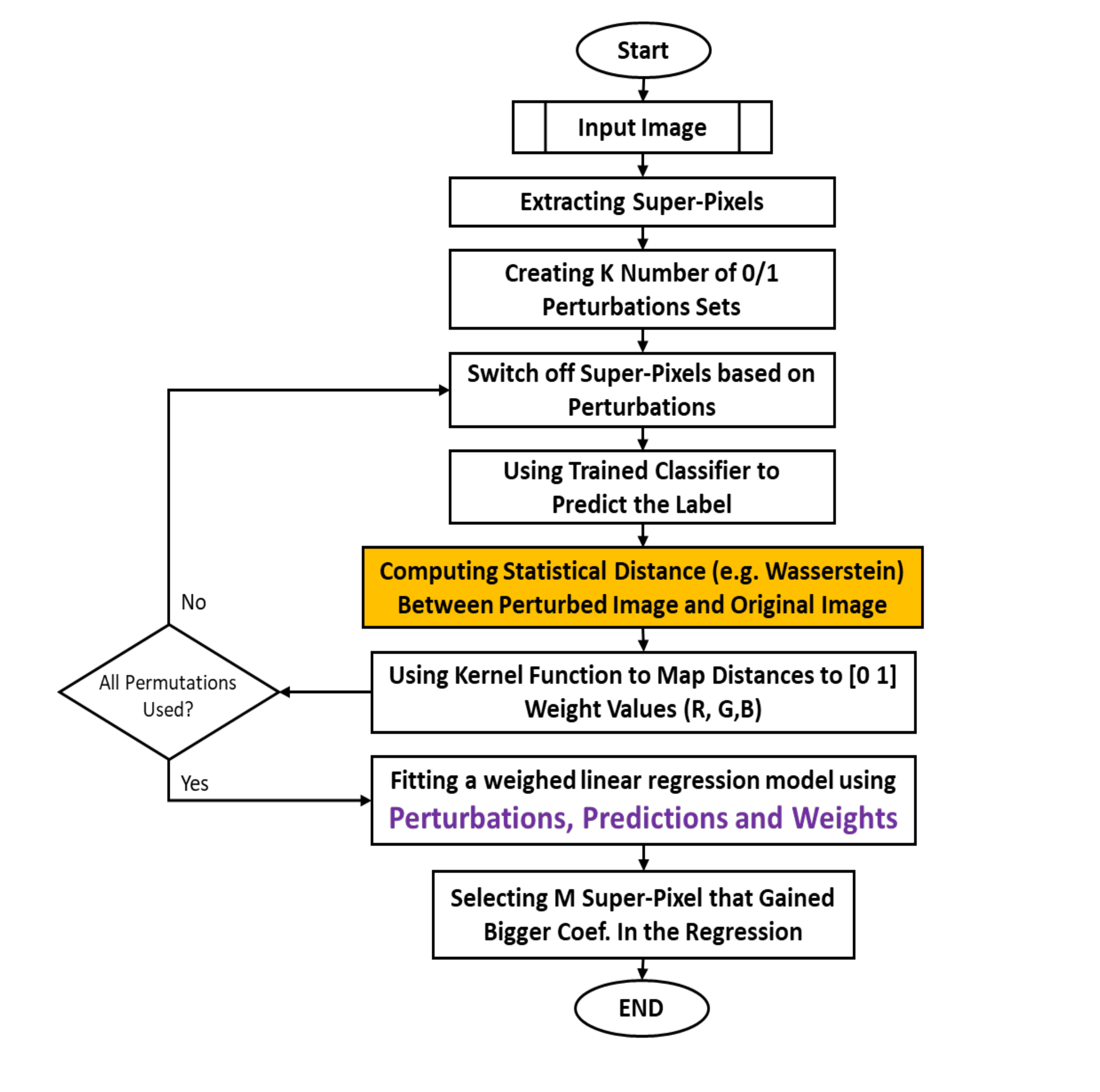}
    \caption{SMILE flowchart for explaining image classification~\cite{aslansefat2023explaining}}
    \label{fig:smile}
\end{figure}

Initially developed for image classification tasks, as illustrated in Fig.~\ref{fig:smile}~\cite{aslansefat2023explaining}, SMILE identifies critical features driving model decisions by segmenting inputs, such as super-pixels for images or logical sections for text and generating perturbations through selective modification of these components. These perturbations reveal the contribution of different input elements to the final output.

This work adapts gSMILE for two domains. For large language models (LLMs), the method assesses how variations in textual inputs influence the generated outputs, offering insights into the relationship between prompts and responses. gSMILE explains how different visual components and input instructions influence the edited outputs for instruction-based image editing models.

This unified interpretability framework clarifies decision-making processes in both textual and visual domains and supports the analyses presented in Chapters~\ref{chap:llm_explainability} and~\ref{chap:image_editing_explainability}.

\begin{figure*}[ht]
    \centering
    \includegraphics[width=0.9\linewidth]{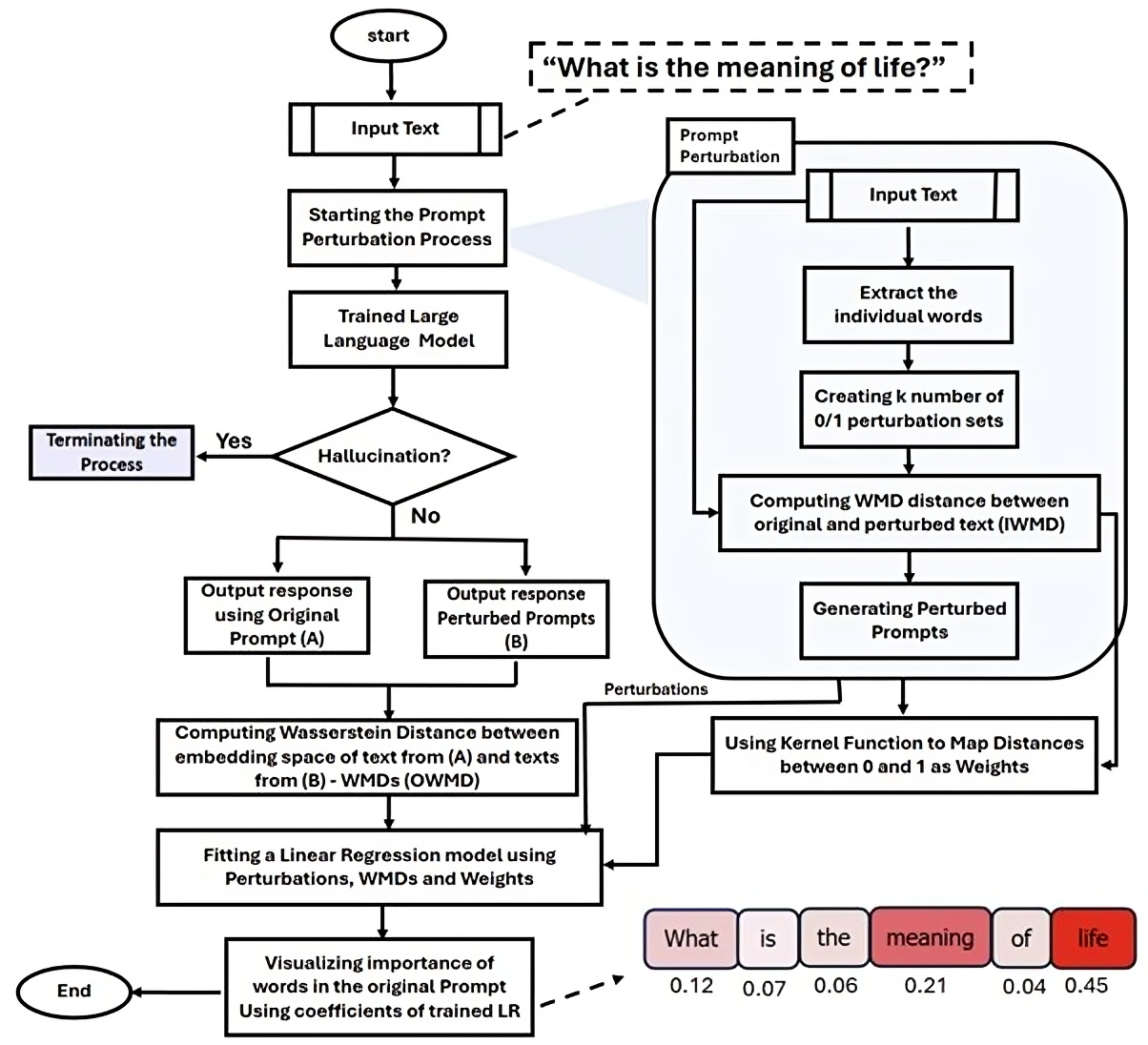}
    \caption{gSMILE flowchart for explaining text generation models}
    \label{fig:flow}
\end{figure*}

\begin{figure*}[ht]
    \centering
    \includegraphics[width=1\linewidth]{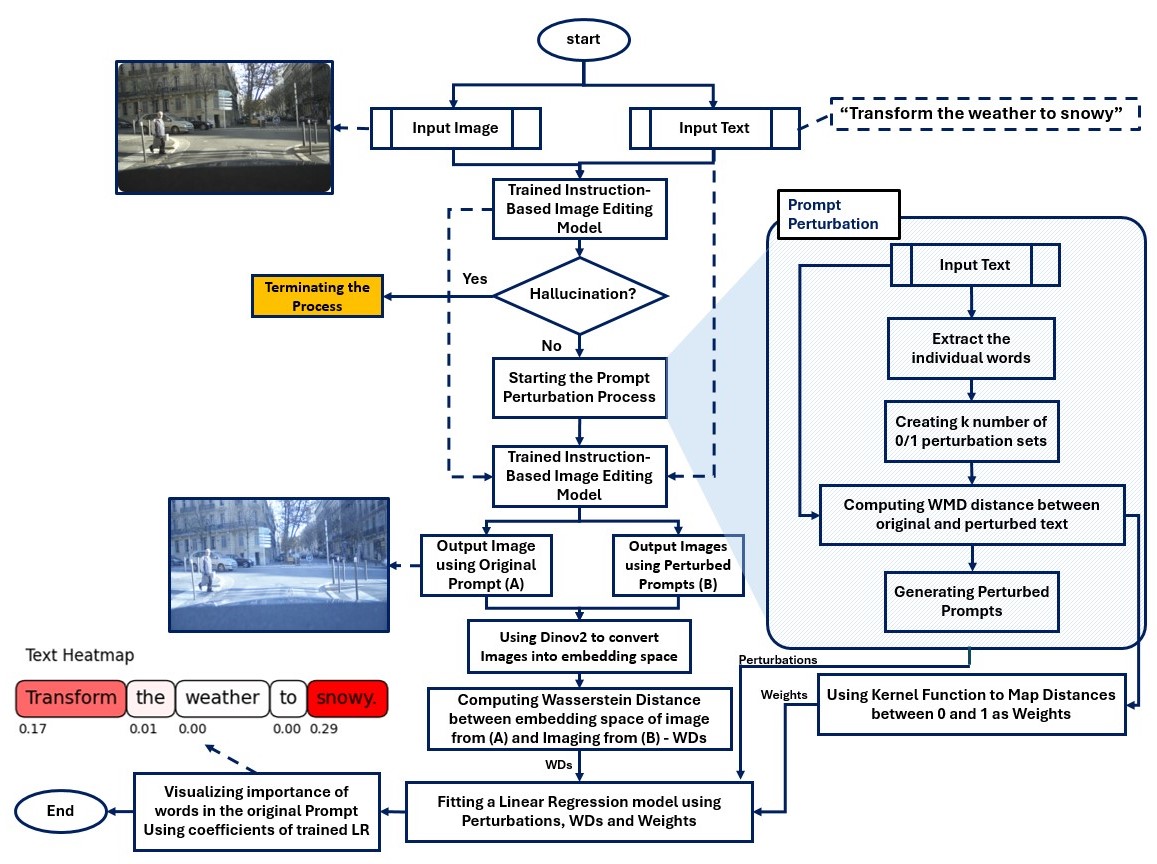}
    \caption{Flowchart of the adapted gSMILE framework for instruction-based image editing models.}

    \label{fig:flowchart}
\end{figure*}

Inspired by SMILE's capabilities, we propose a unified method for interpreting text-prompt-based generation models, applicable to both text and image outputs, as illustrated in Figs.~\ref {fig:flow} and~\ref {fig:flowchart}. Instead of perturbing the generated output directly, the instruction is modified by including or excluding certain words. A corresponding output is generated for each altered text prompt: text outputs for language models and images for image generation models. The Wasserstein distance is computed between each perturbed production and the output produced by the original text prompt. This distance is a similarity measure that allows us to identify which words in the text prompt have the most significant impact on the generated text or image.

These similarity scores are then used as the outcome variable in a weighted linear regression model, with weights based on the distances between the original and perturbed outputs. This regression model helps determine how each word in the prompt affects the final output. The coefficients derived from the model quantify the influence of each word, which is used to create a visual heatmap highlighting the most influential words. This heatmap provides an intuitive visual representation of which prompt elements contribute most to the generated output, whether text or an image.

By employing this enhanced three-step process, which builds on SMILE's methodology, we enable users to understand how particular elements of text prompts impact generation outcomes in both text and image domains~\cite{huang2024smartedit}. This technique enhances user control and predictability in instruction-driven generation models by providing insightful information about which aspects of textual input, such as stylistic instructions or descriptive terms, have the most significant impact on the final output.

In the case of text generation, the model receives various perturbed prompts and generates corresponding text outputs, with the only change being the prompt content. Similarly, for image editing, the model gets the same input image paired with different perturbed text prompts, generating images that vary solely based on critical components of the text input.

Different text permutations are created by breaking down the original prompt into individual words and selectively including or excluding certain words to generate various perturbation texts~\cite{liu2023toward}. The respective model produces outputs for each perturbed prompt: text for language models or images for instruction-based image editing models.

While instruction-based image editing models can face challenges with tasks such as counting objects and spatial reasoning, these issues are mitigated by carefully selecting perturbation texts and restricting the input domain~\cite{qiu2022benchmarking}. In both text and image domains, perturbations are designed to vary only the critical components of the prompt, minimising confounding factors and isolating the effect of each word.

Task-specific embedding models assess similarities between the original and perturbed outputs. For text generation, the Word Mover's Distance (WMD) model~\cite{kusner2015word} is applied to extract meaningful semantic embeddings and compute similarities between text outputs. For image generation, the DINOv2 model~\cite{oquab2023dinov2} is utilised to produce high-quality semantic embeddings that capture both visual and conceptual content. Unlike direct image comparisons, which may only reflect surface-level features such as colour or pixel structure, embedding-based comparisons capture deeper semantic shifts~\cite{caron2021emerging}.

This methodology supports a model-agnostic, explainable framework for interpreting text and image outputs. In the image domain, DINOv2's self-supervised learning approach provides general-purpose embeddings without requiring labelled data or model-specific outputs, making it suitable for diverse image types and adaptable across different editing models~\cite{oquab2023dinov2, caron2021emerging}. Similarly, WMD allows for flexible and meaningful comparison across varied text outputs independent of the underlying text generation model.

We use the Wasserstein distance to quantify differences between the outputs generated from the original and perturbed prompts in the text and image domains~\cite{rubner2000earth, arjovsky2017wasserstein}. This distance metric is chosen because it comprehensively measures differences between embeddings, capturing both broad and subtle distributional changes. Unlike other empirical cumulative distribution function (ECDF)-based distances, such as Kolmogorov–Smirnov or Cramér–von Mises, the Wasserstein distance effectively reflects both local and global variations in the data~\cite{rubner2000earth}.

In the \textbf{text domain}, the Wasserstein distance operates over embeddings derived from the Word Mover’s Distance (WMD) model~\cite{kusner2015word}. This approach captures distributional geometry and allows for a meaningful assessment of how perturbations conceptually align with the original prompt.

In the \textbf{image domain}, the Wasserstein distance is computed over embeddings generated by the DINOv2 model~\cite{oquab2023dinov2}. DINOv2 provides rich, semantic embeddings that capture visual and conceptual content, offering a more informative comparison than direct pixel-based measures.

Finally, by applying linear regression, the input prompt space is linked to the output space of the generation models. For text generation, the regression model determines how specific words influence the text output. Instruction-based image editing reveals how words affect the resulting images. In both cases, regression coefficients quantify the impact of individual words, enabling interpretability of the model’s behavior~\cite{murdoch2019interpretable}.

\subsection{Generating Outputs and Perturbed Prompts}

To examine how subtle changes in text prompts affect the outputs of both text generation and image editing models, we generate variations of the original prompt, referred to as \textit{perturbed prompts}~\cite{fu2023guiding, alivanistos2022prompting}. The original text prompt is denoted by \( x \), and perturbations are represented as \( \{\hat{x}_j\}_{j=1}^J \), where each \( \hat{x}_j \) is obtained through minor token-level modifications of \( x \).

For \textbf{text generation}, the black-box model \( \pi^{(n)} \) receives either the original prompt or a perturbed version. The output distribution for the original prompt is given by:
\begin{equation}
\pi^{(n)}(y \mid x),
\label{eq:original_text_generation}
\end{equation}
and for each perturbed prompt \( \hat{x}_j \), the model produces:
\begin{equation}
\pi^{(n)}(y \mid \hat{x}_j), \quad \forall j \in \{1, \dots, J\}.
\label{eq:perturbed_text_generation}
\end{equation}

For \textbf{instruction-based image editing}, each perturbed prompt \( \hat{x}_j \) is paired with a consistent input image \( \alpha_{\text{input}} \). Let \( \phi_{\text{edit}} \) denote the image editing function. The baseline edited image corresponding to the original prompt is:
\begin{equation}
\alpha'_{\text{org}} = \phi_{\text{edit}}(x, \alpha_{\text{input}}),
\label{eq:original_image_generation}
\end{equation}
And for each perturbed prompt:
\begin{equation}
\alpha'_{\text{pert}, j} = \phi_{\text{edit}}(\hat{x}_j, \alpha_{\text{input}}), \quad \forall j \in \{1, \dots, J\}.
\label{eq:perturbed_image_generation}
\end{equation}

This formulation ensures consistency across text and image domains. In both cases, the set of perturbations \( \{\hat{x}_j\}_{j=1}^J \) enables systematic evaluation of how individual prompt components influence the model’s outputs~\cite{ribeiro2016should, brooks2023instructpix2pix, zhu2023promptbench}.

\subsection{Creating the Interpretable Space}

To map the output spaces of both text generation and image editing models into a one-dimensional similarity space, we compute the Wasserstein distance between outputs generated from the original and perturbed prompts~\cite{arjovsky2017wasserstein, peyre2019computational}. This approach enables the consistent measurement of the influence of each perturbation across domains.

For \textbf{text generation}, we quantify the semantic distance between the baseline output distribution \( \pi^{(n)}(y \mid x) \) and the distribution for each perturbed prompt \( \pi^{(n)}(y \mid \hat{x}_j) \). Embeddings are extracted using the Word Mover’s Distance (WMD), which captures semantic relationships between words and sentences~\cite{kusner2015word}.

For \textbf{instruction-based image editing}, we similarly compare the baseline edited image \( \phi_{\text{edit}}(x, \alpha_{\text{input}}) \) with the outputs generated under perturbed prompts \( \phi_{\text{edit}}(\hat{x}_j, \alpha_{\text{input}}) \). To obtain semantically meaningful representations, we employ the DINOv2 model~\cite{oquab2023dinov2}, a self-supervised feature extractor that captures high-level visual and conceptual features.

The output-level semantic shift is defined as:
\begin{equation}
\Delta(x, \hat{x}_j) = W \big( \pi^{(n)}(y \mid x), \pi^{(n)}(y \mid \hat{x}_j) \big),
\label{eq:output-shift-ch3}
\end{equation}
Where \( W(\cdot, \cdot) \) denotes the Wasserstein distance between two distributions in the embedding space. In practice, \( W \) is computed either over WMD-based embeddings for text outputs or DINOv2 embeddings for image outputs. The norm order \( p \) in the Wasserstein-\( p \) metric controls sensitivity: smaller values emphasise local variations, while larger values capture broader distributional shifts~\cite{peyre2019computational}.

\begin{algorithm}[H]
\caption{Bootstrap-Based P-Value Calculation for Wasserstein Distance (text: OWMD using WMD embeddings; image: Wasserstein distance using DINOv2 embeddings)}
\label{alg:bootstrap}
\SetAlgoLined
\KwIn{Two sets of output embeddings: $X$ (baseline) and $Y$ (perturbed)}
\KwOut{Observed Outcome Wasserstein Distance $\Delta(x, \hat{x}_j)$ and corresponding p-value $pVal$}

$\texttt{MaxItr} \leftarrow 10^5$\;
$\Delta(x, \hat{x}_j) \leftarrow \texttt{Wasserstein\_Dist}(X, Y)$\;
$\texttt{XY} \leftarrow$ concatenate $X$ and $Y$\;
$\texttt{LX} \leftarrow$ length of $X$\;
$\texttt{LY} \leftarrow$ length of $Y$\;
$\texttt{bigger} \leftarrow 0$\;

\For{$i \leftarrow 1$ \KwTo $\texttt{MaxItr}$}{
    Draw a random sample $e$ of size $\texttt{LX}$ from $XY$\;
    Draw a random sample $f$ of size $\texttt{LY}$ from $XY$\;
    $\texttt{bootWD} \leftarrow \texttt{Wasserstein\_Dist}(e, f)$\;
    \If{$\texttt{bootWD} \geq \Delta(x, \hat{x}_j)$}{
        $\texttt{bigger} \leftarrow \texttt{bigger} + 1$\;
    }
}
$pVal \leftarrow \texttt{bigger} / \texttt{MaxItr}$\;
\Return{$\Delta(x, \hat{x}_j), pVal$}\;
\end{algorithm}

To ensure that observed distances correspond to meaningful differences rather than noise, we estimate statistical significance using a Bootstrap procedure~\cite{tibshirani1993introduction, gilleland2020bootstrap}. For each perturbation, we compute a p-value that reflects the probability of obtaining a Wasserstein distance as considerable as the observed one under random resampling. Perturbations with high p-values are discarded as non-significant~\cite{gulrajani2017improved}.

The Bootstrap procedure is summarised in Algorithm~\ref{alg:bootstrap}.

\subsection{Similarity Weighting with a Gaussian Kernel}

To ensure that the interpretable surrogate model appropriately accounts for the relevance of each perturbed prompt, we apply a similarity-based weighting scheme. Perturbations that are more semantically similar to the original input are assigned higher importance, while dissimilar perturbations receive lower weights. This reduces the influence of outliers or radically altered prompts.

For both \textbf{text generation} and \textbf{image editing}, similarity is computed at the input level, since perturbations are applied to the text prompts. We adopt the Word Mover’s Distance (WMD)~\cite{kusner2015word}, which computes the minimal cumulative distance required to transform the word distribution of the original prompt \( x \) into that of a perturbed prompt \( \hat{x}_j \):

\begin{equation}
\label{eq:wmd}
\mathrm{IWMD}(x, \hat{x}_j) = \min_{T \geq 0} \sum_{k=1}^{n} \sum_{l=1}^{m} T_{kl} \, d(w_k, w_l),
\end{equation}

Subject to the flow constraints:
\[
\sum_{l=1}^{m} T_{kl} = p_k, \quad \forall k \in S_x, \qquad
\sum_{k=1}^{n} T_{kl} = q_l, \quad \forall l \in S_{\hat{x}_j}, \qquad
T_{kl} \geq 0.
\]

Here, \(w_k\) and \(w_l\) denote word embeddings, and \(p_k, q_l\) are the normalized bag-of-words distributions for \(x\) and \(\hat{x}_j\), respectively. We denote this input-level semantic distance as:
\[
\delta_{x_j} = \mathrm{IWMD}(x, \hat{x}_j).
\]

To transform these distances into similarity-based weights for regression, we use a Gaussian kernel~\cite{ribeiro2016should}:

\begin{equation}
w_j = \exp \left( - \left( \frac{\delta_{x_j}}{\sigma} \right)^2 \right), 
\quad \forall j \in \{1, \dots, J\},
\label{eq:weights-ch3}
\end{equation}

Where \( w_j \in (0,1] \) is the relevance weight for perturbation \(\hat{x}_j\), and \(\sigma > 0\) is the kernel width parameter controlling the rate of exponential decay.

Here, \( \delta_{x_j} = \mathrm{IWMD}(x, \hat{x}_j) \) represents the WMD-based input distance, and \( \sigma \) controls the sensitivity of the Gaussian kernel~\cite{hastie2009elements}. The resulting weight \( w_j \) reflects the similarity of each perturbed prompt to the original and is used during the regression analysis to prioritise semantically closer perturbations.

\subsection{Developing the Interpretable Surrogate Model}

To interpret how different words in the prompt influence the model’s output, we train a local surrogate model that maps the relationship between perturbed input prompts and the corresponding output shifts. This approach enables clear attribution of influence to individual tokens in both text generation and image editing tasks~\cite{habib2024exploring}.

In the \textbf{text generation} case, the perturbed prompts are represented as feature vectors \( z_j \), typically one-hot encoded or binary indicators of word inclusion or exclusion. The response variable is the output-level semantic shift \(\Delta(x, \hat{x}_j)\), defined in Equation~\ref{eq:output-shift-ch3}. The similarity weights \( w_j \), derived from the Gaussian kernel (Eq.~\ref{eq:weights-ch3}), assign higher importance to perturbations more similar to the original prompt.

For \textbf{instruction-based image editing}, the same procedure applies. The independent variables are again the perturbed prompt vectors \( z_j \), while the response variable \(\Delta(x, \hat{x}_j)\) is computed as the Wasserstein distance between the baseline image and the perturbed image in the DINOv2 embedding space~\cite{oquab2023dinov2}. The weighting ensures that semantically faithful perturbations contribute more strongly to the regression outcome~\cite{molnar2020interpretable}.

The objective function for the weighted linear surrogate model is:

\begin{equation}
\label{eq:loss-ch3}
\min_{\theta} \frac{1}{J} \sum_{j=1}^{J} w_j \, \big( h_\theta(z_j) - \Delta(x, \hat{x}_j) \big)^2,
\end{equation}

After feature extraction, the surrogate model \( h_\theta \) is trained over the space of perturbed prompt vectors and their corresponding output shifts. The learned coefficients \( \theta \) quantify the influence of each token in the prompt on the model’s behaviour, providing intuitive and interpretable explanations.

\subsection{Evaluation Metrics}

To address \textbf{Research Question 1 (RQ1)}, we adopt a suite of evaluation metrics inspired by foundational work provided by Google~\cite{sanchez2020evaluating}, emphasising the multifaceted nature of assessing explainable models. This framework emphasises the importance of evaluating \textbf{Accuracy}, \textbf{Stability}, \textbf{Fidelity}, and \textbf{Consistency} as essential tools for a comprehensive assessment of model behaviour, particularly when comparing explainable models to traditional black-box models.

\textbf{Accuracy} refers to the correctness of the model outputs relative to the expected results, where applicable. \textbf{Stability} measures the robustness of the interpretability framework across different perturbations and model runs. \textbf{Fidelity} assesses how well the surrogate interpretable model approximates the original black-box model’s behaviour. \textbf{Consistency} evaluates the alignment of explanations across similar inputs, ensuring the interpretability results are coherent and reliable.

By adopting these metrics, we provide a structured methodology to rigorously dissect and understand the reliability and performance of our unified interpretability framework for both text generation and image editing models~\cite{sanchez2020evaluating}, thereby directly supporting the investigation and resolution of RQ1.

\begin{figure}[H]
    \centering
    \includegraphics[width=1\linewidth]{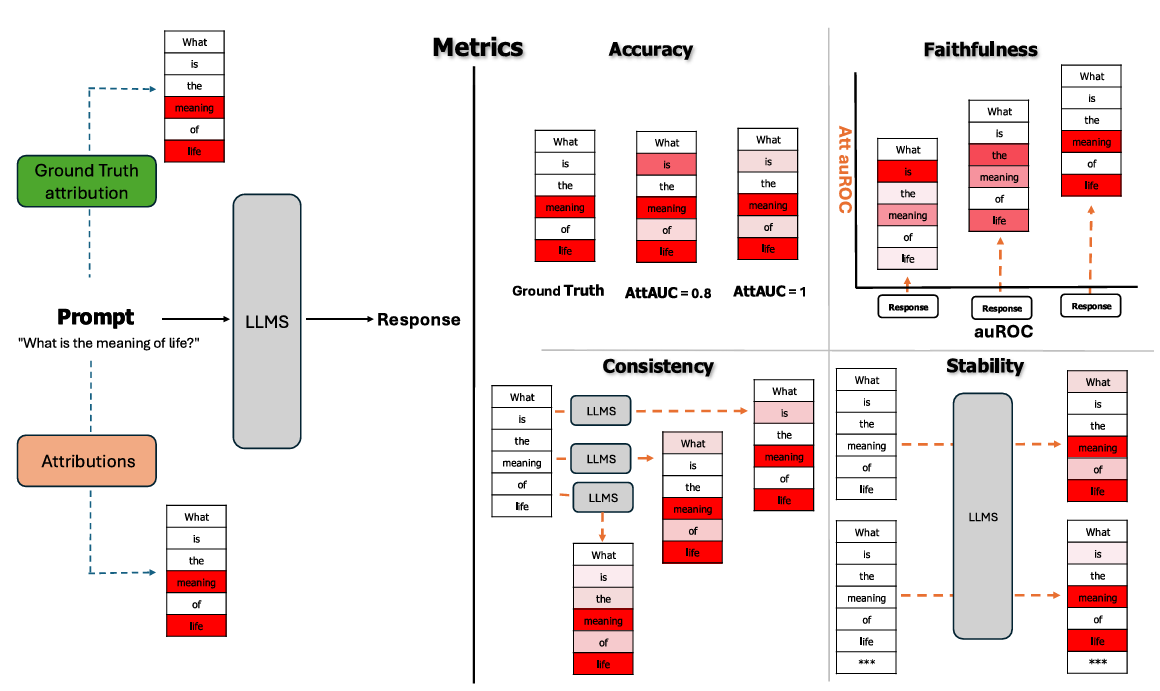}
    \caption{Evaluation framework for interpretability in LLMs using attribution-based metrics. The figure illustrates how the system compares ground-truth attributions with model-generated attributions to compute Accuracy (AUC/AUMAC), Faithfulness (via aUROC which compares input-output attribution alignment), Consistency (agreement across model outputs), and Stability (robustness of attributions to perturbations). This approach enables a comprehensive assessment of interpretability methods in natural language generation tasks.}
    \label{fig:LIMEDEV}
\end{figure}
\subsubsection{ATT Accuracy}

Accuracy measures how well the model's output aligns with the expected results (ground truth). In the context of our interpretability framework, this involves comparing the attribution scores, which reflect the influence of specific prompt words, against ground-truth labels that identify which prompt elements are genuinely relevant to the generated outputs~\cite{fong2017interpretable}.

For \textbf{text generation}, attribution scores are compared to human-annotated ground truth, identifying important words contributing to the generated content. For \textbf{image editing}, the model’s attributions are compared to labels, identifying which prompt words correspond to specific visual changes or regions in the edited images.

To quantify accuracy, we use the \textbf{Area Under the Curve (AUC)} of the Receiver Operating Characteristic (ROC)~\cite{hanley1982meaning}. The AUC reflects the model's ability to rank relevant prompt elements (as determined by ground truth) higher than irrelevant ones:
\begin{itemize}
    \item \textbf{AUC $\sim$ 1:} The model effectively distinguishes relevant elements, closely matching the human-identified ground truth.
    \item \textbf{AUC $\sim$ 0.5:} The model's ranking is random, indicating poor alignment with the ground truth.
\end{itemize}

\begin{figure}[H]
    \centering
    \begin{tikzpicture}
    \definecolor{whitebox}{rgb}{1, 1, 1}
    \definecolor{lightpink}{rgb}{1, 0.8, 0.8}
    \definecolor{mediumred}{rgb}{1, 0.6, 0.6}
    \definecolor{red}{rgb}{1, 0.4, 0.4}
    \definecolor{darkred}{rgb}{1, 0.2, 0.2}

    \node[align=center, font=\bfseries] at (-2, 5) {Attribution Accuracy Test};
    \node[align=center, font=\bfseries] at (-5, 4.5) {Ground Truth};

    \node[draw, rounded corners, fill=whitebox, minimum width=1.3cm] (t1) at (-5.5, 4) {could};
    \node[draw, rounded corners, fill=whitebox, minimum width=1.3cm] (t2) at (-4.2, 4) {you};
    \node[draw, rounded corners, fill=whitebox, minimum width=1.3cm] (t3) at (-2.9, 4) {please};
    \node[draw, rounded corners, fill=darkred, minimum width=1.3cm] (t4) at (-1.6, 4) {make};
    \node[draw, rounded corners, fill=whitebox, minimum width=1.3cm] (t5) at (-0.3, 4) {this};
    \node[draw, rounded corners, fill=darkred, minimum width=1.3cm] (t6) at (1, 4) {rainy};

    \node[draw, rounded corners, fill=blue!20, minimum width=2.5cm, minimum height=0.5cm] (pix2pix1) at (-2.3, 2.5) {Generative Model};

    \node[align=center, font=\bfseries] at (-5, 2) {Model 1: AttAUC = 0.8};

    \node[draw, rounded corners, fill=lightpink, minimum width=1.3cm] (w1) at (-5.5, 1.5) {could};
    \node[draw, rounded corners, fill=darkred, minimum width=1.3cm] (w2) at (-4.2, 1.5) {you};
    \node[draw, rounded corners, fill=whitebox, minimum width=1.3cm] (w3) at (-2.9, 1.5) {please};
    \node[draw, rounded corners, fill=darkred, minimum width=1.3cm] (w4) at (-1.6, 1.5) {make};
    \node[draw, rounded corners, fill=red, minimum width=1.3cm] (w5) at (-0.3, 1.5) {this};
    \node[draw, rounded corners, fill=darkred, minimum width=1.3cm] (w6) at (1, 1.5) {rainy};

    \node[align=center, font=\bfseries] at (-5, 0.3) {Model 2: AttAUC = 1};

    \node[draw, rounded corners, fill=lightpink, minimum width=1.3cm] (j1) at (-5.5, -0.2) {could};
    \node[draw, rounded corners, fill=lightpink, minimum width=1.3cm] (j2) at (-4.2, -0.2) {you};
    \node[draw, rounded corners, fill=whitebox, minimum width=1.3cm] (j3) at (-2.9, -0.2) {please};
    \node[draw, rounded corners, fill=darkred, minimum width=1.3cm] (j4) at (-1.6, -0.2) {make};
    \node[draw, rounded corners, fill=mediumred, minimum width=1.3cm] (j5) at (-0.3, -0.2) {this};
    \node[draw, rounded corners, fill=darkred, minimum width=1.3cm] (j6) at (1, -0.2) {rainy};

    \node at (-5.5, 3.5) {0};
    \node at (-4.2, 3.5) {0};
    \node at (-2.9, 3.5) {0};
    \node at (-1.6, 3.5) {1};
    \node at (-0.3, 3.5) {0};
    \node at (1, 3.5) {1};

    \node at (-5.5, 1) {0.10};
    \node at (-4.2, 1) {0.80};
    \node at (-2.9, 1) {0.01};
    \node at (-1.6, 1) {0.70};
    \node at (-0.3, 1) {0.50};
    \node at (1, 1) {0.90};

    \node at (-5.5, -0.7) {0.10};
    \node at (-4.2, -0.7) {0.10};
    \node at (-2.9, -0.7) {0.01};
    \node at (-1.6, -0.7) {0.70};
    \node at (-0.3, -0.7) {0.20};
    \node at (1, -0.7) {0.90};

    \end{tikzpicture}
    \caption{AttAUC (Attention Area Under Curve) metric evaluation of model accuracy for the image editing model. Darker red shades denote stronger attribution of text elements to image features, indicating the influence of specific prompt words on the Editing images.}

    \label{fig:accuracy-evaluation}
\end{figure}
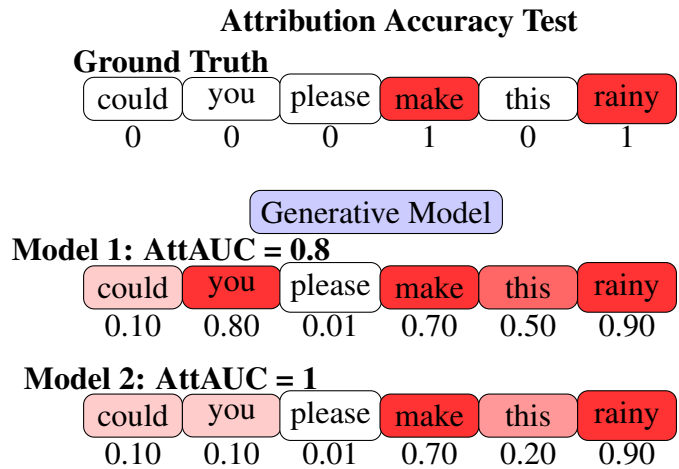

For example, Fig.~\ref{fig:accuracy-evaluation} illustrates a case where the ground truth identifies the words ``make'' and ``rainy'' as the most relevant text elements influencing the generated output. The metric \textbf{AttAUC} evaluates the alignment between the model’s attribution scores and the ground truth, where:
\begin{itemize}
    \item \textbf{AttAUC = 1.0:} Indicates perfect alignment.
    \item \textbf{Lower AttAUC (e.g., 0.8):} Reflects reduced accuracy~\cite{powers2020evaluation}.
\end{itemize}

The attribution heatmaps produced by the model are shown in Fig.~\ref{fig:accuracy-evaluation}. In these heatmaps, darker red shades indicate higher attribution to specific text elements. An accurate model should assign greater attention to the text components that most significantly influence the generated output, whether textual or visual. This enhances the interpretability and reliability of the model’s predictions~\cite{fong2017interpretable}.

Such evaluations assist stakeholders in making informed decisions about model deployment and guide fine-tuning efforts to improve interpretability performance.

\vspace{1em}

\textbf{Attention Accuracy (ATT ACC)}, \textbf{Attention F1 Score (ATT F1)}, and \textbf{Attention Area Under the Curve (ATT AUROC)} are defined as follows:

\begin{equation}\label{eq:att_acc}
\text{ATT ACC} = \frac{\text{Number of correctly attributed elements}}{\text{Total number of elements}} = \frac{\sum_{i=1}^N \mathbb{1}[\hat{y}_i = y_i]}{N}
\end{equation}

\textbf{Description:} ATT ACC measures the proportion of correctly attributed elements, where:
\begin{itemize}
    \item $N$ is the total number of elements,
    \item $\hat{y}_i$ is the predicted attribution score for the $i$-th element,
    \item $y_i$ is the corresponding ground truth label,
    \item $\mathbb{1}[\hat{y}_i = y_i]$ is an indicator function returning 1 if $\hat{y}_i = y_i$, and 0 otherwise.
\end{itemize}

\begin{equation}\label{eq:att_f1}
\text{ATT F1} = 2 \cdot \frac{\text{Precision} \cdot \text{Recall}}{\text{Precision} + \text{Recall}}
\end{equation}

\textbf{Description:} ATT F1 provides a harmonic mean of precision and recall, balancing false positives and false negatives. A higher F1 score indicates better overall model performance regarding accurate and relevant attributions. Precision and recall are defined as:

\begin{itemize}
    \item $\text{Precision} = \frac{\text{True Positives (TP)}}{\text{True Positives (TP)} + \text{False Positives (FP)}}$,
    \item $\text{Recall} = \frac{\text{True Positives (TP)}}{\text{True Positives (TP)} + \text{False Negatives (FN)}}$.
\end{itemize}

\vspace{1em}

\begin{equation}\label{eq:att_auroc}
\text{ATT AUROC} = \frac{1}{|P| \cdot |N|} \sum_{p \in P} \sum_{n \in N} \mathbb{1}[\hat{y}_p > \hat{y}_n]
\end{equation}

\textbf{Description:} ATT AUROC evaluates the model's ability to rank relevant elements (positive ground truth) higher than irrelevant ones (negative ground truth). It ranges from 0.5 (random ranking) to 1.0 (perfect ranking). Here:

\begin{itemize}
    \item $P$ and $N$ are the sets of positive and negative ground truth labels, respectively,
    \item $|P|$ and $|N|$ are the sizes of these sets,
    \item $\hat{y}_p$ and $\hat{y}_n$ are the model's predicted scores for positive and negative samples,
    \item $\mathbb{1}[\hat{y}_p > \hat{y}_n]$ is an indicator function returning 1 if the positive sample's score is higher than the negative sample's score, and 0 otherwise.
\end{itemize}

\vspace{1em}

These metrics ensure the model is rigorously tested for accuracy, consistency, and relevance, offering a robust assessment of its alignment with human-annotated ground truth labels.

\subsubsection{ATT Stability}

Stability in an explainable model refers to its ability to provide consistent explanations even when minor changes are made to the input data~\cite{samek2017explainable}. This means that small perturbations in the input should not significantly affect the model's predictions or explanations. Stability is a key property for fostering trust in explainable artificial intelligence (XAI) systems~\cite{doshi2017towards}. It reassures users that the model's interpretations are not arbitrary or overly sensitive to irrelevant variations~\cite{lipton2018mythos}.

For \textbf{text generation}, stability assesses whether small changes to the input prompt (e.g., adding or removing non-critical words or symbols) result in similar attribution patterns over the prompt elements. For these experiments, minor text variations were introduced by adding the symbol \textbf{\texttt{***}} at the end of the prompt. For \textbf{image editing}, stability evaluates whether small prompt changes lead to consistent attribution of text elements to visual features across different edited images, where the variation was introduced by adding the symbol \textbf{\texttt{\#\#\#}} at the end of the prompt.

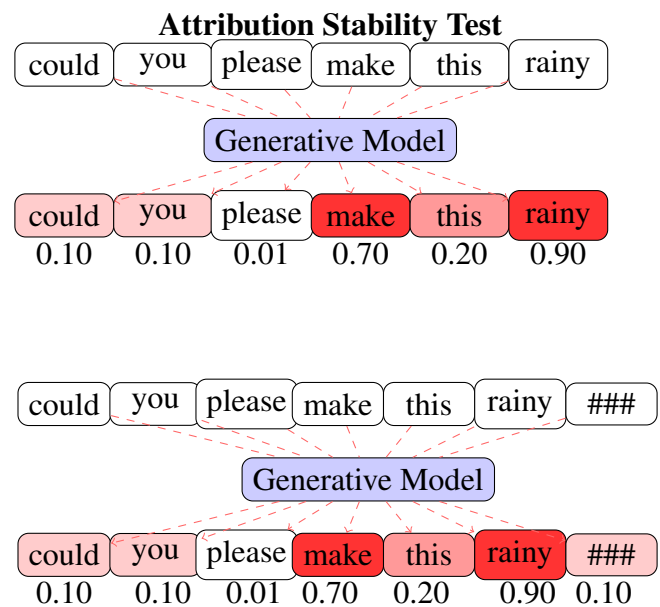
\begin{figure}[H]
    \centering
    \begin{tikzpicture}
    \definecolor{whitebox}{rgb}{1, 1, 1}
    \definecolor{lightpink}{rgb}{1, 0.8, 0.8}
    \definecolor{mediumred}{rgb}{1, 0.6, 0.6}
    \definecolor{red}{rgb}{1, 0.4, 0.4}
    \definecolor{darkred}{rgb}{1, 0.2, 0.2}

    \node[align=center, font=\bfseries] at (-2, 4.5) {Attribution Stability Test};

    \node[draw, rounded corners, fill=whitebox, minimum width=1.3cm] (w1) at (-5.5, 4) {could};
    \node[draw, rounded corners, fill=whitebox, minimum width=1.3cm] (w2) at (-4.2, 4) {you};
    \node[draw, rounded corners, fill=whitebox, minimum width=1.3cm] (w3) at (-2.9, 4) {please};
    \node[draw, rounded corners, fill=whitebox, minimum width=1.3cm] (w4) at (-1.6, 4) {make};
    \node[draw, rounded corners, fill=whitebox, minimum width=1.3cm] (w5) at (-0.3, 4) {this};
    \node[draw, rounded corners, fill=whitebox, minimum width=1.3cm] (w6) at (1, 4) {rainy};

    \node[draw, rounded corners, fill=blue!20, minimum width=2.5cm, minimum height=0.5cm] (pix2pix1) at (-2, 3) {Generative Model};

    \node[draw, rounded corners, fill=lightpink, minimum width=1.3cm] (j1) at (-5.5, 2) {could};
    \node[draw, rounded corners, fill=lightpink, minimum width=1.3cm] (j2) at (-4.2, 2) {you};
    \node[draw, rounded corners, fill=whitebox, minimum width=1.3cm] (j3) at (-2.9, 2) {please};
    \node[draw, rounded corners, fill=darkred, minimum width=1.3cm] (j4) at (-1.6, 2) {make};
    \node[draw, rounded corners, fill=mediumred, minimum width=1.3cm] (j5) at (-0.3, 2) {this};
    \node[draw, rounded corners, fill=darkred, minimum width=1.3cm] (j6) at (1, 2) {rainy};

    \node[draw, rounded corners, fill=whitebox, minimum width=1.2cm] (t1) at (-5.5, -0.5) {could};
    \node[draw, rounded corners, fill=whitebox, minimum width=1.2cm] (t2) at (-4.3, -0.5) {you};
    \node[draw, rounded corners, fill=whitebox, minimum width=1.2cm] (t3) at (-3.1, -0.5) {please};
    \node[draw, rounded corners, fill=whitebox, minimum width=1.2cm] (t4) at (-1.9, -0.5) {make};
    \node[draw, rounded corners, fill=whitebox, minimum width=1.2cm] (t5) at (-0.7, -0.5) {this};
    \node[draw, rounded corners, fill=whitebox, minimum width=1.2cm] (t6) at (0.5, -0.5) {rainy};
    \node[draw, rounded corners, fill=whitebox, minimum width=1.2cm] (t7) at (1.7, -0.5) {\#\#\#};

    \node[draw, rounded corners, fill=blue!20, minimum width=2.5cm, minimum height=0.5cm] (pix2pix) at (-1.5, -1.5) {Generative Model};

    \node[draw, rounded corners, fill=lightpink, minimum width=1.2cm] (b1) at (-5.5, -2.5) {could};
    \node[draw, rounded corners, fill=lightpink, minimum width=1.2cm] (b2) at (-4.3, -2.5) {you};
    \node[draw, rounded corners, fill=whitebox, minimum width=1.2cm] (b3) at (-3.1, -2.5) {please};
    \node[draw, rounded corners, fill=darkred, minimum width=1.2cm] (b4) at (-1.9, -2.5) {make};
    \node[draw, rounded corners, fill=mediumred, minimum width=1.2cm] (b5) at (-0.7, -2.5) {this};
    \node[draw, rounded corners, fill=darkred, minimum width=1.2cm] (b6) at (0.5, -2.5) {rainy};
    \node[draw, rounded corners, fill=lightpink, minimum width=1.2cm] (b7) at (1.7, -2.5) {\#\#\#};

    \node at (-5.5, 1.5) {0.10};
    \node at (-4.2, 1.5) {0.10};
    \node at (-2.9, 1.5) {0.01};
    \node at (-1.6, 1.5) {0.70};
    \node at (-0.3, 1.5) {0.20};
    \node at (1, 1.5) {0.90};

    \node at (-5.5, -3) {0.10};
    \node at (-4.2, -3) {0.10};
    \node at (-3, -3) {0.01};
    \node at (-2, -3) {0.70};
    \node at (-0.8, -3) {0.20};
    \node at (0.6, -3) {0.90};
    \node at (1.6, -3) {0.10};

    \draw[-, dashed, red] (w1) -- (pix2pix1);
    \draw[-, dashed, red] (w2) -- (pix2pix1);
    \draw[-, dashed, red] (w3) -- (pix2pix1);
    \draw[-, dashed, red] (w4) -- (pix2pix1);
    \draw[-, dashed, red] (w5) -- (pix2pix1);
    \draw[-, dashed, red] (w6) -- (pix2pix1);

    \draw[->, dashed, red] (pix2pix1) -- (j1);
    \draw[->, dashed, red] (pix2pix1) -- (j2);
    \draw[->, dashed, red] (pix2pix1) -- (j3);
    \draw[->, dashed, red] (pix2pix1) -- (j4);
    \draw[->, dashed, red] (pix2pix1) -- (j5);
    \draw[->, dashed, red] (pix2pix1) -- (j6);

    \draw[-, dashed, red] (t1) -- (pix2pix);
    \draw[-, dashed, red] (t2) -- (pix2pix);
    \draw[-, dashed, red] (t3) -- (pix2pix);
    \draw[-, dashed, red] (t4) -- (pix2pix);
    \draw[-, dashed, red] (t5) -- (pix2pix);
    \draw[-, dashed, red] (t6) -- (pix2pix);
    \draw[-, dashed, red] (t7) -- (pix2pix);

    \draw[->, dashed, red] (pix2pix) -- (b1);
    \draw[->, dashed, red] (pix2pix) -- (b2);
    \draw[->, dashed, red] (pix2pix) -- (b3);
    \draw[->, dashed, red] (pix2pix) -- (b4);
    \draw[->, dashed, red] (pix2pix) -- (b5);
    \draw[->, dashed, red] (pix2pix) -- (b6);
    \draw[->, dashed, red] (pix2pix) -- (b7);

    \end{tikzpicture}
    \caption{Stability evaluation of the image editing model’s attention with slight text variations. Darker red shades indicate consistent attribution of text elements across perturbed prompts.}

    \label{fig:stability-evaluation}
\end{figure}

The example visualizations present the attention scores for two slightly different text inputs. The heatmaps demonstrate how the model assigns attention to each word. A stable model, as shown in Fig.~\ref{fig:stability-evaluation}, assigns similar importance to the words \textit{make} and \textit{rainy} in both inputs, even though the additional symbol (\texttt{***} for text generation or \texttt{\#\#\#} for image editing) was introduced in one case. This consistency is critical for ensuring the model behaves predictably across minor input variations~\cite{ahmadi2024explainability}.

Without such stability, the explanations generated by the model could lead to confusion or misinterpretation, particularly in high-stakes scenarios such as medical diagnosis or financial decision-making~\cite{ribeiro2016should}.

We use the \textbf{Jaccard index} to quantify stability, which measures the similarity between two sets of model attributions. For two sets \( A \) and \( B \), the Jaccard index is calculated as:

\begin{equation}
\label{eq:jaccard}
\text{Jaccard}(A, B) = \frac{|A \cap B|}{|A \cup B|}
\end{equation}

This metric reflects the proportion of shared elements between the two attribution sets, where values closer to 1 indicate greater stability and similarity in the explanations~\cite{samek2017explainable}. A higher Jaccard index confirms the model’s ability to provide stable outputs and reliable interpretations~\cite{vaswani2017attention}.

By ensuring stability, the model enhances not only robustness but also usability and credibility in real-world applications~\cite{doshi2017towards}.

\subsubsection{ATT Consistency}

Consistency evaluates whether the model produces stable and reliable outputs when the same input is repeatedly provided. This property ensures that the model behaves predictably, reinforcing trust in its outputs and usability across diverse scenarios~\cite{ribeiro2016should}.

For \textbf{text generation}, consistency assesses whether repeated executions with the exact prompt yield similar attribution patterns and generated content. For \textbf{image editing}, it evaluates whether multiple runs using the same input prompt and image consistently highlight the same prompt elements and produce similar visual edits.

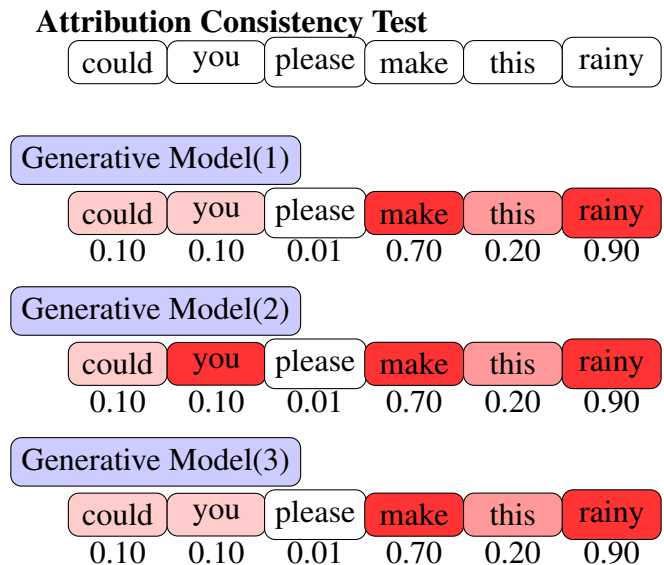
\begin{figure}[H]
    \centering
    \begin{tikzpicture}
    \definecolor{whitebox}{rgb}{1, 1, 1}
    \definecolor{lightpink}{rgb}{1, 0.8, 0.8}
    \definecolor{mediumred}{rgb}{1, 0.6, 0.6}
    \definecolor{red}{rgb}{1, 0.4, 0.4}
    \definecolor{darkred}{rgb}{1, 0.2, 0.2}

    \node[align=center, font=\bfseries] at (-4., 6) {Attribution Consistency Test};

    \node[draw, rounded corners, fill=whitebox, minimum width=1.3cm] (b1) at (-5.5, 5.5) {could};
    \node[draw, rounded corners, fill=whitebox, minimum width=1.3cm] (b2) at (-4.2, 5.5) {you};
    \node[draw, rounded corners, fill=whitebox, minimum width=1.3cm] (b3) at (-2.9, 5.5) {please};
    \node[draw, rounded corners, fill=whitebox, minimum width=1.3cm] (b4) at (-1.6, 5.5) {make};
    \node[draw, rounded corners, fill=whitebox, minimum width=1.3cm] (b5) at (-0.3, 5.5) {this};
    \node[draw, rounded corners, fill=whitebox, minimum width=1.3cm] (b6) at (1, 5.5) {rainy};

    \node[draw, rounded corners, fill=blue!20, minimum width=1.5cm, minimum height=0.5cm] (pix2pix1) at (-5, 4.2) {Generative Model(1)};

    \node[draw, rounded corners, fill=lightpink, minimum width=1.3cm] (t1) at (-5.5, 3.5) {could};
    \node[draw, rounded corners, fill=lightpink, minimum width=1.3cm] (t2) at (-4.2, 3.5) {you};
    \node[draw, rounded corners, fill=whitebox, minimum width=1.3cm] (t3) at (-2.9, 3.5) {please};
    \node[draw, rounded corners, fill=darkred, minimum width=1.3cm] (t4) at (-1.6, 3.5) {make};
    \node[draw, rounded corners, fill=mediumred, minimum width=1.3cm] (t5) at (-0.3, 3.5) {this};
    \node[draw, rounded corners, fill=darkred, minimum width=1.3cm] (t6) at (1, 3.5) {rainy};

    \node[draw, rounded corners, fill=blue!20, minimum width=1.5cm, minimum height=0.5cm] (pix2pix2) at (-5, 2.2) {Generative Model(2)};

    \node[draw, rounded corners, fill=lightpink, minimum width=1.3cm] (w1) at (-5.5, 1.5) {could};
    \node[draw, rounded corners, fill=darkred, minimum width=1.3cm] (w2) at (-4.2, 1.5) {you};
    \node[draw, rounded corners, fill=whitebox, minimum width=1.3cm] (w3) at (-2.9, 1.5) {please};
    \node[draw, rounded corners, fill=darkred, minimum width=1.3cm] (w4) at (-1.6, 1.5) {make};
    \node[draw, rounded corners, fill=mediumred, minimum width=1.3cm] (w5) at (-0.3, 1.5) {this};
    \node[draw, rounded corners, fill=darkred, minimum width=1.3cm] (w6) at (1, 1.5) {rainy};

    \node[draw, rounded corners, fill=blue!20, minimum width=1.5cm, minimum height=0.5cm] (pix2pix3) at (-5, 0.2) {Generative Model(3)};

    \node[draw, rounded corners, fill=lightpink, minimum width=1.3cm] (j1) at (-5.5, -0.5) {could};
    \node[draw, rounded corners, fill=lightpink, minimum width=1.3cm] (j2) at (-4.2, -0.5) {you};
    \node[draw, rounded corners, fill=whitebox, minimum width=1.3cm] (j3) at (-2.9, -0.5) {please};
    \node[draw, rounded corners, fill=darkred, minimum width=1.3cm] (j4) at (-1.6, -0.5) {make};
    \node[draw, rounded corners, fill=mediumred, minimum width=1.3cm] (j5) at (-0.3, -0.5) {this};
    \node[draw, rounded corners, fill=darkred, minimum width=1.3cm] (j6) at (1, -0.5) {rainy};

    \node at (-5.5, 3) {0.10};
    \node at (-4.2, 3) {0.10};
    \node at (-2.9, 3) {0.01};
    \node at (-1.6, 3) {0.70};
    \node at (-0.3, 3) {0.20};
    \node at (1, 3) {0.90};

    \node at (-5.5, 1) {0.10};
    \node at (-4.2, 1) {0.10};
    \node at (-2.9, 1) {0.01};
    \node at (-1.6, 1) {0.70};
    \node at (-0.3, 1) {0.20};
    \node at (1, 1) {0.90};

    \node at (-5.5, -1) {0.10};
    \node at (-4.2, -1) {0.10};
    \node at (-2.9, -1) {0.01};
    \node at (-1.6, -1) {0.70};
    \node at (-0.3, -1) {0.20};
    \node at (1, -1) {0.90};

    \end{tikzpicture}
    \caption{Consistency analysis of the Instruct-Pix2Pix models. The heatmaps show the influence of each word on the model's output across three iterations.}

    \label{fig:pix2pix-consistency}
\end{figure}

As shown in Fig.~\ref{fig:pix2pix-consistency}, when the input phrase ``Could you please make this rainy?'' was provided to the model multiple times, the outputs remained stable across iterations. In both text and image tasks, the model consistently assigned higher weights to the words ``make'' and ``rainy'', demonstrating that it reliably identifies the most semantically essential elements of the input~\cite{vaswani2017attention}.

Such consistent behaviour underscores the model's ability to focus on the semantic core of the prompt, a trait essential for maintaining coherence and usability in practical applications~\cite{lipton2018mythos}. In fields such as automated content generation, weather simulation, and interactive systems, users rely on models to deliver outputs that are consistent and predictable. Any inconsistencies could undermine user trust and compromise the system’s overall effectiveness.

Our model enhances user confidence by demonstrating consistency across repeated runs and establishes a foundation for further optimisation and integration into more complex systems. This ability to maintain stable explanations and outputs directly contributes to the robustness and scalability of the model, making it well-suited for deployment in dynamic, real-world environments~\cite{doshi2017towards}.

\subsubsection{ATT Fidelity}

Fidelity evaluates how well the explainable surrogate model aligns with the original black-box model. This metric measures the degree to which the surrogate’s predictions approximate those of the black-box model under various input perturbations~\cite{ribeiro2016should}. 

Formally, let the ground-truth output shift from the black-box model be 
\(\Delta(x, \hat{x}_j)\) (see Eq.~\ref{eq:output-shift-ch3}), and let the surrogate’s predicted shift be 
\(\hat{\Delta}_j = h_\theta(z_j)\). Fidelity is quantified by comparing 
\(\hat{\Delta}_j\) against \(\Delta(x, \hat{x}_j)\) across all perturbations 
\(j \in \{1, \dots, J\}\).

\paragraph{Coefficient-based metrics.}  
The \textbf{coefficient of determination} \(R^2\) assesses how well the surrogate correlates with the black-box model~\cite{draper1998applied}:
\begin{equation}
\label{eq:coefficient_of_determination}
R^2 = 1 - \frac{\sum_{j=1}^{J} \left( \Delta(x,\hat{x}_j) - \hat{\Delta}_j \right)^2}
                  {\sum_{j=1}^{J} \left( \Delta(x,\hat{x}_j) - \overline{\Delta} \right)^2},
\end{equation}
where \(\overline{\Delta}\) is the mean of the black-box distances. A value close to 1 indicates a strong alignment between the surrogate and the black-box.

When perturbations are weighted, the \textbf{weighted coefficient of determination} \(R_w^2\) is used~\cite{ahmadi2024explainability, biometrics19763211hocking1976analysis}:
\begin{equation}
\label{eq:weighted_coefficient_of_determination}
R_w^2 = 1 - \frac{\sum_{j=1}^{J} w_j \left( \Delta(x,\hat{x}_j) - \hat{\Delta}_j \right)^2}
                  {\sum_{j=1}^{J} w_j \left( \Delta(x,\hat{x}_j) - \overline{\Delta}_w \right)^2},
\end{equation}
where \(\overline{\Delta}_w = \frac{\sum_{j=1}^J w_j \Delta(x,\hat{x}_j)}{\sum_{j=1}^J w_j}\) is the weighted mean.

To account for the number of features \(N_s\) and perturbations \(J\), the \textbf{weighted adjusted coefficient of determination} is defined as~\cite{ahmadi2024explainability}:
\begin{equation}
\label{eq:weighted_adjusted_coefficient_of_determination}
\hat{R}_w^2 = 1 - (1 - R_w^2) \left[ \frac{J - 1}{J - N_s - 1} \right].
\end{equation}

\paragraph{Error-based metrics.}  
Error magnitude between surrogate and black-box predictions is measured using the 
\textbf{Weighted Mean Squared Error (WMSE)} and \textbf{Weighted Mean Absolute Error (WMAE)}~\cite{willmott2005advantages}:
\begin{equation}
\label{eq:mse}
\text{WMSE} = \frac{\sum_{j=1}^{J} w_j \big( \Delta(x,\hat{x}_j) - \hat{\Delta}_j \big)^2}{\sum_{j=1}^{J} w_j},
\end{equation}
\begin{equation}
\label{eq:mae}
\text{WMAE} = \frac{\sum_{j=1}^{J} w_j \, \big| \Delta(x,\hat{x}_j) - \hat{\Delta}_j \big|}{\sum_{j=1}^{J} w_j}.
\end{equation}

Additionally, unweighted \textbf{mean \(L_1\)} and \textbf{mean \(L_2\)} losses are computed~\cite{goodfellow2016deep, ahmadi2024explainability}:
\begin{equation}
\label{eq:l1_loss}
L_1 = \frac{1}{J} \sum_{j=1}^{J} \big| \Delta(x,\hat{x}_j) - \hat{\Delta}_j \big|,
\end{equation}
\begin{equation}
\label{eq:l2_loss}
L_2 = \frac{1}{J} \sum_{j=1}^{J} \big( \Delta(x,\hat{x}_j) - \hat{\Delta}_j \big)^2.
\end{equation}

\paragraph{Interpretation.}  
Lower values of WMSE, WMAE, \(L_1\), and \(L_2\), together with higher values of \(R^2\), \(R_w^2\), and \(\hat{R}_w^2\), indicate better fidelity, meaning the surrogate closely approximates the behaviour of the black-box model.

\bigskip

We compute fidelity across a range of scenarios. For \textbf{text generation}, fidelity is evaluated by comparing the explainable model's predicted distances to those of the black-box language model when prompted with perturbed and original texts. For \textbf{image editing}, the fidelity is computed using distances based on DINOv2 image embeddings compared between the surrogate and the original image editing model.

\begin{figure}[H]
    \centering
    \includegraphics[width=1\linewidth]{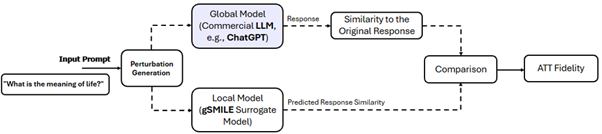}
    \caption{ATT fidelity evaluation framework. The input prompt is perturbed, 
    and both a Global Model (e.g., ChatGPT for text generation, 
    InstructPix2Pix for image editing) and a Local Surrogate Model (gSMILE) 
    generate similarity signals. Their comparison quantifies how closely 
    The surrogate approximates the black-box model’s behaviour.}
    \label{fig:fidelity}
\end{figure}

The overall workflow for computing ATT fidelity is shown in Fig~\ref{fig:fidelity}. 
As depicted, input prompts are perturbed, and both the Global Model and the Local Surrogate Model 
produce similarity signals that are subsequently compared to evaluate ATT fidelity. 
For \textbf{text generation}, similarity is computed using Outcome Word Mover’s Distance (OWMD). 
For \textbf{instruction-based image editing}, similarity is calculated using Wasserstein distance 
over DINOv2 embeddings.

\section*{Conclusion}
\addcontentsline{toc}{section}{Discussion and Conclusion}

In this chapter, we introduced the gSMILE framework as a unified method for explaining both large language models (LLMs) and instruction-based image editing systems. By extending LIME with statistical distance measures and model-agnostic techniques, gSMILE provides a flexible and consistent approach to understanding how input changes affect model outputs, whether in text or image generation tasks.

We walked through the complete gSMILE framework in this chapter, from generating minor variations in inputs to building simplified models that help explain how the original models behave. We also discussed how we evaluate the quality of these explanations using metrics such as stability, fidelity, and consistency. What makes gSMILE stand out is that it does not require access to what is going on inside the model, which means it can be used even with closed-source or API-only systems.

This chapter sets the stage for what follows, where we put gSMILE to the test on real-world generative models, both in language and image tasks, to assess its actual performance.

\chapter{Explainability of Large Language Models}
\label{chap:llm_explainability}

\begin{figure}[H]
    \centering
    \includegraphics[width=0.95\textwidth]{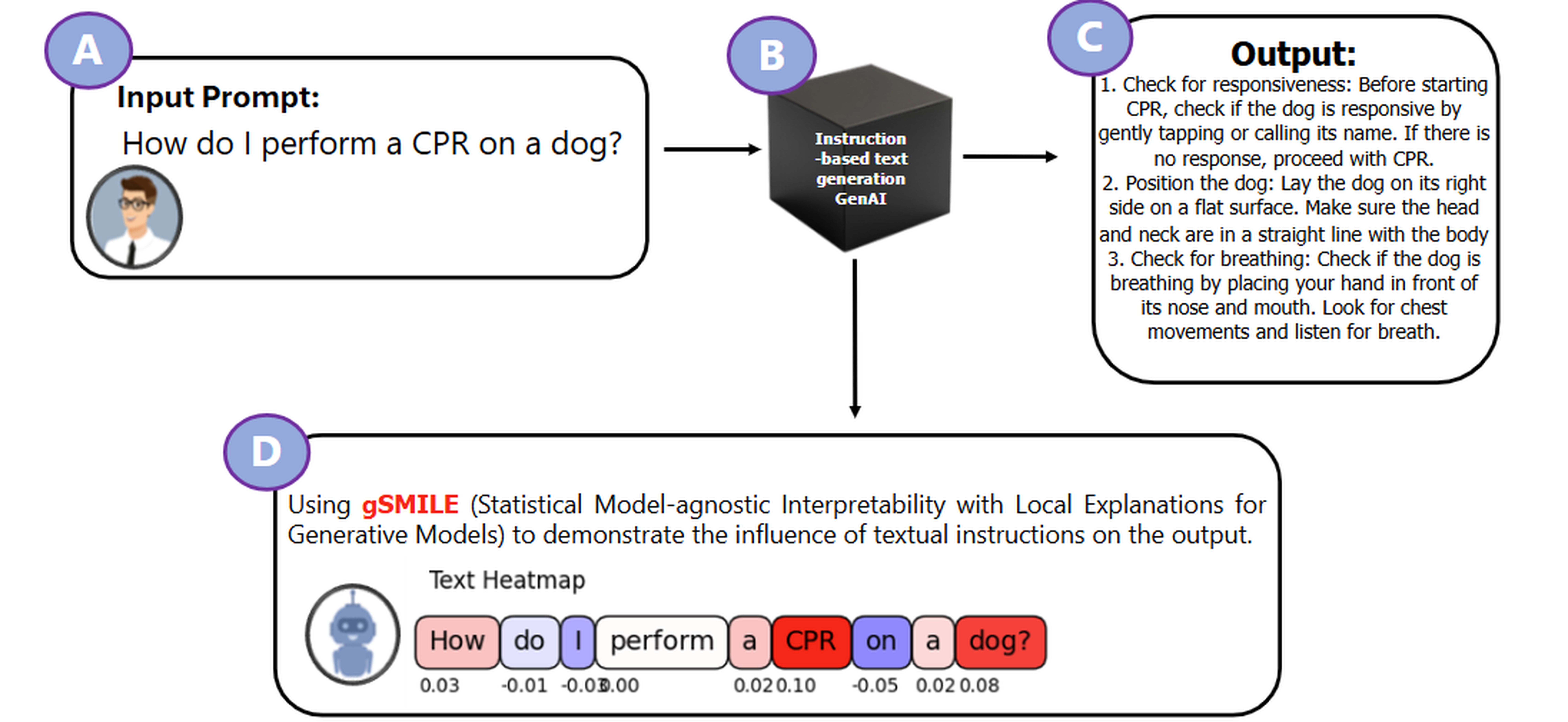}
    \caption{An overview of the explainability process using gSMILE (Statistical Model-agnostic Interpretability with Local Explanations) applied to a Large Language Model (LLM).}
    \label{fig:smile_llm_overview}
\end{figure}

This chapter directly addresses \textbf{Research Question 2 (RQ2)}, which investigates how interpretability methods can reveal the influence of individual input elements in large language models (LLMs). It introduces the challenges and approaches for interpreting the outputs of LLMs, with a focus on model-agnostic methods. This study presents the gSMILE framework, developed to provide local explanations for LLM predictions. An overview of the explainability workflow applied in this research is shown in Fig.~\ref{fig:smile_llm_overview}. We tested gSMILE on several leading large language models (LLMs) and used metrics such as accuracy, consistency, stability, and fidelity to demonstrate that it provides clear and reliable explanations. By making these models easier to understand, gSMILE brings us one step closer to making AI more transparent and trustworthy.

\section{Problem Definition}\label{sec:problem_definition}

This section introduces the core concepts and mathematical framework underlying our approach. We begin by defining the key notation and problem setup, followed by a formal description of how we quantify the impact of input perturbations. We then describe the use of a surrogate model to approximate model behaviour locally and provide theoretical justification based on the Wasserstein distance~\cite{ribeiro2016should, arjovsky2017wasserstein, peyre2019computational}. Lastly, we outline a structured methodology for implementing a Wasserstein-based surrogate framework tailored to large language models~\cite{molnar2020interpretable}.

Word Mover’s Distance (WMD) is the 1-Wasserstein (Earth Mover’s) distance between two documents represented as normalised bags-of-words in an embedding space, using the Euclidean distance between word embeddings as the ground cost~\cite{kusner2015word}. WMD provides a semantically meaningful measure of dissimilarity between documents by capturing the minimum amount of semantic ``movement'' needed to align their word distributions, making it well-suited for our interpretability framework.

\subsection{Notation and Problem Setting}

\paragraph{Setting the Stage.}
We aim to construct a local surrogate model that explains the behaviour of a black-box large language model (LLM), denoted as \( \pi^{(n)} \), in response to a natural language input prompt. The goal is to understand how specific parts of the input affect the model's output distribution. The following notation is used:

\begin{itemize}
    \item \textbf{Input:} \( x \in \mathcal{X} \), representing the original input prompt.
    \item \textbf{Perturbations:} \( \{\hat{x}_j\}_{j=1}^{J} \), a set of \( J \) perturbed versions of \( x \), generated via minor token-level edits.
    \item \textbf{Model Outputs:} \( \pi^{(n)}(y \mid \hat{x}_j) \), the model’s probability distribution over output space \( \mathcal{Y} \), conditioned on the perturbed input \( \hat{x}_j \). For the unperturbed input \( x \), we write \( \pi^{(n)}(y \mid x) \).
\end{itemize}

\subsection{Input-Level Distance}

To quantify the semantic difference between the original prompt \( x \) and each perturbed version \( \hat{x}_j \), we compute the Word Mover’s Distance (WMD)~\cite{kusner2015word}. We denote this input-level semantic distance as IWMD (Inner WMD):

\begin{equation}
\delta_{x_j} = \mathrm{IWMD}(x, \hat{x}_j),
\label{eq:input-distance}
\end{equation}

Where \( \delta_{x_j} \in \mathbb{R}_{\geq 0} \) denotes the semantic distance between the original and perturbed prompts.

\subsection{Output-Level Distribution Shift}

To assess the effect of input perturbation on the model’s behaviour, we compute the Wasserstein distance between the output distributions of the original and perturbed inputs. We refer to this as OWMD (Outcome WMD):

\begin{equation}
\mathrm{OWMD}(x, \hat{x}_j) = W \left( \pi^{(n)}(y \mid x), \pi^{(n)}(y \mid \hat{x}_j) \right),
\label{eq:owmd-definition}
\end{equation}

\begin{equation}
\Delta(x, \hat{x}_j) = \mathrm{OWMD}(x, \hat{x}_j),
\label{eq:output-shift}
\end{equation}

where \( y \in \mathcal{Y} \), and \( W(p, q) \) denotes the Wasserstein distance between two probability distributions \( p \) and \( q \).

\subsection{Weighting via Input Similarity}

Perturbations that are semantically closer to the original input are weighted more heavily. We use a Gaussian-based weighting scheme~\cite{ribeiro2016should}:

\begin{equation}
w_j = \exp \left( - \left( \frac{ \delta_{x_j} }{ \sigma } \right)^2 \right),
\label{eq:weights}
\end{equation}

Where \( w_j \in (0,1] \) denotes the relevance weight of \( \hat{x}_j \), and \( \sigma > 0 \) is a kernel width parameter that controls the rate of exponential decay.

\subsection{Fitting a Local Surrogate}

Each perturbed input \( \hat{x}_j \) is mapped to a feature vector \( z_j \in \mathbb{R}^d \). We fit a linear surrogate model of the form:

\begin{equation}
h_\theta(z_j) = \theta_0 + \theta^\top z_j,
\label{eq:surrogate}
\end{equation}

where \( \theta_0 \in \mathbb{R} \) is a bias term and \( \theta \in \mathbb{R}^d \) is the weight vector. Parameters \( \theta \) are estimated by minimising the weighted squared error.

\begin{equation}
\min_{\theta} \sum_{j=1}^{J} w_j \left( h_\theta(z_j) - \Delta(x, \hat{x}_j) \right)^2,
\label{eq:loss}
\end{equation}

Where the loss ensures local ATT fidelity between the surrogate and the observed output shifts.

\subsection{Theoretical Justification (Lipschitz Smoothness)}

\paragraph{Shift Function Definition.}

We define a shift function that locally quantifies the model's response to perturbation:


\begin{equation}
f(x_a, \hat{x}_a) = W \left( \pi^{(n)}(y \mid x_a), \pi^{(n)}(y \mid \hat{x}_a) \right),
\label{eq:shift-function}
\end{equation}

If \( \pi^{(n)} \) is Lipschitz continuous with respect to its input, then the function \( f \) is also Lipschitz~\cite{beliakov2007aggregation}:

\begin{equation}
\left| f(x_a, \hat{x}_a) - f(x_b, \hat{x}_b) \right| \leq L \times \left( \|x_a - x_b\| + \|\hat{x}_a - \hat{x}_b\| \right), \forall a,b \in \{1, \dots, N_{pert}\}
\label{eq:lipschitz}
\end{equation}


The notation \( x_a \) and \( \hat{x}_a \) represents a specific input and its corresponding perturbation. The subscript allows us to extend the formulation to multiple input pairs \( (x_i, \hat{x}_i) \) when analysing the Lipschitz continuity or stability across perturbations. \( N_{pert} \) denotes the total number of perturbations. 


where \( L > 0 \) is the Lipschitz constant, and \( \|\cdot\| \) denotes a norm over the input space (e.g., token embedding space).

\paragraph{Relation to Total Variation.}

For two distributions \( p \) and \( q \) over output space \( \mathcal{Y} \), the Kantorovich–Rubinstein duality yields:

\begin{equation}
W(p, q) \leq \text{diam}(\mathcal{Y}) \cdot \mathrm{TV}(p, q),
\label{eq:wasserstein-tv}
\end{equation}

Where $\text{diam}(\mathcal{Y}) = \sup_{y_1, y_2 \in \mathcal{Y}} d(y_1, y_2)$ denotes the \emph{diameter} of the output space under the chosen ground metric $d$, i.e., the maximum possible distance between any two outputs.

The total variation distance ($TV$) is given by:

\begin{equation}
\mathrm{TV}(p, q) = \frac{1}{2} \sum_{y \in \mathcal{Y}} |p(y) - q(y)|.
\label{eq:tv}
\end{equation}

This bound highlights that small changes in distribution (in the TV sense) imply a bounded Wasserstein distance.

\subsection{Linear Surrogates Approximate Lipschitz Functions}

Due to the differentiability of Lipschitz functions, the shift function \( f \) can be locally approximated by a first-order Taylor expansion:

\begin{equation}
f(x') \approx f(x) + \nabla f(x)^\top (x' - x),
\label{eq:linear-approximation}
\end{equation}

where \( \nabla f(x)^\top (x' - x) \) denotes the dot product between the gradient of \( f \) and the local displacement in input space.

The complete Wasserstein-based LIME surrogate procedure consists of generating perturbations $\{\hat{x}_j\}_{j=1}^{J}$ near input prompt $x$. The process then computes semantic distances $\delta_{x_j}$ (Equation~\ref{eq:input-distance}) and output shifts $\Delta(x, \hat{x}_j)$ (Equation~\ref{eq:output-shift}). Following this, similarity weights $w_j$ are assigned via Equation~\ref{eq:weights}. The procedure continues by extracting feature representations $z_j$ for each perturbation and fitting a local surrogate model $h_\theta$ by minimising the loss (Equation~\ref{eq:loss}). Finally, the fitted model $h_\theta$ is used to interpret the influence of each input feature on model behaviour. This framework offers a principled and interpretable approach to analysing large language models through localised perturbation and response analysis.

\section{Method Summary for LLMs}

This section outlines how the general gSMILE framework, previously described in Chapter~\ref{chap:gSMILE Methodology}, is adapted to interpret the behaviour of large language models (LLMs) in text generation tasks. The objective is to identify which prompt components most strongly influence the model’s output.

We begin by decomposing the original input prompt into individual words. Perturbed prompts are generated by systematically removing or altering one word at a time, resulting in modified inputs. Each perturbed prompt is passed to the LLM, which produces a corresponding output. The text generated from the original prompt serves as the baseline for comparison.

To measure how the outputs change, we compute the Wasserstein distance~\cite{arjovsky2017wasserstein} between each perturbed output and the baseline output. These distances are calculated in a semantic embedding space using Word Mover’s Distance (WMD)~\cite{kusner2015word}, quantifying the dissimilarity between texts based on the underlying word embeddings.

These distance values are then transformed using a Gaussian kernel to generate weights that reflect the proximity of each perturbed prompt to the original. A locally weighted linear regression model~\cite{ribeiro2016should, molnar2020interpretable} is trained using these weights, with each perturbed prompt represented as a binary feature vector indicating the inclusion or exclusion of specific words.

The regression coefficients reflect the influence of each word in the original prompt on the generated output. These values are visualised as a heatmap over the input prompt, highlighting the most influential words in their contribution to the final text.

The overall process leverages SMILE’s model-agnostic, distribution-based explanation strategy~\cite{aslansefat2023explaining} and adapts it to the unique characteristics of text generation, where semantic consistency and local perturbation effects are critical. Fig.~\ref{fig:flow} illustrates this adapted pipeline.

\section{Experimental Results}

This section evaluates the proposed method's ability to enhance explainability in instruction-based text generation. We analyse how the structure of textual prompts affects interpretability, how variations in input text influence the clarity of explanations, and how the proposed solution performs across diverse scenarios. 

Experiments were conducted using a range of datasets and three state-of-the-art text generation models, OpenAI GPT~\cite{brown2020language}, LLaMA~\cite{touvron2023llama}, and Claude-AI~\cite{anthropic2023claude} to ensure a comprehensive assessment.

\subsection{Qualitative Results}

The qualitative evaluation demonstrates the proposed framework's interpretability using scenario-based testing and visualisations. We designed diverse testing scenarios by varying textual prompts, including changes in sentence complexity, rephrasing, and the inclusion of domain-specific terminology. These scenarios examine how the model interprets and generates outputs based on nuanced variations in textual inputs.

\subsubsection{Input Prompt Explainability}

The proposed framework applies scenario-based testing across multiple prompt variations to evaluate the interpretability of the model's behaviour regarding different inputs. These variations include rephrased questions, descriptive modifiers, and changes in word choice. The system generates a text heatmap highlighting individual words' contribution to the model's output for each input, with colours indicating the importance scores.

\begin{figure}[H]
    \centering
    \includegraphics[width=1\textwidth]{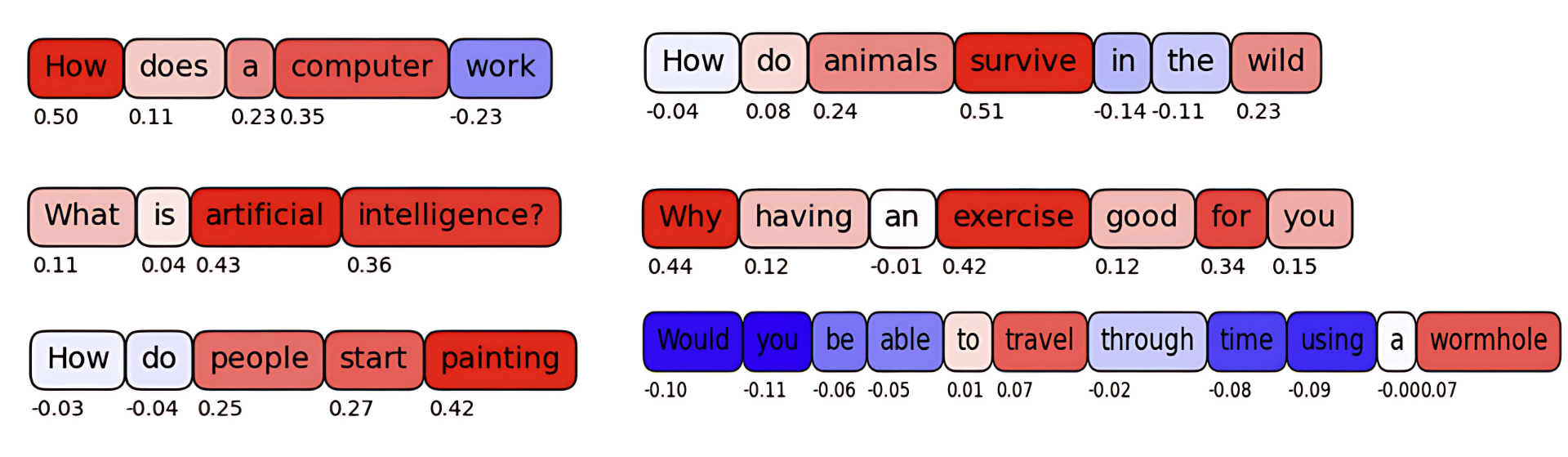}
    \caption{Text heatmaps showing word-level importance for various input prompts. The colour intensity represents the influence of each word on the model's output.}
    \label{fig:prompt_heatmaps}
\end{figure}

Fig~\ref{fig:prompt_heatmaps} shows how varying the input prompt's phrasing or focus leads to different word importance patterns, demonstrating the model's sensitivity to linguistic changes.

\subsubsection{Gender Fairness via Token-Level Attributions}

\begin{figure}[H]
    \centering
    \includegraphics[width=\linewidth]{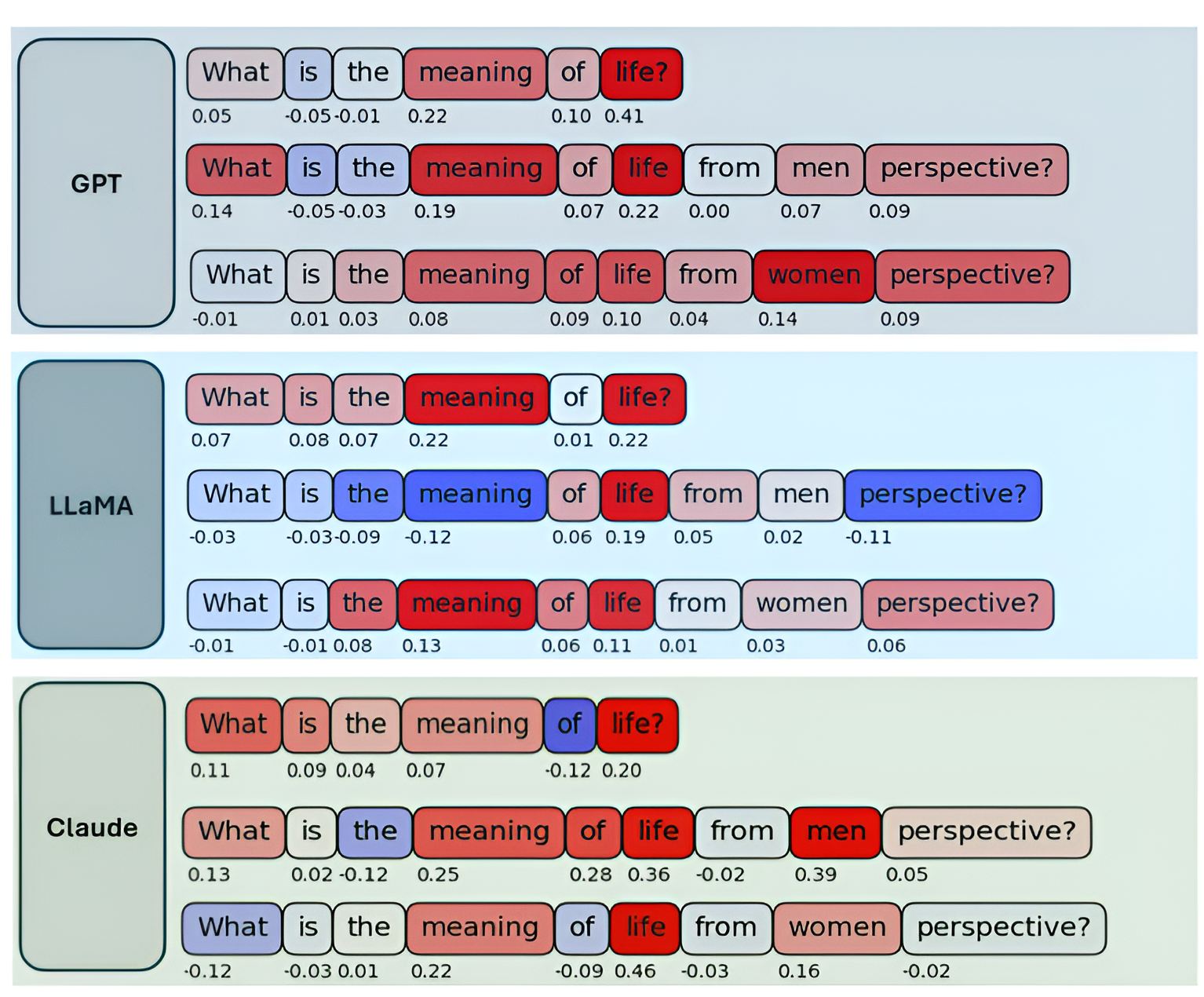}
    \caption{Comparison of gSMILE token-level attributions for GPT-3.5, LLaMA~3.1, and Claude~3.5 under three prompts: neutral, ``men perspective'', and ``women perspective''. Colour intensity encodes the relative influence of each token on the output distribution.}
    \label{fig:fairness-all}
\end{figure}

We further examined whether gendered phrasing affects attribution patterns by applying gSMILE to three closely related prompts: ``What is the meaning of life?'', ``What is the meaning of life from men's perspective?'', and ``What is the meaning of life from women's perspective?''. In all three models, the neutral formulation consistently highlighted the tokens \emph{meaning} and \emph{life}, underscoring their central role in shaping the response. Once a gender qualifier was introduced, however, the attribution heatmaps shifted, revealing model-specific sensitivities. As shown in Fig.~\ref{fig:fairness-all}, GPT-3.5 tended to assign greater weight to the token \emph{women} than \emph{men}, suggesting an asymmetric influence of gender terms. Claude~3.5 distributed its attention more evenly, balancing the contribution of both gendered tokens while still raising their importance relative to the baseline. By contrast, LLaMA~3.1 exhibited higher variability, with attribution sometimes amplifying one gender reference more strongly than the other.

These findings illustrate how gSMILE exposes fairness-related sensitivities that remain hidden if only the generated text is inspected. A gender term that absorbs disproportionate attribution suggests an underlying association that could bias the model’s reasoning. From a practical standpoint, this analysis highlights that avoiding gendered modifiers in prompts helps maintain attention on the semantic core of the task, thereby reducing the risk of unintended attribution drift.

\subsubsection{Reasoning Path and Token Attribution}

\begin{figure}[H]
    \centering
    \includegraphics[width=1\linewidth]{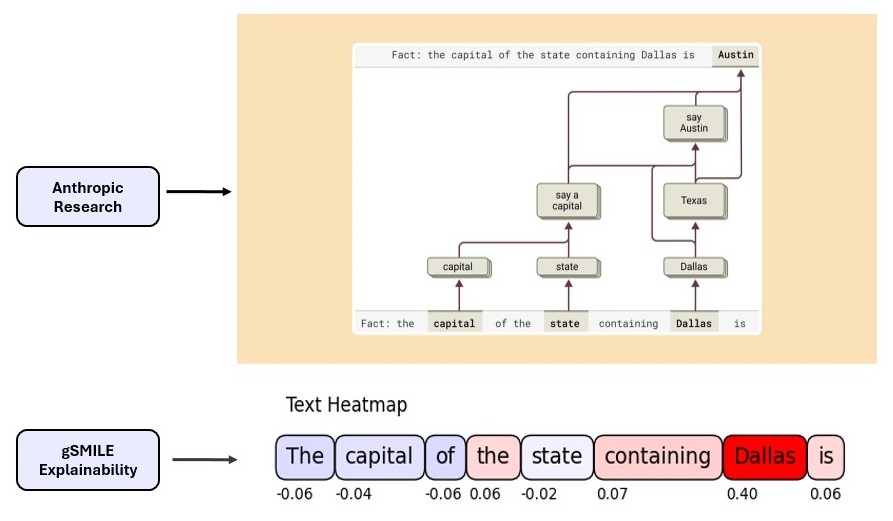}
    \caption{Comparison between attribution visualisations by Anthropic's Attribution Graphs and the gSMILE framework. The upper section shows the reasoning path identified by Anthropic's method, while the lower section illustrates gSMILE's token-level importance heatmap for the same prompt.~\cite{anthropic2025attribution}}
    \label{fig:smile_heatmap_llm}
\end{figure}

Beyond token-level importance visualisations, the proposed framework provides structured reasoning paths that illustrate how the model logically connects input elements to produce its response. Fig.~\ref{fig:smile_heatmap_llm} presents an example where the gSMILE method not only assigns attribution scores to individual tokens but also generates a reasoning flow diagram. This combined visualisation offers a holistic understanding of local word-level contributions and the logical structure of the model's decision-making.

Such comprehensive visual explanations align with advanced techniques used by leading AI research organisations, including Anthropic and OpenAI, while providing greater flexibility across diverse tasks and models. As shown in Fig.~\ref{fig:smile_heatmap_llm}, gSMILE's explainability aligns with or exceeds the interpretability achieved by leading industry methods~\cite{anthropic2025attribution}.

\subsubsection{Word-Level Attribution Analysis on MMLU Using gSMILE}

To evaluate the interpretability of ChatGPT using the gSMILE framework, we used examples from the MMLU (Massive Multitask Language Understanding) dataset~\cite{hendrycks2020measuring}. This benchmark contains four-choice multiple-choice questions covering various academic and professional subjects, including science, history, law, and the humanities. For this study, we accessed a pre-formatted version of the dataset available on Kaggle~\cite{mmlu_kaggle} to facilitate prompt construction and experimentation.

\begin{figure}[H]
    \centering
    \includegraphics[width=0.8\textwidth]{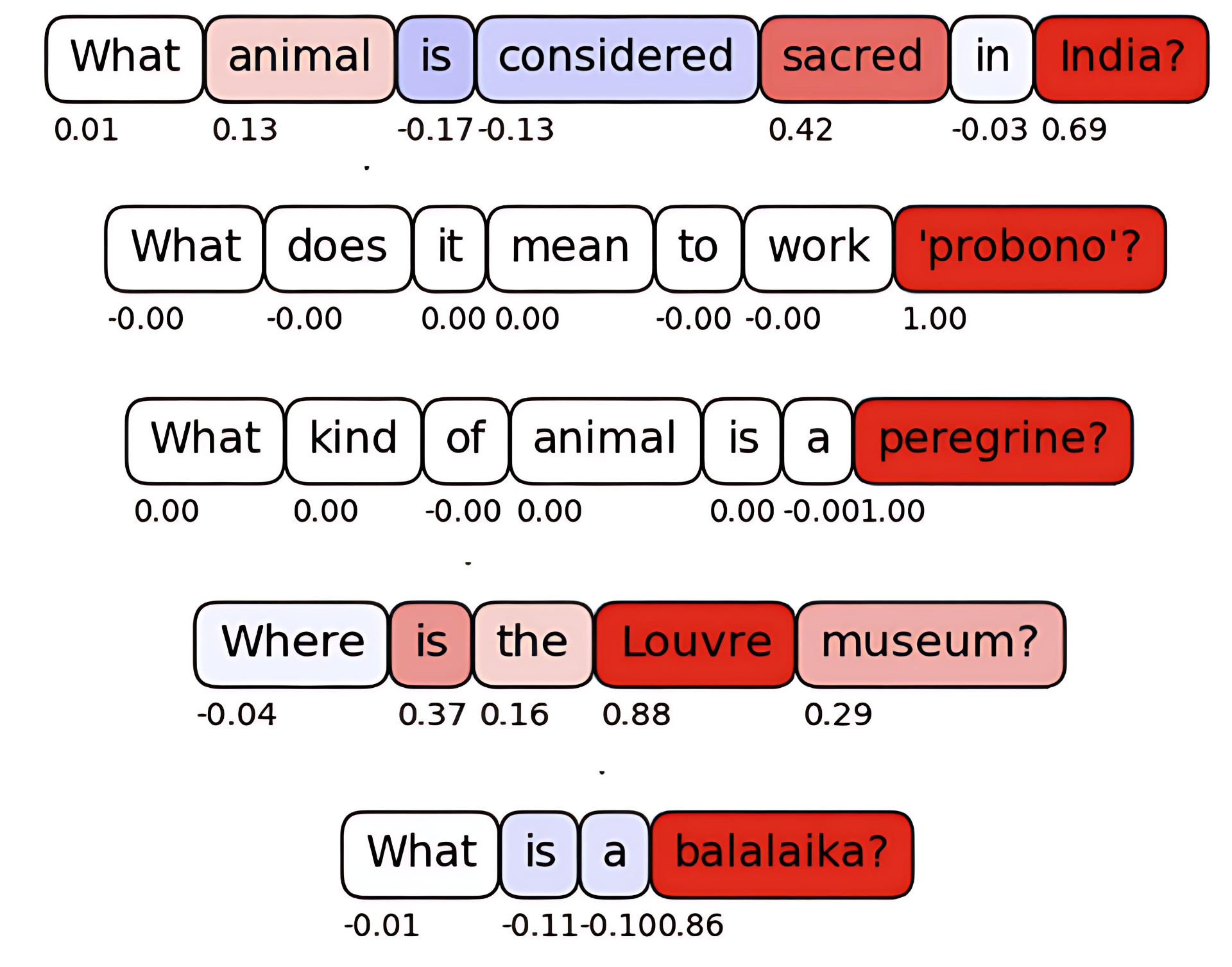}
    \caption{Attribution heatmaps generated by gSMILE using ChatGPT (GPT-4) on selected samples from the MMLU dataset. Highlighted words represent those with the highest influence on the model's output.}
    \label{fig:llm_heatmaps_mmlu}
\end{figure}

Fig.~\ref{fig:llm_heatmaps_mmlu} shows a set of attribution heatmaps that visualise which parts of the input prompt most influenced the model's decision. Darker red tones indicate higher influence, while neutral or irrelevant tokens are shaded lighter. These visualisations demonstrate gSMILE's ability to reveal the model's internal reasoning by tracing the most impactful words in multi-choice question contexts.

\subsection{Quantitative Results}

In this section, we apply several evaluation metrics, including accuracy, stability, consistency, fidelity, and computational complexity, to assess our explainability method across various text generation models and scenarios.

\subsubsection{ATT Accuracy}

To evaluate the performance of instruction-based text generation models, we tested ten different prompts and their variations through 64 perturbations. Heatmaps were generated to display the weights of each keyword in the prompts across various models, including OpenAI GPT, LLAMA, and Claude-AI. A ground truth was defined, assigning weights of 1 to critical keywords in the prompts and 0 to less relevant or auxiliary words. These ground truth values were compared with the extracted heatmaps to assess the interpretability and accuracy of each model.

To quantify accuracy, we employed multiple metrics, including Attention Accuracy (ATT ACC), F1-Score for Attention (ATT F1), and Area Under the Receiver Operating Characteristic Curve for Attention (ATT AUROC). The results were averaged across the ten different prompts, and each model's performance is summarised in Table~\ref{tab:Accuracy_models}, showcasing their effectiveness based on these metrics.

Among the evaluated models, \textbf{Claude-AI demonstrated the strongest performance}, achieving the highest scores in all three metrics: \textbf{0.82} in ATT ACC, \textbf{0.67} in ATT F1, and \textbf{0.88} in ATT AUROC. These results indicate Claude-AI’s superior ability to focus on essential input tokens across varying prompts consistently. In contrast, \textbf{LLAMA} achieved the second-highest accuracy but showed a notably lower F1-score (\textbf{0.40}), suggesting potential inconsistency in identifying key tokens. \textbf{OpenAI GPT} yielded balanced results but lagged slightly behind Claude-AI in all metrics, highlighting room for improvement in its attention fidelity. Ll, Claude-AI provides more reliable and semantically focused attention compared to its counterparts.

\begin{table}[H]
\centering
\caption{Performance metrics (ATT ACC, ATT F1, and ATT AUROC) for various\\models, evaluated under two conditions: different prompts and different images.}
\label{tab:Accuracy_models}
\renewcommand{\arraystretch}{1.3}
\scalebox{0.8}{
\begin{tabular}{lccc}
\hline \hline
\textbf{Model name} & \textbf{ATT ACC} & \textbf{ATT F1} & \textbf{ATT AUROC} \\ \hline
OpenAI-GPT    & 0.70 & 0.59 & 0.84 \\ 
LLaMA         & 0.76 & 0.40 & 0.84 \\
Claude-AI & \textbf{0.82} & \textbf{0.67} & \textbf{0.88} \\
\hline \hline
\end{tabular}
}
\end{table}

\subsubsection{ATT Stability}

To quantify stability, we use the Jaccard index, a metric that measures the similarity between sets by comparing their overlap. For text generation models such as OpenAI GPT, LLAMA, and Claude-AI, we tested ten different prompts and extracted the coefficients and weights assigned to each word. To evaluate stability, we added only the token \texttt{\*\*\*} at the end of the text, extracted the coefficients again, and computed the Jaccard index between the coefficients before and after the perturbation. 
The averaged Jaccard indices across all ten prompts for each model are reported in Table~\ref{tab:stability_models_llm}. Higher scores indicate greater stability in response to minor input changes.

\begin{table}[H]
\centering
\caption{Stability scores across models based on average Jaccard index over ten prompts}
\label{tab:stability_models_llm}
\renewcommand{\arraystretch}{1.3}
\scalebox{0.8}{
\begin{tabular}{lc}
\hline \hline
\textbf{Model name}  & \textbf{Jaccard Index}  \\ \hline
\textbf{OpenAI-GPT}    & \textbf{0.62} \\
LLaMA         & 0.45 \\
Claude-AI     & 0.44 \\ 
\hline \hline
\end{tabular}
}
\end{table}

\subsubsection{ATT Consistency}

To compute consistency, we ran the same code with 64 perturbations for each prompt over 10 iterations, ensuring that the outputs remained consistent across these repetitions for text generation models such as OpenAI GPT, LLAMA, and Claude-AI, as shown in Table~\ref{tab:consistency_models_llm}. We computed variance and standard deviation metrics for each word coefficient. This step is essential to confirm that the model's predictions are not random or heavily dependent on initialisation factors but are instead deterministic and robust.

Among all models, OpenAI-GPT exhibited the highest consistency, achieving a variance of 0.0000 and a standard deviation of 0.0046. This indicates that its predictions are the most stable across repeated runs, confirming its robustness and reliability in generating interpretable outputs.

\begin{table}[H]
\centering
\caption{Consistency metrics for different models on the prompt\\``What is the meaning of life?''}
\label{tab:consistency_models_llm}
\renewcommand{\arraystretch}{1.3}
\scalebox{0.8}{
\begin{tabular}{lcc}
\hline \hline
\textbf{Model name}  & \textbf{Variance} & \textbf{Standard Deviation} \\ \hline
\textbf{OpenAI-GPT}    & \textbf{0.0000} & \textbf{0.0046} \\
LLaMA         & 0.0048 & 0.0663  \\
Claude-AI     & 0.0011 & 0.0331 \\ 
\hline \hline
\end{tabular}
}
\end{table}

\subsubsection{ATT Fidelity for Different Instruct image editing diffusion-based Models (IED)}

The fidelity computation for various text generation models applied to the prompt ``What is the meaning of life?'' is displayed below. We compare several models, including OpenAI GPT, LLAMA, and Claude-AI, using 32 perturbations. The comparisons are made using a weighted linear regression as the surrogate model and the Wasserstein distance to compute distances across metrics such as MSE, (R$^2_\omega$), MAE, mean \(L_1\), and mean \(L_2\) losses. The results are presented in Table~\ref{tab:fidelity_models_llm}.

\begin{table}[H]
\centering
\caption{Fidelity metrics for different models on the prompt\\``What is the meaning of life?''}
\label{tab:fidelity_models_llm}
\renewcommand{\arraystretch}{1.3}
\scalebox{0.8}{
\begin{tabular}{lcccccc}
\hline \hline
\textbf{Model name} & \textbf{WMSE} & \textbf{R$^2_\omega$} & \textbf{WMAE} & \textbf{mean-L1} & \textbf{mean-L2} & \textbf{R$^2_{\hat{\omega}}$} \\ \hline
OpenAI-GPT    & 0.0388 & \textbf{0.7104}  & 0.1731 & 0.2035 & \textbf{0.0609} & \textbf{0.6409} \\
LLaMA         & 0.0368 & 0.7068 & 0.1617 & 0.2387 & 0.0805 & 0.6364 \\
Claude-AI     & \textbf{0.0265} & 0.6967  & \textbf{0.1251} & \textbf{0.1981} & 0.0708 & 0.6240 \\
\hline \hline
\end{tabular}
}
\end{table}

From the results, we observe that Claude-AI achieves the lowest WMSE, WMAE, and mean-L1, indicating superior performance in minimising error. On the other hand, OpenAI-GPT performs best in terms of \(R^2_\omega\), mean-L2, and adjusted \(R^2_{\hat{\omega}}\), demonstrating more substantial alignment and generalisation of the surrogate model. LLaMA ranks in the middle across most metrics but does not lead in any of them. This suggests that different models excel in different fidelity dimensions, highlighting the importance of multi-metric evaluation when interpreting the behaviour of black-box models.

\subsubsection{ATT Fidelity Across Different Numbers of Text Perturbations}

Tables~\ref{tab:fidelity_per_llm},~\ref{tab:fidelity_per_turbo_llm}, and~\ref{tab:fidelity_per_turbo_claude} present the results of fidelity computations for varying numbers of perturbations across the OpenAI GPT, LLAMA, and Claude-AI models, respectively. These results are derived using a weighted linear regression as the surrogate model and the Wasserstein distance for computing distances. The evaluation metrics include weighted mean squared error (WMSE), weighted mean absolute error (WMAE), coefficient of determination ($R^2_\omega$ and $R^2_{\hat{\omega}}$), and average loss metrics such as mean-L1 and mean-L2.

\begin{table}[H]
\centering
\caption{Performance metrics for different numbers of perturbations in OpenAI GPT}
\label{tab:fidelity_per_llm}
\renewcommand{\arraystretch}{1.3}
\scalebox{0.8}{
\begin{tabular}{rcccccc}
\hline \hline
\textbf{\#Perturb} & \textbf{WMSE} & \textbf{R$^2_\omega$} & \textbf{WMAE} & \textbf{mean-L1} & \textbf{mean-L2} & \textbf{R$^2_{\hat{\omega}}$} \\ \hline
32   & 0.0388 & \textbf{0.7104}  & 0.1731 & 0.2035 & 0.0609 & \textbf{0.6409} \\
64   & 0.0617 & 0.3406  & 0.1695 & 0.1756 & 0.0526 & 0.2712  \\
128  & 0.0329 & 0.4468  & 0.1385 & 0.1519 & 0.0378 & 0.4193 \\
256  & \textbf{0.0116} & 0.4276  & \textbf{0.0675} & \textbf{0.0872} & \textbf{0.0163} & 0.4138 \\
\hline \hline
\end{tabular}
}
\end{table}

\begin{table}[H]
\centering
\caption{Performance metrics for different numbers of perturbations in LLaMA}
\label{tab:fidelity_per_turbo_llm}
\renewcommand{\arraystretch}{1.3}
\scalebox{0.8}{
\begin{tabular}{rcccccc}
\hline \hline
\textbf{\#Perturb} & \textbf{WMSE} & \textbf{R$^2_\omega$} & \textbf{WMAE} & \textbf{mean-L1} & \textbf{mean-L2} & \textbf{R$^2_{\hat{\omega}}$} \\ \hline
32   & 0.0368 & \textbf{0.7068} & 0.1617 & 0.2387 & 0.0805 & \textbf{0.6364} \\
64   & 0.0240 & 0.4844 & 0.1284 & 0.1433 & 0.0382 & 0.4301  \\
128  & 0.0085 & 0.2430  & 0.0638 & \textbf{0.0612} & \textbf{0.0096} & 0.2055 \\
256  & \textbf{0.0058} & 0.2387 & \textbf{0.0605}  & 0.0667 & 0.0118 & 0.2203  \\
\hline \hline
\end{tabular}
}
\end{table}

\begin{table}[H]
\centering
\caption{Performance metrics for different numbers of perturbations in Claude-AI}
\label{tab:fidelity_per_turbo_claude}
\renewcommand{\arraystretch}{1.3}
\scalebox{0.8}{
\begin{tabular}{rcccccc}
\hline \hline
\textbf{\#Perturb} & \textbf{WMSE} & \textbf{R$^2_\omega$} & \textbf{WMAE} & \textbf{mean-L1} & \textbf{mean-L2} & \textbf{R$^2_{\hat{\omega}}$} \\ \hline
32   & 0.0368 & \textbf{0.7209}  & 0.1495 & 0.2262 & 0.0733 & \textbf{0.6539} \\
64   & \textbf{0.0115} & 0.3708  & \textbf{0.0564} & 0.1036 & 0.0370 & 0.3046 \\
128  & 0.0089 &  0.1506 & 0.0535 & 0.0920 & 0.0192 & 0.1085 \\
256  & 0.0191 & 0.2670 & 0.0902 & \textbf{0.0873} & \textbf{0.0163} & 0.2494 \\
\hline \hline
\end{tabular}
}
\end{table}

For OpenAI GPT and LLaMA, increasing the number of perturbations generally improves fidelity across most metrics, with 256 perturbations achieving the best WMSE, WMAE, and loss values. Interestingly, for Claude-AI, 64 perturbations yield the lowest WMSE and WMAE, while 256 achieves the best loss (mean-L2 and mean-L1). However, the highest \(R^2\) scores for all three models occur at 32 perturbations, suggesting that beyond a certain point, more perturbations may dilute interpretability despite improving local fit. Therefore, there is a trade-off between the number of perturbations and the reliability of the surrogate model, depending on the metric of interest.

\subsubsection{ATT Fidelity for Different Distance Metrics and Surrogate Models}

Fidelity is also computed by comparing various distance metrics for text generation models, such as OpenAI GPT, using text-to-text comparisons. The surrogate models used are Weighted Linear Regression (WLR) and Bayesian Ridge (BayLIME), as shown in Table~\ref{tab:fidelity_dist_llm}. We explore different combinations of distance metrics, including Cosine similarity and Wasserstein Distance (WD). Fidelity is measured using WMSE, WMAE, mean-\(L_1\), mean-\(L_2\), and coefficient of determination ($R^2_\omega$, $R^2_{\hat{\omega}}$).

\begin{table}[H]
\centering
\caption{Fidelity results for different distance measures with 30 perturbations}
\label{tab:fidelity_dist_llm}
\renewcommand{\arraystretch}{1.2}
\scalebox{0.75}{
    \begin{tabular}{llcccccc}
    \hline \hline
    \multicolumn{2}{c}{\textbf{WLR}} & \multicolumn{6}{c}{\textbf{Fidelity Metrics}} \\
    \cmidrule(lr){1-2} \cmidrule(lr){3-8}
    \textbf{T vs T} & \textbf{T vs T} &  \textbf{WMSE} & \textbf{R$^2_\omega$} & \textbf{WMAE} & \textbf{mean-L1} & \textbf{mean-L2} & \textbf{R$^2_{\hat{\omega}}$} \\
    \midrule
    Cosine & Cosine   & 0.0172 & 0.3151 & 0.0659 & 0.1277 & 0.0412 & 0.1508 \\
    Cosine & WD       & 0.0216 & 0.4197 & 0.0899 & 0.1332 & 0.0385 & 0.2805 \\
    WD     & WD       & 0.0388 & \textbf{0.7104} & 0.1731 & 0.2035 & 0.0609 & \textbf{0.6409} \\
    WD     & Cosine   & 0.0048 & 0.4026 & 0.0329 & 0.0871 & 0.0296 & 0.2593 \\
    WD+C   & WD+C     & 0.0349 & 0.5349 & 0.1589 & 0.3050 & 0.1468 & 0.4233 \\
    \midrule
    \multicolumn{2}{c}{\textbf{BayLIME}} & \multicolumn{6}{c}{} \\
    \midrule
    Cosine & Cosine   & 0.0200 & 0.2285 & 0.0499 & 0.1067 & 0.0464 & 0.0434 \\
    Cosine & WD       & 0.0241 & 0.3572 & 0.0736 & 0.1281 & 0.0430 & 0.2029 \\
    WD     & WD       & 0.0118 & \textbf{0.5837} & 0.0629 & 0.1170 & 0.0361 & \textbf{0.4838} \\
    WD     & Cosine   & 0.0058 & 0.5273 & 0.0459 & 0.0978 & 0.0304 & 0.4139 \\
    WD+C   & WD+C     & 0.0048 & 0.4227 & 0.0284 & 0.0690 & 0.0213 & 0.2841 \\
    \hline \hline
    \end{tabular}
}
\end{table}

\vspace{1em}

\paragraph{Our Proposed Method: gSMILE.}  
Based on the results in Table~\ref{tab:fidelity_dist}, the combination where Wasserstein Distance (WD) is used for both input similarity and output shift consistently yields the highest values of the weighted coefficient of determination ($R^2_\omega$ and $R^2_{\hat{\omega}}$). Specifically, underweighted linear regression, the WD–WD configuration achieves $R^2_\omega = \textbf{0.7104}$ and $R^2_{\hat{\omega}} = \textbf{0.6409}$, which significantly outperform other configurations. 

Motivated by these findings, we introduce our method, called \textbf{gSMILE} (generalised SMILE), which utilises the Wasserstein distance for both input and output comparisons. This choice enhances the surrogate model's fidelity and stability in capturing local shifts in model behaviour.

We use gSMILE throughout subsequent sections as the foundation for explainability evaluations on large language models (LLMs). These include assessments of fidelity, consistency, and stability, ensuring the proposed approach is both theoretically grounded and empirically adequate.

\subsubsection{Computation Complexity}

We evaluate the computational efficiency of our proposed method, \textbf{gSMILE}, using three instruction-based text generation models: OpenAI GPT, LLaMA, and Claude-AI. Although these models differ in scale and architecture, the objective here is not to compare the models themselves but rather to assess the performance of different \textit{explainability methods} applied to each model under identical hardware and runtime conditions. All evaluations are conducted with 60 perturbations per input prompt.

We compare three explainability approaches—LIME, gSMILE, and Bay-LIME—based on their execution time. As reported in Table~\ref{tab:execution_times_llm}, gSMILE consistently requires the highest runtime across all model frameworks. This is mainly due to its reliance on the Wasserstein distance to compute similarity between output distributions, which involves solving high-dimensional optimisation problems with time complexity \(O(Nd^3)\), where \(N\) is the number of perturbations and \(d\) is the embedding dimensionality~\cite{cuturi2014fast}. By contrast, LIME and Bay-LIME utilise cosine similarity with lower complexity \(O(Nd)\), making them faster but potentially less precise.

Despite the higher computational cost, gSMILE delivers significantly better \textit{fidelity}, \textit{stability}, and robustness to adversarial perturbations—essential properties for trustworthy explanations. These advantages are supported by results in Tables~\ref{tab:fidelity_dist} and~\ref{tab:stability_models_llm} and align with findings from prior work on SMILE~\cite{aslansefat2023explaining}, which demonstrated superior robustness against adversarial input variations.

These findings suggest that gSMILE’s computational overhead is justified by the quality of explanations it provides, making it highly suitable for applications where interpretability, consistency, and resistance to manipulation are paramount.

\begin{table}[H]
\centering
\caption{Execution times (in seconds) for each framework and explainability method with 60 perturbations}
\label{tab:execution_times_llm}
\renewcommand{\arraystretch}{1.3}
\scalebox{0.8}{
\begin{tabular}{lccc}
\hline \hline
\textbf{Model name} & \textbf{LIME} & \textbf{gSMILE} & \textbf{Bay-LIME} \\ \hline
OpenAI-GPT  & 156.82 & \textbf{170.70} & 161.43 \\
LLaMA       & 113.00 & \textbf{118.99} & 116.40 \\
Claude-AI   & 355.09 & \textbf{372.95} & 349.88 \\
\hline \hline
\end{tabular}
}
\end{table}

\subsubsection{Token-Level Attention for Different Sentence Structures}

To investigate how sentence structure affects token-level attention in large language models (LLMs), we analysed the attention scores assigned to the tokens \textit{``meaning''} and \textit{``life''} across ten semantically similar but syntactically varied prompts. These prompts ranged from questions (e.g., ``What is the meaning of life?''), commands (e.g., ``Please give me the meaning of life''), to declarative formulations (e.g., ``You must explain the meaning of life'').

\begin{figure}[H]
\centering
\includegraphics[width=0.9\linewidth]{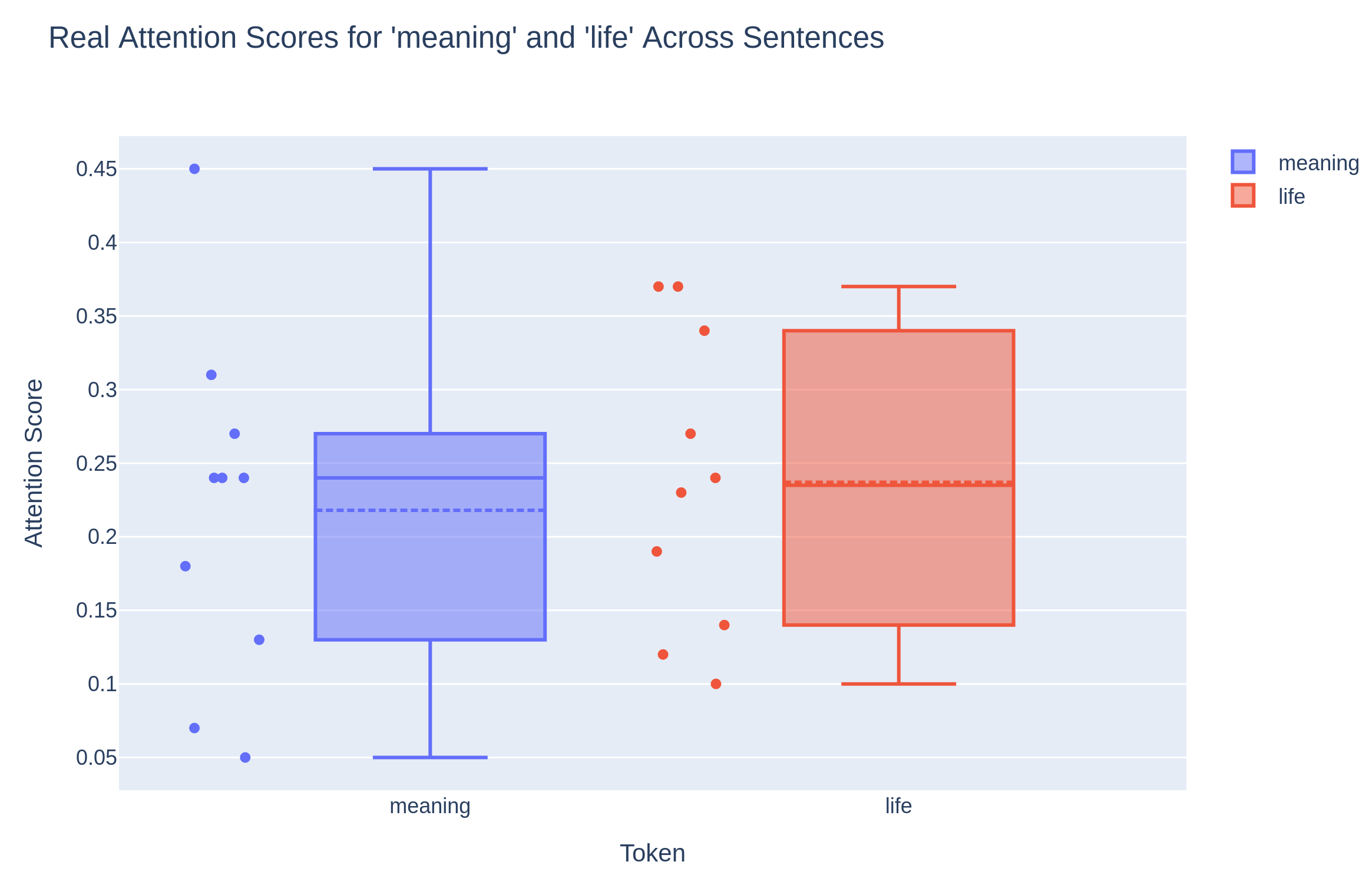}
\caption{Box plot showing attention scores for the tokens ``meaning'' and ``life'' across ten prompts with varying sentence structures. Each point represents the attention score assigned to the token in a specific sentence instance.}
\label{fig:token-attention-structure}
\end{figure}

For each sentence, we extracted the attention values directly from the final layer of the LLM. We focused on the attention allocated to the tokens ``meaning'' and ``life.'' The results were visualised using a box plot (see Fig~\ref{fig:token-attention-structure}), where each point represents a unique sentence. Full-sentence prompts are viewable through hover text, allowing for a qualitative comparison of sentence structure and attention distribution.

The analysis reveals noticeable variability in how these tokens are weighted depending on syntactic form. For instance, in imperative constructions, attention to ``life'' tends to dominate, whereas in interrogative or abstract prompts, ``meaning'' often receives comparable or higher attention. This suggests that sentence formulation can subtly steer the internal focus of the model, even when the semantic content remains consistent.

These findings underscore the sensitivity of LLMs to surface-level structure in shaping token-level interpretability, offering insight into both model behaviour and prompt engineering strategies.

\section{Conclusion}

This work introduced gSMILE, a generative, model-agnostic framework for explaining LLM behaviour through token-level attribution and visual reasoning aids.  Whereas most existing interpretability methods highlight essential tokens in the model’s output~\cite{ribeiro2016should, lundberg2017unified}, gSMILE shifts attention to the input space. By integrating perturbation analysis with Wasserstein-based distance measures and weighted linear surrogates, gSMILE offers a principled approach to local interpretability without requiring access to internal model parameters.

Our evaluation across three  instruction-tuned LLMs revealed clear performance contrasts:

\begin{itemize}
    \item \textbf{Claude 2.1} achieved the highest attribution fidelity (AttAUROC = 0.88) and lowest error measures, indicating more semantically focused attention.
    \item \textbf{GPT-3.5-turbo-instruct} exhibited exceptional consistency (lowest variance and standard deviation across repeated runs), making it highly predictable in explanation output.
    \item \textbf{LLaMA 3.1} performed competitively but showed greater variability in token attribution, suggesting sensitivity to prompt phrasing.
\end{itemize}

Overall, the resulting heatmaps and attribution scores provide an intuitive understanding of model behaviour, empowering users to control and refine prompt design better. 

\textbf{gSMILE} was evaluated and compared against existing methods (LIME, Bay-LIME) using multiple explainability metrics, including fidelity, stability, and consistency. We systematically assessed how well each technique captured token-level influence in LLMs. The results show that gSMILE consistently achieved higher fidelity in approximating black-box behaviour, while also producing more stable and repeatable attributions under small prompt perturbations. Compared to LIME and Bay-LIME, gSMILE demonstrated improved robustness and alignment with human-interpretable patterns, underscoring its suitability for practical use in settings where reliable interpretability is critical.

These findings justify gSMILE’s contribution in two ways. First, it demonstrates that explainability metrics can meaningfully differentiate between LLMs beyond standard benchmark accuracy, providing new insight into their operational reliability. Second, it indicates that a purely black-box, post-hoc approach can produce explanations with stability and fidelity comparable to those of privileged-access interpretability tools.

At the same time, the limitations identified (prompt sensitivity, hallucination effects, computational overhead, and deployment constraints) highlight that the path toward fully practical interpretability remains open. Addressing these will be central to future research. Looking ahead, we see three promising directions:

\begin{enumerate}
    \item \textbf{Adaptive perturbation strategies:} Future work can move beyond random sampling by using targeted perturbations that focus only on influential tokens or phrases. This reduces computational overhead while preserving fidelity. Inspired by approaches such as IBM’s CELL framework, perturbations could involve changing specific words rather than only removing or keeping them. This enables the efficient highlighting of selected words in longer sentences.  

    \item \textbf{Causal verification methods:} To mitigate hallucination-driven or spurious attributions, causal analysis can be integrated (e.g., perturbing tokens while holding other context fixed) to confirm whether an attribution genuinely drives the model’s behaviour. This would strengthen robustness in high-stakes applications.  

    \item \textbf{Extension to multimodal and multilingual models:} gSMILE currently focuses on English text. Extending the framework to multilingual LLMs and multimodal systems (text–image or text–speech) would test whether stability and fidelity hold across diverse input spaces, improving generalisability for real-world deployment.  
\end{enumerate}

By uniting rigorous evaluation with interpretable visualisations, gSMILE makes LLMs more transparent without privileged access to their internals, thereby providing both researchers and practitioners with a practical tool for building more trustworthy AI systems.

\chapter{Explainability of Instruction-based Image Editing}
\label{chap:image_editing_explainability}

\begin{figure}[H]
    \centering
    \includegraphics[width=0.9\textwidth]{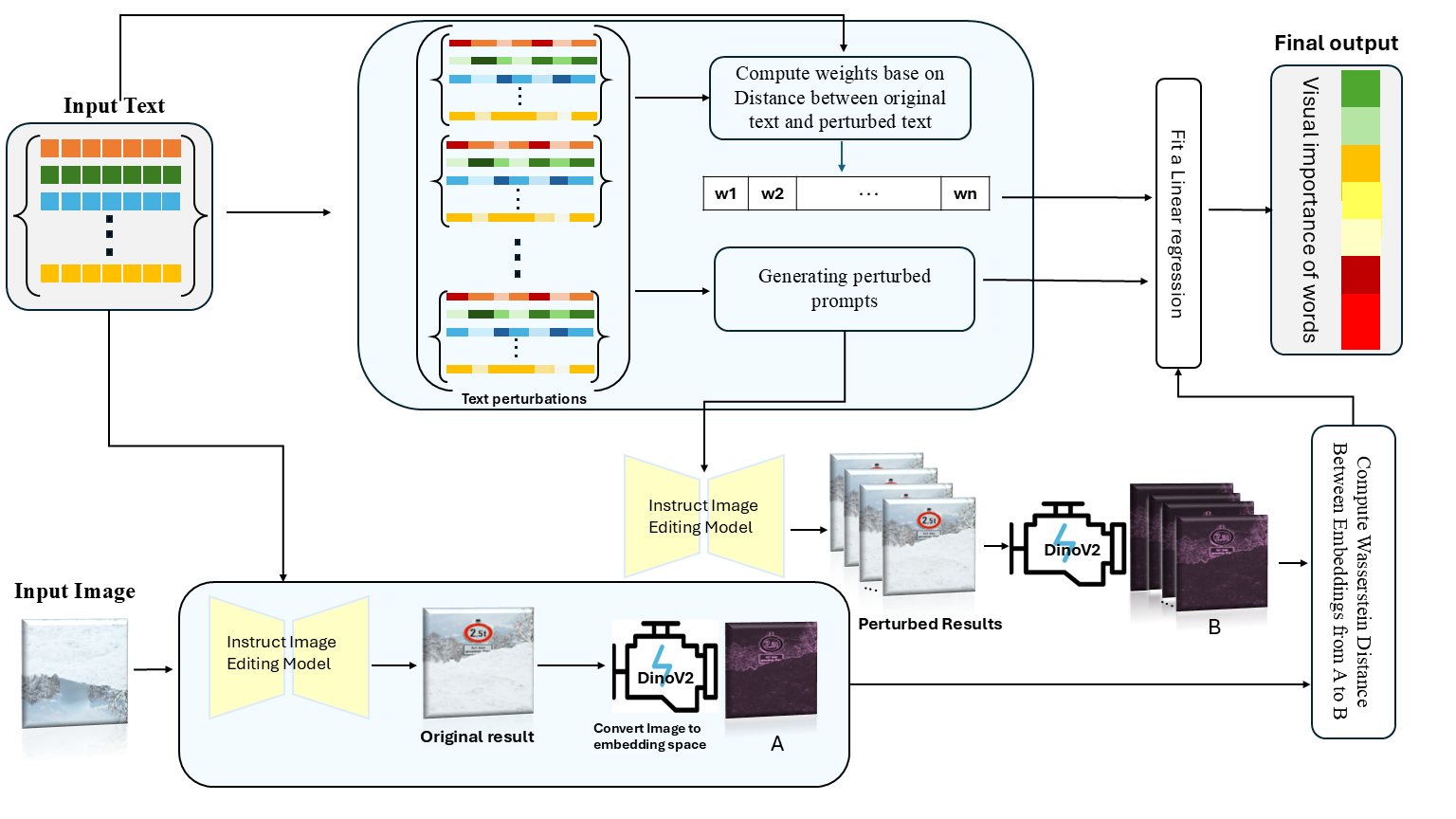} 
    \caption{The gSMILE framework for instruction-based image editing explainability.}
    \label{fig:smile_image_editing_flow}
\end{figure}

This chapter addresses \textbf{Research Question 3 (RQ3)} by exploring how interpretability methods, specifically the gSMILE framework, can be adapted to multimodal systems, such as instruction-based image editing models. It introduces techniques for interpreting how these models respond to natural language prompts. We present the gSMILE framework, which generates local visual explanations linking text instructions to specific image modifications. Our extensive testing evaluates the framework’s effectiveness across multiple metrics, including stability, accuracy, fidelity, and consistency, demonstrating its robustness in diverse editing scenarios. An overview of the proposed workflow is illustrated in Fig.~\ref{fig:smile_image_editing_flow}.

\section{Problem Definition}

Interpretability poses a significant challenge in complex deep learning models, as well as in extensive language and text-based image editing models. Due to the Enormous number of parameters and complex processing mechanisms, these models essentially act as black boxes, making it problematic for users to understand why and how the model responds to certain inputs or shows specific behaviors~\cite{doshi2017towards, rudin2019stop}. The need for interpretability poses a significant challenge to the safe deployment and control of these models. It makes it harder to manage them effectively and increases the risk of unexpected or undesirable behaviour.

For instance, to improve the interpretability of large language models, the Anthropic team applied Sparse Autoencoders to extract semantic and abstract features from these models. This technique enabled them to identify critical concepts and internal behaviours, thus providing a more transparent view of how these models operate~\cite{templeton2024scaling}.

Inspired by this approach and recognising the importance of interpretability in text-based image editing models, we present an innovative, model-agnostic approach for evaluating and enhancing the interpretability of these models. We leverage SMILE (Statistical et al. with Local Explanations) as a foundational tool to generate localised explanations and visual heatmaps. These heatmaps illustrate the influence of specific text elements on the image generation process, helping users better understand the relationship between textual inputs and visual outputs~\cite{aslansefat2023explaining}. This model-agnostic method is crucial for making instruction image editing models more transparent and trustworthy.

\begin{figure}[ht]
    \centering
    \includegraphics[width=1\linewidth]{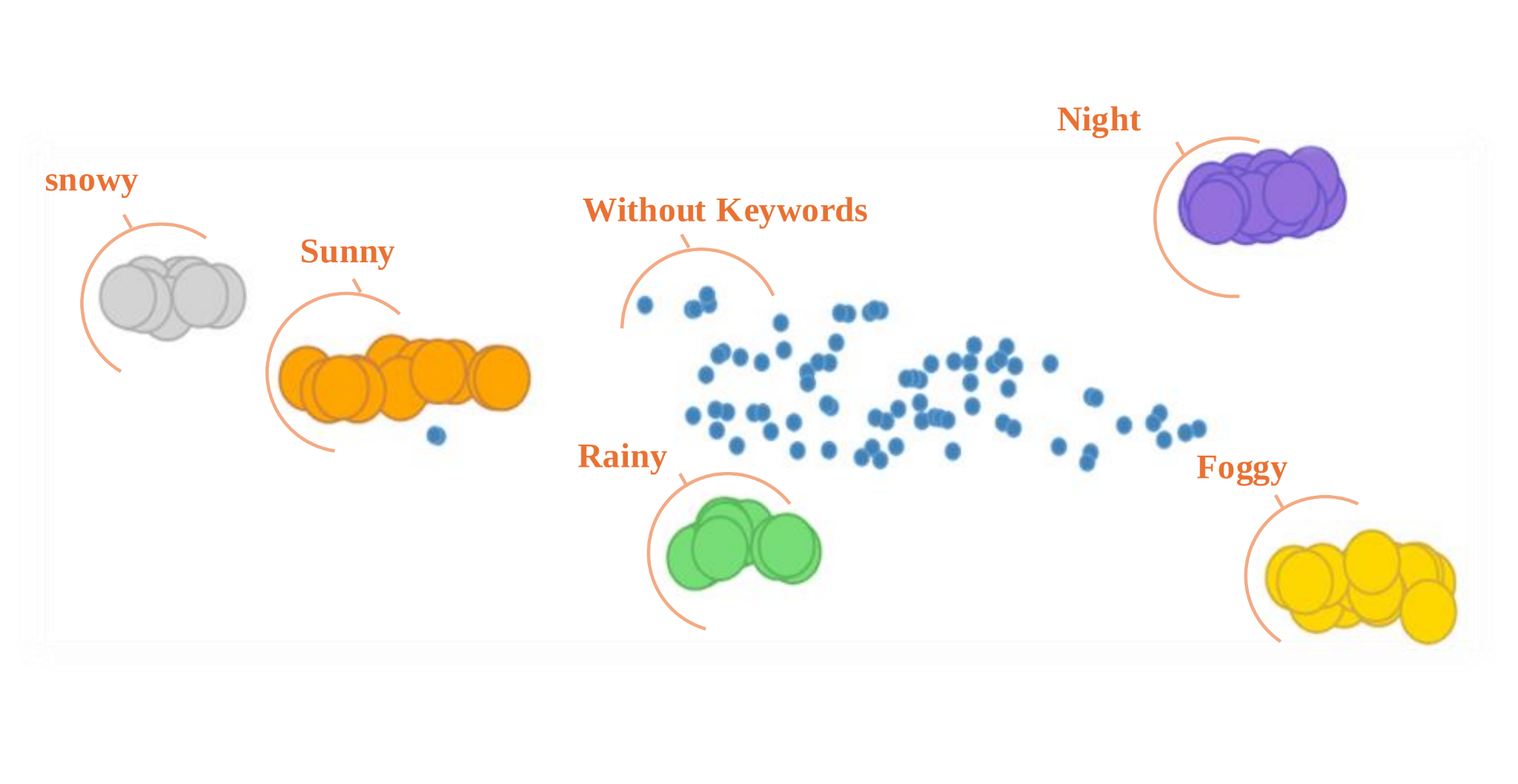}
   \caption{2D t-SNE visualisation of image embeddings, illustrating cluster separability for different prompt keywords and perturbed text inputs.}

    \label{fig:tsne_5}
\end{figure}

In addition, to further examine the model's interpretability, we incorporated a t-SNE scatter plot, as illustrated in Fig.~\ref{fig:tsne_5}, to visualise the embeddings of images modified with various descriptive sentences, each embedding colour-coded based on specific keywords. In this plot, images generated with sentences containing the keyword ``sunny'' are represented as orange points, ``rainy'' as green points, ``foggy`` as yellow points, ``snowy'' as grey points, and ``night'' as purple points. Additional points generated from sentences with perturbed texts lacking targeted keywords are represented as dark blue points.

This visualisation shows that embeddings of each keyword create discrete clusters, clearly separated from perturbation points without specific keywords. This separation suggests that a surrogate model, like weighted linear regression, could effectively capture the model's behaviour. Moreover, the regression coefficients provide an interpretable and straightforward measure of the importance of each keyword, indicating how it influences the model's output. This method enhances interpretability by offering a clear, quantitative view of how specific textual components impact the generated images.

\section{Method Summary for Instruction-based Image Editing}

In this section, we describe how the gSMILE framework introduced in Chapter~\ref{chap:gSMILE Methodology} is adapted to explain instruction-based image editing models. The objective is to identify which parts of a text prompt most significantly influence the visual transformations applied to an input image.

The process begins with a fixed input image and a corresponding instruction (text prompt). The prompt is first tokenised into individual words, and perturbed prompts are created by selectively removing or altering one word at a time. Each perturbed instruction is then used to guide the image editing model, resulting in a modified output image. The image generated using the original prompt serves as the baseline reference.

To quantify the impact of each prompt variation, we compute the Wasserstein distance~\cite{arjovsky2017wasserstein} between each perturbed image and the baseline image. These distances are calculated in the semantic embedding space using DINOv2~\cite{oquab2023dinov2}, a self-supervised vision transformer known for producing robust and general-purpose image representations. This allows us to compare outputs based on high-level visual semantics rather than low-level pixel differences.

The computed distances are converted into weights using a Gaussian kernel, representing the degree of similarity between each perturbed prompt and the original. These weights are then applied in a locally weighted linear regression model~\cite{ribeiro2016should, molnar2020interpretable}, where each input is a binary vector indicating which words were included in the perturbed prompt.

The resulting regression coefficients reveal the contribution of each word to the visual transformation performed by the model. These values are visualised as a heatmap overlay on the prompt, highlighting which words had the most significant effect on the generated image.

This adaptation enables interpretable explanations in multimodal tasks by extending SMILE’s statistical and model-agnostic approach to vision-language settings. It is especially valuable when stylistic or semantic image edits must be traced to specific textual instructions. Fig.~\ref{fig:flowchart} illustrates the SMILE pipeline applied to instruction-based image editing.

\section{Experimental Results}

This section evaluates the proposed ability to enhance explainability in instruction-based image editing. We analyse how the structure of textual prompts affects interpretability, how input images influence the clarity of explanations, and how the proposed solution performs across diverse scenarios. Experiments were conducted using a range of datasets and three state-of-the-art diffusion models: Instruct-Pix2Pix~\cite{an2023fine}, Img2Img-Turbo~\cite{parmar2024one}, and Diffusers-Inpaint~\cite{rombach2022high}, to ensure a comprehensive assessment.

\subsection{Qualitative Results}

The qualitative evaluation demonstrates the interpretability of the proposed framework using scenario-based testing and visualisations. Leveraging the Operational Design Domain (ODD)~\cite{zhang2024odd}, we generated diverse testing scenarios that encompass environmental factors, such as weather transformations (e.g., rain, fog, snow), and non-environmental variables (e.g., object attributes, human-centric factors, temporal conditions). The specific transformation scenarios, keywords, and controlled variables used in these evaluations are summarised in Tables~\ref{tab:weather_transformations} and~\ref{tab:table_of_co_2}, covering various environmental conditions and human-centric modifications.

\begin{table}[H]
    \centering
    \includegraphics[width=\linewidth]{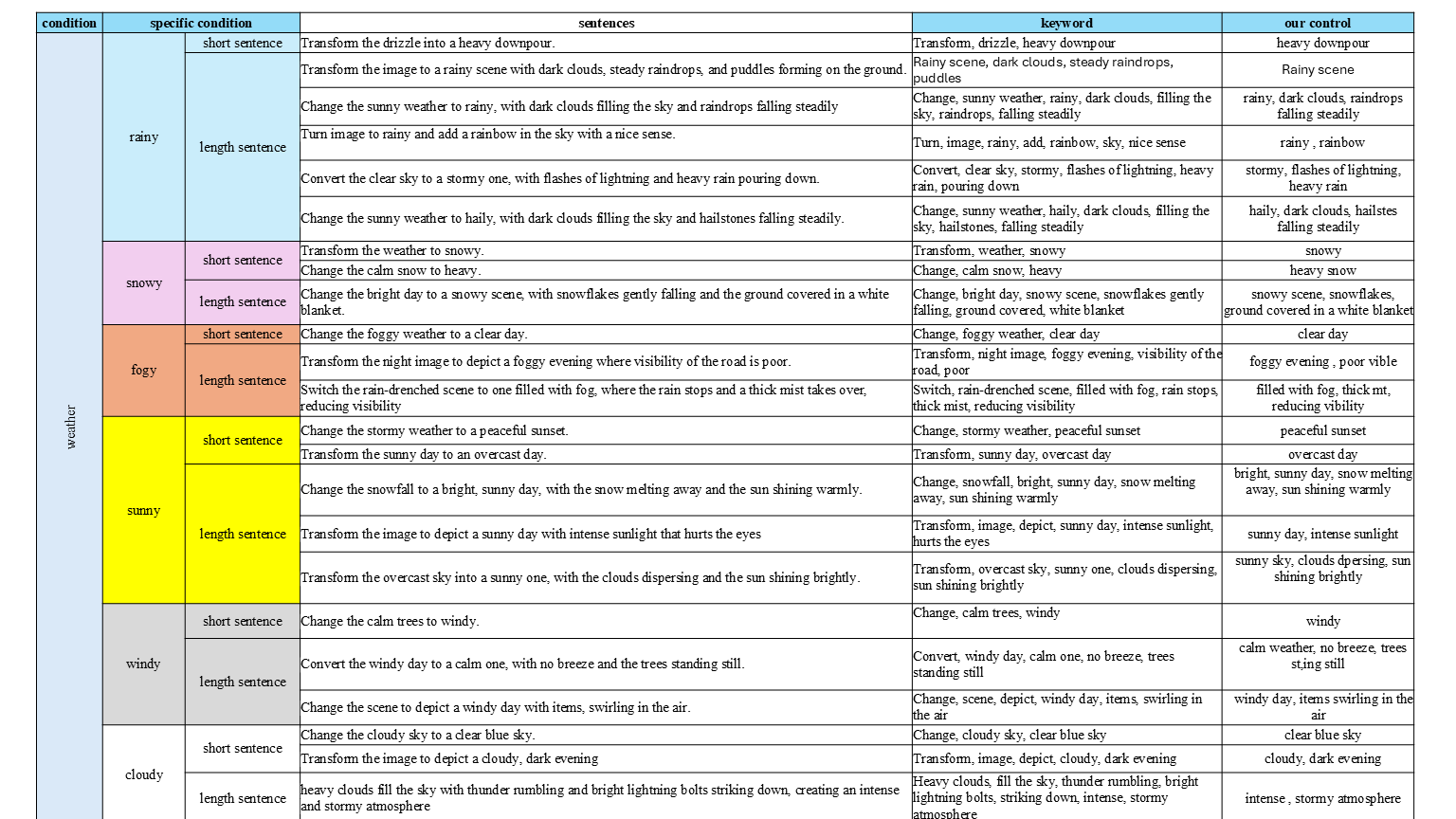}
    \caption{Weather Condition Transformation and Corresponding Keywords.}
    \label{tab:weather_transformations}
\end{table}

\begin{table}[H]
    \centering
    \includegraphics[width=\linewidth]{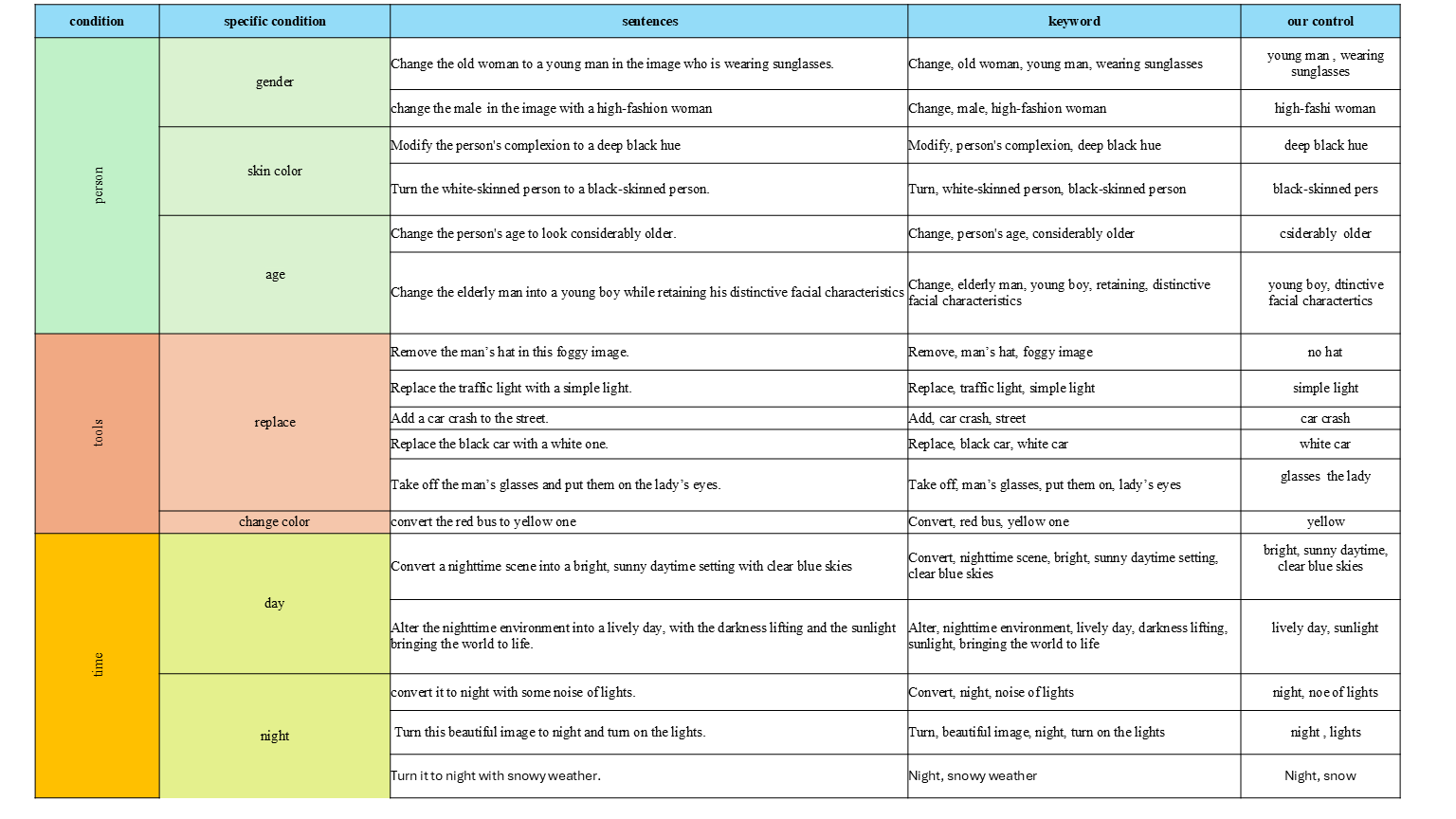}
    \caption{Transformation Guidelines for Visual and Descriptive Changes Based on Time, and Human Attributes.}
    \label{tab:table_of_co_2}
\end{table}

We analysed input prompt explainability by mapping word contributions to visual changes, highlighting the weight of keywords such as ``snowing'' in the editing process. Using t-SNE scatter plots, we qualitatively examined the clustering of images generated from perturbed prompts, revealing the impact of word-level variations on the generated outputs (Figs.~\ref{fig:TSNE1}).

\subsubsection{Input Prompt Explainability}

The proposed scenario-based testing framework leverages the system's Operational Design Domain (ODD) to generate scenarios that comprehensively test Automated Driving Systems (ADS) under a wide array of conditions~\cite{koopman2016challenges, zhang2024odd}. 

The concept of ODD, as highlighted in recent high-risk AI frameworks, is fundamental to understanding and ensuring the operational boundaries and limitations of ADS~\cite{wang2024survey}. As noted in broader discussions on high-risk AI systems, the ODD provides a structured way to classify environmental and situational variables, helping to consistently assess and validate system performance across various conditions~\cite{stettinger2024trustworthiness}.

Tables~\ref{tab:weather_transformations} and~\ref{tab:table_of_co_2} are developed based on the Operational Design Domain (ODD) framework to systematically simulate and assess a wide range of environmental and situational conditions affecting the performance of Automated Driving Systems (ADS). 

The ODD, central to high-risk AI frameworks, defines ADS's operational boundaries and limitations, enabling the structured creation of diverse testing scenarios~\cite{koopman2016challenges}. Table~\ref{tab:weather_transformations} is generated using a variety of instructions, focusing on environmental conditions and classifying scenarios by weather states such as rain, fog, and snow, ensuring thorough evaluation under diverse environmental factors~\cite{zhang2024odd}. 

Table~\ref{tab:table_of_co_2} incorporates various instructions to address non-environmental variables, including person-related attributes (e.g., age, gender, and skin colour), object variations (e.g., vehicle colours and traffic light configurations), and temporal factors (e.g., day and night scenarios), ensuring thorough testing across a range of human-centric and object-driven conditions~\cite{stettinger2024trustworthiness}. 

We implemented interpretability techniques using various image editing models, including Instruct-Pix2Pix~\cite{an2023fine}, Img2Img-Turbo~\cite{parmar2024one} and Diffusers-Inpaint~\cite{rombach2022high}. We based our approach on the analysis provided in the tables and incorporated heatmaps to demonstrate how specific keywords influence image modifications~\cite{koopman2016challenges}. This structured methodology is aligned with high-risk AI frameworks, such as the EU AI Act, which emphasises trustworthiness, safety, and regulatory compliance throughout the system's lifecycle~\cite{cancela2024eu}.

Additionally, it strengthens the robustness, reliability, and transparency of the Automated Decision System (ADS) evaluation process. Our design aims to enhance the trustworthiness of high-risk AI systems through instruction-based image editing models.

\begin{figure}[H]
    \centering
    \includegraphics[width=\linewidth]{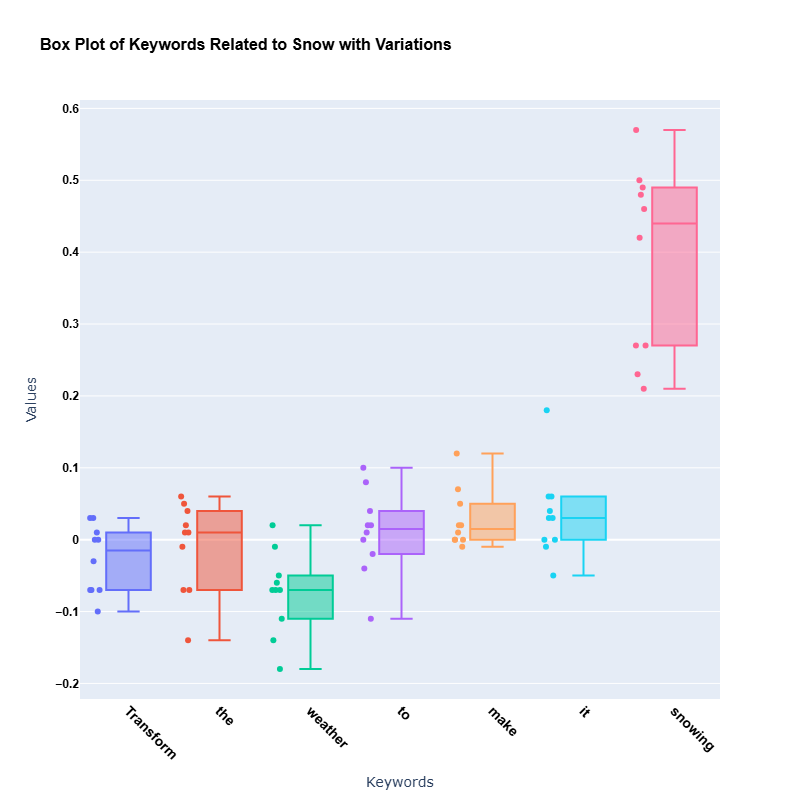}
    \caption{Box plot showing the distribution of word weights in the prompt ``Transform the weather to make it snowing''.}
    \label{fig:snow_keywords_boxplot}
\end{figure}

By aggregating the weights from the heatmaps across all ten images, we created a box plot to display the distribution of influence for each word in the prompt. As shown in Fig.~\ref{fig:snow_keywords_boxplot}, the box plot reveals the range and concentration of influence each word exerts.
The keyword ``snowing'' has the most substantial influence compared to other words in the prompt, indicating a significant effect on the model's editing process for creating snow and related adjustments.

\subsubsection{t-SNE Scatter Plot of Image Embeddings}

In this section, we present the qualitative results of our model using perturbed prompts. To better understand the influence of various prompt structures on the generated images, we visualised the results using a 2D plot. This plot displays the distances between images generated from different perturbations of the same base scenario. Each point in the plot represents an image, with the red dot corresponding to the image generated by the original prompt and the blue points representing the images generated by perturbed versions of the original prompt.

We systematically modified the prompt by adding or removing keywords, such as ``snowing''. As shown in the plot, perturbations that include the keyword ``snowing'' change the generated images, transforming the scenes into snowy ones. In contrast, the resulting images do not depict snow when the perturbations do not include this keyword.

\begin{figure}[H]
    \centering
    \includegraphics[width=\linewidth]{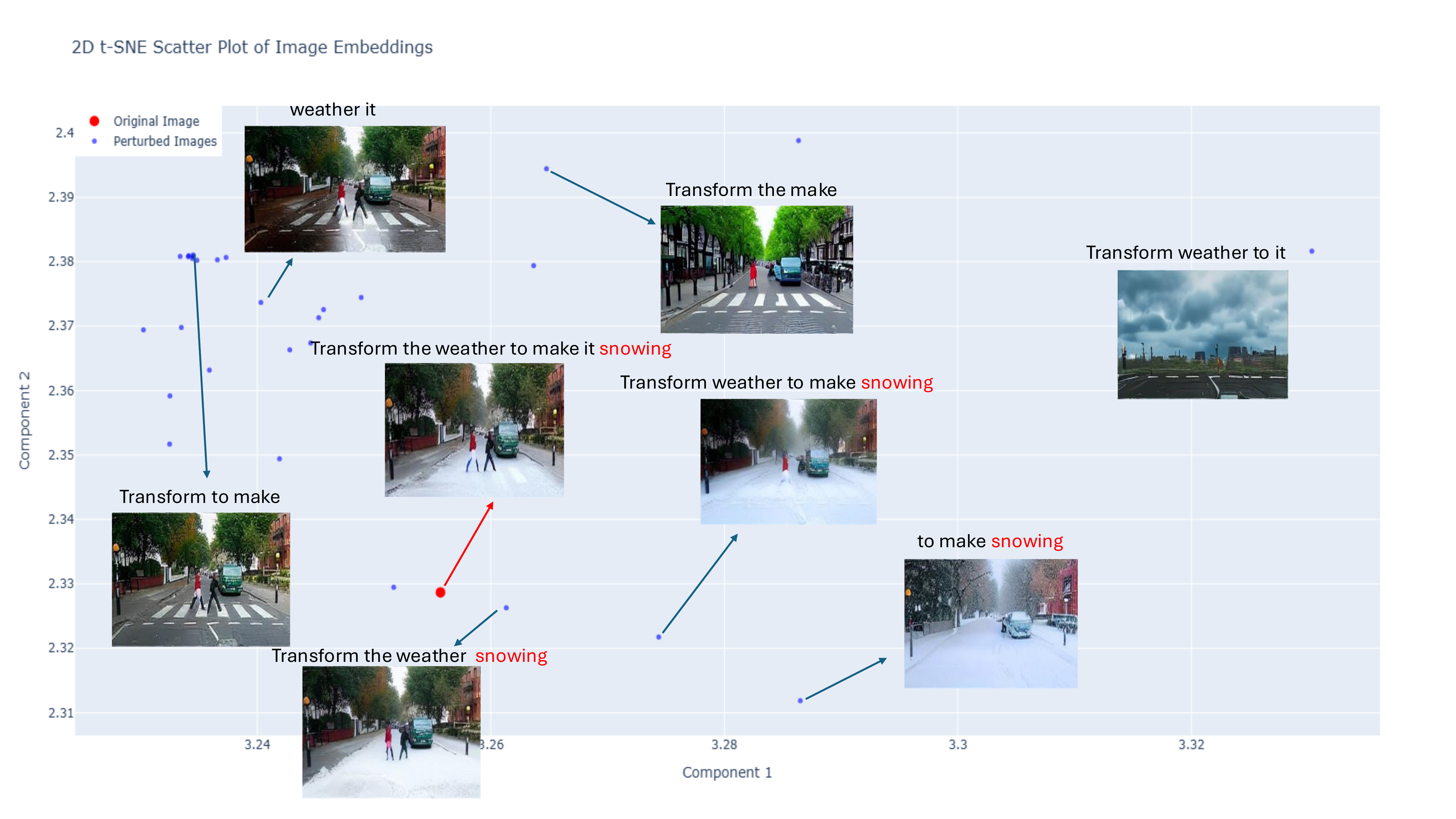}
    \caption{2D visualisation of image embeddings, showing the effect of including or omitting ``snowing'' in the prompt.}
    \label{fig:TSNE1}
\end{figure}

The 2D plot shown in Fig.~\ref{fig:TSNE1} illustrates the proximity or distance between images based on their visual features. Perturbations containing the keyword ``snowing'' form clusters in the snowy regions closer to the original image. In contrast, other perturbations result in non-snowy scenes. This visualisation allows us to qualitatively assess the impact of each word in the prompt and how the model responds to different perturbations.

\subsection{Quantitative Results}

In this section, we apply evaluation metrics, including accuracy, stability, consistency, fidelity, and computational complexity, to assess our explainability method across various image editing models and scenarios.

\subsubsection{ATT Accuracy}

To evaluate the performance of instruction-based image editing models, we tested ten different prompts on a single image and ten different images with a single instruction, each undergoing 30 perturbations. Heatmaps were extracted to display the weights of each keyword in the instructions across various models, including Instruct-Pix2Pix, Img2Img-Turbo, and Diffusers-inpaint. A ground truth was defined, assigning weights of 1 to keywords in the instructions and 0 to other words. These ground truth values were compared with the extracted heatmaps to assess the model's performance.

To quantify accuracy, we employed multiple metrics, including Attention Accuracy (ATT ACC)~\ref{eq:att_acc}, F1-Score~\ref{eq:att_f1} for Attention (ATT F1), and Area Under the Receiver Operating Characteristic Curve for Attention (ATT AUROC)~\ref{eq:att_auroc}. The results were averaged across the ten different prompts and images, and each model's outcomes are summarised in Table~\ref{tab:accuracy_models}, showcasing their performance based on these metrics.

\begin{table}[H]
\centering
\caption{Performance metrics (ATT ACC, ATT F1, and ATT AUROC) for various\\models, evaluated under two conditions: different prompts and different images.}
\label{tab:accuracy_models}
\renewcommand{\arraystretch}{1.3}
\scalebox{0.8}{
\begin{tabular}{lccc}
\hline \hline
\textbf{Model name} & \textbf{ATT ACC} & \textbf{ATT F1} & \textbf{ATT AUROC} \\ \hline
I-Pix2Pix (different prompts)   & 0.7845 & 0.6869 & 0.8943 \\ 
\textbf{I-Pix2Pix (different images)}    & \textbf{0.9571} & \textbf{0.9166} & \textbf{1.000} \\ 
Diffusers\_I (different prompts) & 0.6621 & 0.1167 & 0.4417 \\
Diffusers\_I (different images)  & 0.7714 & 0.1000 & 0.4000 \\
I2I-Turbo (different prompts)   & 0.7944 & 0.6824 & 0.8400 \\  
I2I-Turbo (different images)    & 0.8286 & 0.6333 & \textbf{1.000} \\ 
\hline \hline
\end{tabular}
}
\end{table}

\subsubsection{ATT Stability}

To quantify stability, we use the Jaccard index, a metric that measures the similarity between sets by comparing their overlap. We modify an image with ten different prompts for image editing models, such as Instruct-Pix2Pix, Img2Img-Turbo, and Diffusers-inpaint, extracting the coefficients and weights of each word. Then, by adding `\#\#\#` at the end of each prompt as input text, we extract the coefficients again and compute the Jaccard index to compare them. Finally, we calculate the average metric for each model and present the results in Table~\ref{tab:stability_models}.

\begin{table}[H]
\centering
\caption{Stability across different models for ten prompts and 30 perturbations. The table presents the Jaccard Index for different models.}
\label{tab:stability_models}
\renewcommand{\arraystretch}{1.3}
\scalebox{0.8}{
\begin{tabular}{lc}
\hline \hline
\textbf{Model Name}  & \textbf{Jaccard Index}  \\ \hline
I-Pix2Pix   & 0.85 \\ 
Diffusers\_I  & 0.85 \\ 
I2I-Turbo   & 0.85 \\ 
\hline \hline
\end{tabular}
}
\end{table}

\subsubsection{ATT Consistency}

For computational consistency, we ran the same code with 30 perturbations for each one over 1000 iterations, ensuring that the outputs remained consistent across these repetitions for image editing models such as Instruct-Pix2Pix, Img2Img-Turbo, and Diffusers-inpaint, as shown in Table~\ref {tab:consistency_models}. We computed variance and standard deviation metrics for each word coefficient. This step was crucial to confirm that the model's predictions were not random or overly dependent on initialisation factors but rather deterministic and robust. Lower variance and standard deviation values indicate better consistency.

\begin{table}[H]
\centering
\caption{Consistency metrics for different diffusion models. Variance and standard deviation calculated for the prompt: ``Transform the weather to make it snowing.''}
\label{tab:consistency_models}
\renewcommand{\arraystretch}{1.3}
\scalebox{0.8}{
\begin{tabular}{lcc}
\hline \hline
\textbf{Model Name}  & \textbf{Variance} & \textbf{Standard Deviation} \\ \hline
I-Pix2Pix   & 0.0161 & 0.1271 \\ 
Diffusers\_I  & 0.0776 & \textbf{0.0060} \\ 
I2I-Turbo   & \textbf{0.0001} & 0.0081 \\ 
\hline \hline
\end{tabular}
}
\end{table}

\subsubsection{ATT Fidelity for Different Instruct image editing diffusion-based Models (IED)}

The fidelity computation for various IED models applied to the text prompt ``Transform the weather to make it snowing'' is displayed below. We compare several models, including Instruct-Pix2Pix, Img2Img-Turbo, and Diffusers-inpaint, using 64 perturbations. The comparisons are made using a weighted linear regression as the surrogate model and the Wasserstein distance to compute distances across metrics such as MSE, (R$^2_\omega$), MAE, mean \(L_1\), and mean \(L_2\) losses. The results are presented in Table~\ref{tab:fidelity_models}.

As shown in table~\ref{tab:fidelity_models}, the fidelity for the Diffusers-inpaint model was calculated, but its value is relatively low due to its poor performance. For this reason, fidelity was not calculated for the other scenarios involving this model.

\begin{table}[H]
\centering
\caption{Fidelity metrics for different diffusion models. The fidelity metrics were calculated for the prompt: ``Transform the weather to make it snowing.''}
\label{tab:fidelity_models}
\renewcommand{\arraystretch}{1.3}
\scalebox{0.8}{
\begin{tabular}{lcccccc}
\hline \hline
\textbf{Model Name} & \textbf{WMSE} & \textbf{R$^2_\omega$} & \textbf{WMAE} & \textbf{Mean-L1} & \textbf{Mean-L2} & \textbf{R$^2_{\hat{\omega}}$} \\ \hline
I-Pix2Pix   & 0.0120 & \textbf{0.7208} & 0.0812 & 0.0790 & 0.0124 & \textbf{0.6859} \\ 
Diffusers\_I  & 0.0317 & 0.1255 & 0.1359 & 0.1529 & 0.0410 & 0.0161  \\ 
I2I-Turbo   & 0.0193 & 0.6225 & 0.1078 & 0.1076 & 0.0187 & 0.5753 \\ 
\hline \hline
\end{tabular}
}
\end{table}

\subsubsection{ATT Fidelity Across Different Numbers of Text Perturbations}

Tables~\ref{tab:fidelity_per}, ~\ref{tab:fidelity_per_turbo} show the Results of fidelity computation for different numbers of perturbations using the Instruct-Pix2Pix and Img2Img-Turbo methods. The comparisons are made using a weighted linear regression as the surrogate model and the Wasserstein distance to compute distances, using metrics such as weighted mean squared error (WMSE), coefficient of determination (R$^2_\omega$), weighted mean absolute error (WMAE),  loss metrics of the (mean \(L_1\) and mean \(L_2\)):

The results presented in these tables highlight the robustness and accuracy of the evaluated methods in preserving fidelity across various perturbation intensities. A lower (WMSE) and (WMAE), along with a higher (R$^2_\omega$), indicate better alignment between the generated and target outputs.

\begin{table}[H]
\centering
\caption{Performance metrics for different numbers of perturbations in Instruct-Pix2Pix. Evaluation metrics include WMSE, R$^2_\omega$, WMAE, mean-L1, mean-L2, and R$^2_{\hat{\omega}}$.}
\label{tab:fidelity_per}
\renewcommand{\arraystretch}{1.3}
\scalebox{0.8}{
\begin{tabular}{rcccccc}
\hline \hline
\textbf{\#Perturb} & \textbf{WMSE} & \textbf{R$^2_\omega$} & \textbf{WMAE} & \textbf{Mean-L1} & \textbf{Mean-L2} & \textbf{R$^2_{\hat{\omega}}$} \\ \hline
32  & 0.0178 & 0.7770 & 0.1040 & 0.1004 & 0.0161 & 0.7119 \\ 
64  & 0.0120 & 0.7208 & 0.0812 & 0.0790 & 0.0124 & 0.6859 \\ 
128 & 0.0220 & 0.6816 & 0.1129 & 0.1090 & 0.0207 & 0.6631 \\ 
256 & 0.0080 & 0.5110 & 0.0512 & 0.0509 & 0.0084 & 0.4972 \\ 
\hline \hline
\end{tabular}
}
\end{table}

\begin{table}[H]
\centering
\caption{Performance metrics for different numbers of perturbations in Img2Img-Turbo. Evaluation metrics include WMSE, R$^2_\omega$, WMAE, mean-L1, mean-L2, and R$^2_{\hat{\omega}}$.}
\label{tab:fidelity_per_turbo}
\renewcommand{\arraystretch}{1.3}
\scalebox{0.8}{
\begin{tabular}{rcccccc}
\hline \hline
\textbf{\#Perturb} & \textbf{WMSE} & \textbf{R$^2_\omega$} & \textbf{WMAE} & \textbf{Mean-L1} & \textbf{Mean-L2} & \textbf{R$^2_{\hat{\omega}}$} \\ \hline
32  & 0.0421 & 0.5273 & 0.1635 & 0.1484 & 0.0369 & 0.3894 \\ 
64  & 0.0193 & 0.6225 & 0.1078 & 0.1076 & 0.0187 & 0.5753 \\ 
128 & 0.0316 & 0.3879 & 0.1412 & 0.1330 & 0.0293 & 0.3522 \\ 
256 & 0.0248 & 0.6025 & 0.1275 & 0.1375 & 0.0305 & 0.5912 \\ 
\hline \hline
\end{tabular}
}
\end{table}

\subsubsection{ATT Fidelity for Different Distance Metrics and Surrogate Models}

Fidelity is also computed by comparing various distance metrics for the Instruct-Pix2Pix model with text-versus-text and image-versus-image comparisons. The surrogate models used are Weighted Linear Regression and Bayesian Ridge (BayLIME), as shown in Table~\ref{tab:fidelity_dist}. We explore different combinations of distance metrics, such as Cosine and Wasserstein Distance (WD). Fidelity is measured using MSE, (R$^2_\omega$), MAE, mean \(L_1\), and mean \(L_2\) losses.

\begin{table}[H]
\centering
\caption{Fidelity results for different distance measures with 30 perturbations. The table presents various fidelity metrics for different combinations of T vs T and I vs I distance measures.}
\label{tab:fidelity_dist}
\renewcommand{\arraystretch}{1.3}
\scalebox{0.75}{
\begin{tabular}{llcccccc}
\hline \hline
\multicolumn{2}{c}{\textbf{WLR}} & \multicolumn{6}{c}{\textbf{Fidelity Metrics}} \\
\cline{1-2} \cline{3-8}
\textbf{T vs T} & \textbf{I vs I} & \textbf{WMSE} & \textbf{R$^2_\omega$} & \textbf{WMAE} & \textbf{Mean-L1} & \textbf{Mean-L2} & \textbf{R$^2_{\hat{\omega}}$} \\
\hline
Cosine & Cosine & 0.0494 & 0.7626 & 0.1555 & 0.2865 & 0.2309 & 0.6167 \\
Cosine & WD     & 0.0022 & 0.8557 & 0.0215 & 0.2530 & 0.1148 & 0.7980 \\
WD     & WD     & 0.0128 & \textbf{0.9495} & 0.0674 & 0.2865 & 0.5243 & \textbf{0.9292} \\
WD     & Cosine & \textbf{0.0022} & 0.8558 & \textbf{0.0216} & 0.2530 & 0.1149 & 0.8558 \\
WD+C   & WD+C   & 0.0411 & 0.8744 & 0.1427 & 0.8484 & 1.5264 & 0.8241 \\
\hline
\multicolumn{8}{c}{\textbf{BayLIME}} \\
\hline
Cosine & Cosine & 0.0524 & 0.7097 & 0.1862 & 0.2607 & 0.1363 & 0.5936 \\
Cosine & WD     & 0.2314 & 0.6064 & 0.3942 & 0.5239 & 0.4739 & 0.6064 \\
WD     & WD     & 0.0148 & \textbf{0.9485} & 0.0674 & 0.8065 & 0.8052 & \textbf{0.9278} \\
WD     & Cosine & \textbf{0.0001} & 0.8369 & \textbf{0.0049} & \textbf{0.0501} & \textbf{0.0044} & 0.7715 \\
WD+C   & WD+C   & 0.0421 & 0.8711 & 0.1139 & 0.8282 & 1.4246 & 0.8195 \\
\hline \hline
\end{tabular}
}
\end{table}

\subsubsection{Computation complexity}

We apply our model to three types of instruction-based image editing frameworks: Instruct-pix2pix, Img2Img-Turbo, and Diffusers-inpaint. We measure the time consumed across different explainability methods, including LIME, gSMILE, and Bay-LIME, with 60 perturbations performed on the same computer to ensure consistent hardware conditions.

The models exhibit apparent differences in their computational demands, primarily due to the way they calculate similarity and assign weights. gSMILE, in particular, uses the Wasserstein distance to calculate text distances, making it much more computationally expensive than LIME and Bay-LIME. As shown in Table~\ref{tab:execution_times}, the execution times for gSMILE are consistently higher across all three frameworks. Similar trends are observed in the other frameworks, where gSMILE's computational overhead is evident.

Wasserstein distance is excellent for capturing detailed differences between distributions, especially in text embeddings, but it comes at a cost \cite{peyre2019computational}. For high-dimensional text data, computing the Wasserstein-2 distance involves working with covariance matrices, which requires time proportional to $O(d^3)$, where $d$ is the embedding size. When multiple samples ($N$) are considered, the computational complexity quickly scales to $O(Nd^3)$. In contrast, LIME employs more straightforward similarity measures, such as cosine similarity, which scale more efficiently at $O(Nd)$.

The complexity difference arises because Wasserstein-2 distance calculations involve operations on covariance matrices for $d$-dimensional Gaussians~\cite{cuturi2014fast}. While this higher computational effort can yield more robust and stable explanations, as seen in gSMILE, it significantly increases execution time. On the other hand, LIME's reliance on cosine similarity is computationally lighter and faster, as it avoids such matrix operations.

gSMILE requires more computational resources and is inherently more complex, but delivers highly detailed and stable explanations. This makes it particularly well-suited for tasks where interpretability and precision are critical despite its extra computational costs.

\begin{table}[H]
\centering
\caption{Execution times (in seconds) for each framework and explainability method with 60 perturbations.}
\label{tab:execution_times}
\renewcommand{\arraystretch}{1.3}
\scalebox{0.8}{
\begin{tabular}{lccc}
\hline \hline
\textbf{Framework} & \textbf{LIME} & \textbf{gSMILE} & \textbf{Bay-LIME} \\ \hline
I-Pix2Pix         & 1013.35  & \textbf{1208.08} & 1024.32 \\ 
I2I-Turbo         & 983.17   & \textbf{1222.09} & 980.12  \\
Diffuser-Inpaint  & 263.00   & \textbf{276.83}  & 260.75  \\ 
\hline \hline
\end{tabular}
}
\end{table}

\subsubsection{Influence of Linguistic Elements on Explainability}

We designed an experiment using a structured set of prompts to understand the impact of different phrasing on the model's ability to interpret and execute image editing instructions. We aim to evaluate how variations in wording affect the model's performance and the clarity of the resulting edits. We selected ten diverse images and applied prompts such as ``Transform the image to depict a cloudy, dark evening, ``Transform the weather to snowy,'' and ``Turn the white-skinned person into a black-skinned person.'' For each control word in the prompts, we created variations using verbs, adjectives, nouns, ``ing'' forms, and rephrasing with ``make'' and ``must.'' Each image was processed using all prompt variations, and the outputs were analysed using LIME to generate heatmaps showing the importance of each word.

\begin{figure}[H]
    \centering
    \includegraphics[width=1\linewidth]{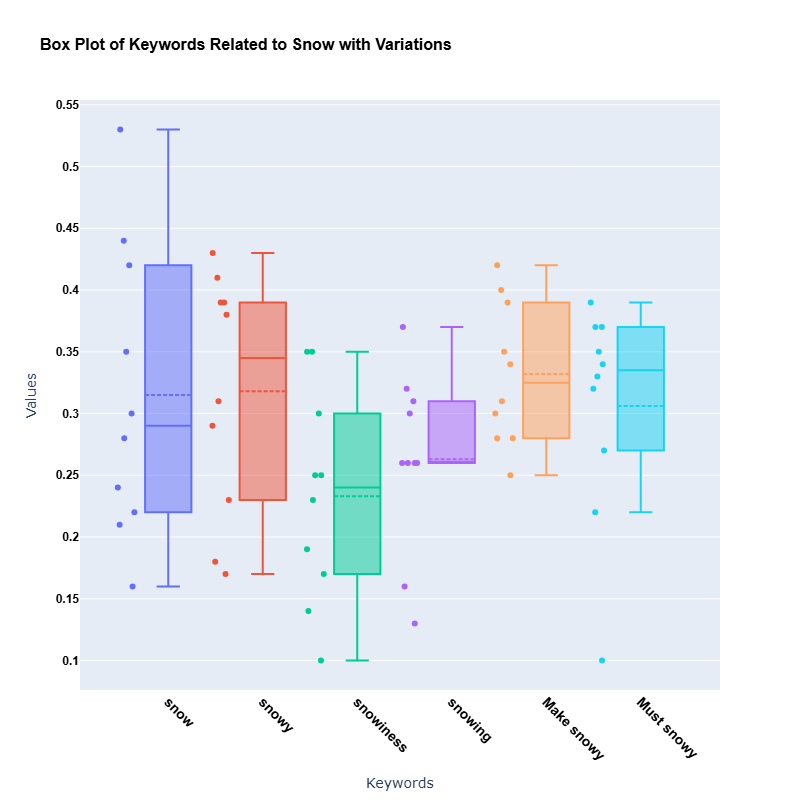}
    \caption{Effect of linguistic variations on responses to ``snow''\\ prompts. Different forms (e.g., nouns, adjectives, gerunds) yield distinct output distributions.}
    \label{fig:fig_LE_S}
\end{figure}

Fig.~\ref{fig:fig_LE_S}, the snow, related prompts, reveals that noun and adjective forms lead to more consistent outputs, while gerund and directive phrases exhibit higher variability. This suggests that the model interprets straightforward descriptors more predictably, while transformation directives introduce flexibility and ambiguity in execution.

The results from this study provide valuable insights for users aiming to achieve specific image transformations using AI models. By carefully selecting the phrasing of their prompts, users can exert greater control over the consistency and quality of the generated edits. Future work will expand the prompt set to include more complex instructions and evaluate how contextual information affects the model's decision-making process.

\begin{figure}[H]
    \centering
    \includegraphics[width=1\linewidth]{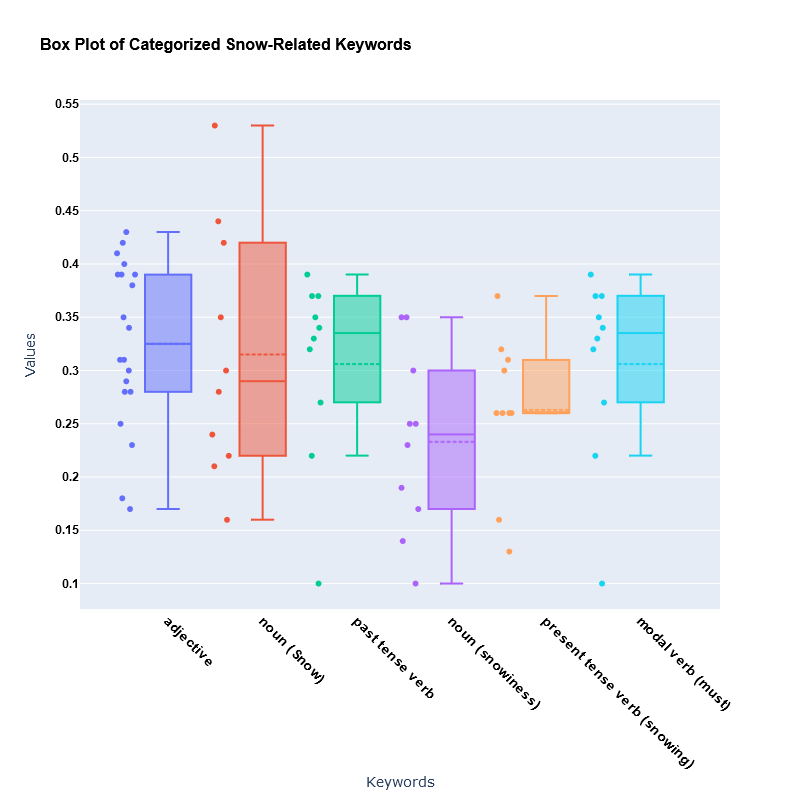}
    \caption{Response variations for a range of prompts across different words. This illustrates the model's general sensitivity to linguistic phrasing in instruction-based image editing.}
    \label{fig:fig_LE_G}
\end{figure}

Fig.~\ref{fig:fig_LE_G} illustrates the model's response to a range of prompts that incorporate various linguistic elements, offering a broader perspective on its sensitivity to phrasing. Testing directive phrases like ``make it'' reveals wider response ranges, suggesting that the model's interpretation becomes more context-dependent with complex instructions.

\begin{figure}[H]
    \centering
    \includegraphics[width=1\linewidth]{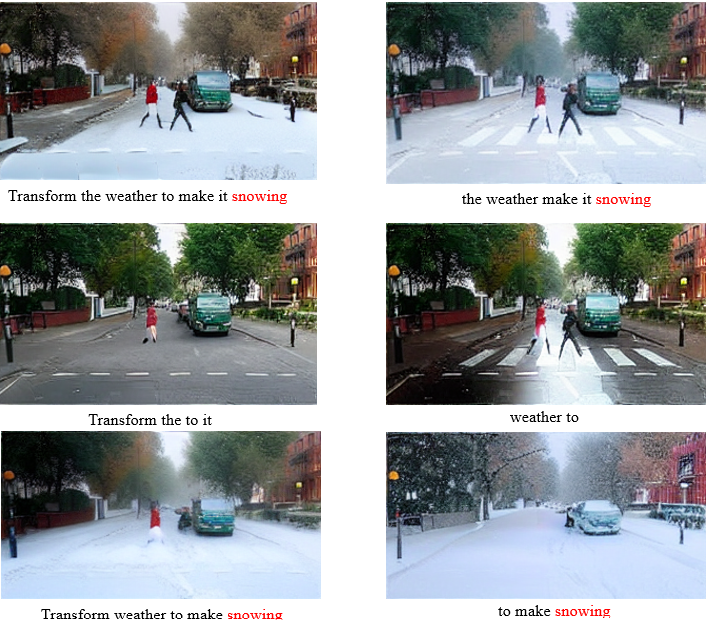}
    \caption{Results generated by Instruct-Pix2Pix showing how the presence of the keyword ``snowing'' in the perturbed prompts leads to snowy scenes, while its absence results in non-snowy images, illustrating the model's responsiveness to specific keywords.}
    \label{fig:Snowing_per}
\end{figure}

In our model, we refined input processing by splitting the text into individual words and creating targeted perturbations based on these words. This approach enhances granularity in understanding how each keyword influences the output. Fig.~\ref{fig:Snowing_per} illustrates how different models respond to prompts containing specific keywords. For instance, with the keyword “snowing” included, the generated image accurately depicts a snowy scene. However, when the keyword is omitted, the image reflects alternative visual outcomes. This highlights the model’s sensitivity to precise textual cues and the importance of keyword selection in achieving desired results.

This method enables us to assess the direct impact of individual keywords on image generation, revealing how each word shapes the output. By identifying and quantifying these effects, we can refine the model to enhance consistency and accuracy in generating context-specific visuals. By examining these effects, we can further optimize the model to ensure that subtle changes in textual input lead to precise and predictable alterations in the generated images. This methodology is a key step toward achieving robust and reliable AI-generated content.

\begin{figure}[H]
    \centering
    \includegraphics[width=1\linewidth]{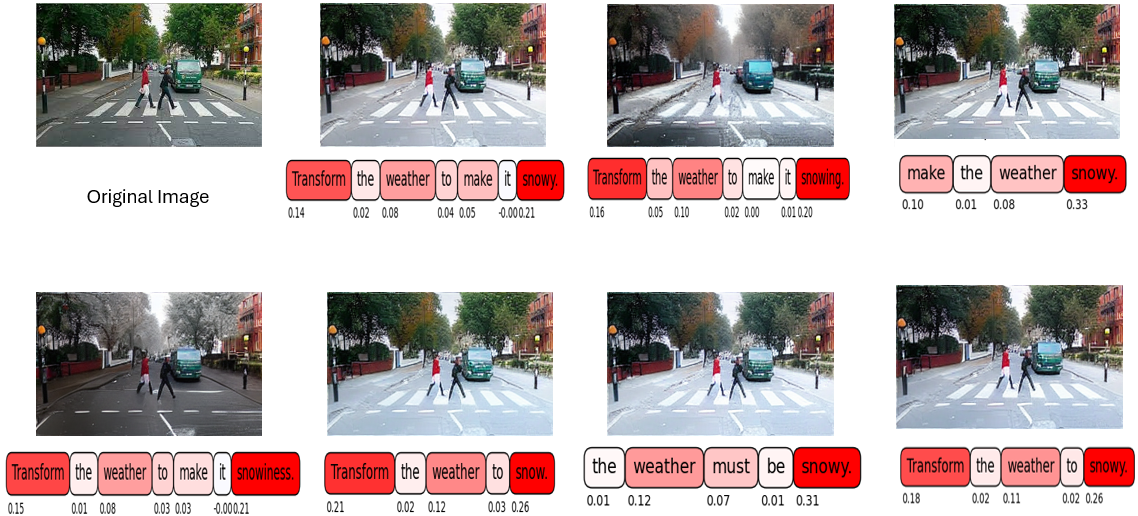}
    \caption{Comparison of different structures of prompts used to generate the same image using Instruct-Pix2Pix, showcasing how variations in the prompt's structure lead to distinct transformations of the same base scenario.}
    \label{fig:Snow_str}
\end{figure}

Fig.~\ref{fig:Snow_str} shows how an image changes when different prompt structures are used. This visualization underscores the pivotal role of prompt design in guiding AI-generated outputs. By tweaking keywords and phrases like “make it snowy” or “the weather must be snowy,” we see noticeable differences in how the image looks. Such variations illustrate the importance of nuanced prompt crafting to achieve specific visual goals. The importance of each word in the prompts is reflected by its weight, with darker colors representing words that have a more substantial influence. This weighted representation provides valuable insights into the linguistic components that drive visual transformations.

This analysis demonstrates the model's sensitivity to specific prompt elements and highlights how prompt phrasing directly impacts the generated image's characteristics. By leveraging this understanding, users can refine their prompts to produce more accurate and contextually relevant outputs, making this approach a powerful tool for optimizing AI-driven image generation.

\begin{figure}[H]
    \centering
    \includegraphics[width=1\linewidth]{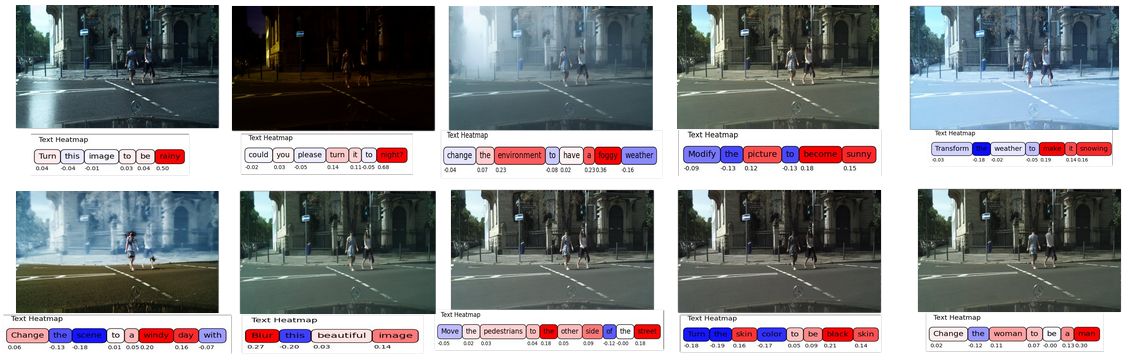}
    \caption{Heatmap analysis of prompts displaying the weights of individual words in guiding the Instruct-Pix2Pix model’s visual output. Each row shows transformations generated from different prompt structures, with heatmaps illustrating how specific words affect the model’s interpretation and the resulting image.}
    \label{fig:diff_prompt}
\end{figure}

Fig.~\ref{fig:diff_prompt} presents the heatmap analysis of prompt words to illustrate the influence of each word on the generated images using the Instruct-Pix2Pix model. This analysis provides a visual breakdown of how the model decodes and assigns importance to textual inputs. Each image corresponds to a specific prompt, with heatmaps showing the weight assigned to each word. Darker colors represent words with higher influence, indicating their more substantial impact on the resulting image transformation. This allows users to pinpoint which words are most critical for achieving desired visual outcomes.

This analysis highlights how the model interprets and prioritizes individual words in the prompt, allowing us to observe the relationship between textual emphasis and visual modifications. Such insights are invaluable for refining prompt engineering strategies, ensuring more precise and predictable image transformations. By understanding these dynamics, we can further optimize the Instruct-Pix2Pix model for enhanced sensitivity to nuanced textual cues, paving the way for improved AI-generated content.

\begin{figure}[H]
    \centering
    \includegraphics[width=1\linewidth]{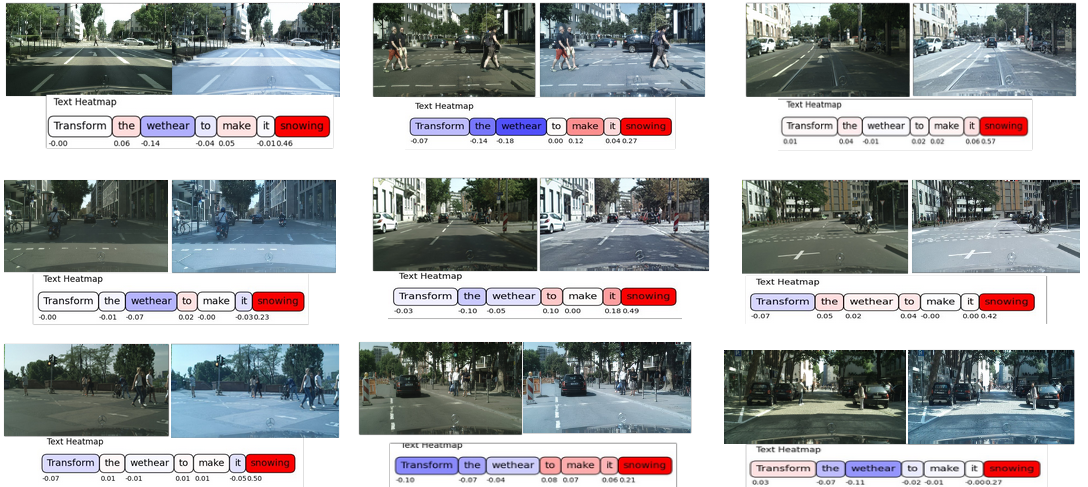}
    \caption{ Heatmap analysis of a single prompt applied across different images. Each heatmap shows the weight of each word in the prompt, indicating how the Instruct-Pix2Pix model maintains consistent word influence across varied scenes to achieve the desired transformations.}
    \label{fig:diff_images}
\end{figure}

\begin{figure}[H]
    \centering
    \includegraphics[width=1\linewidth]{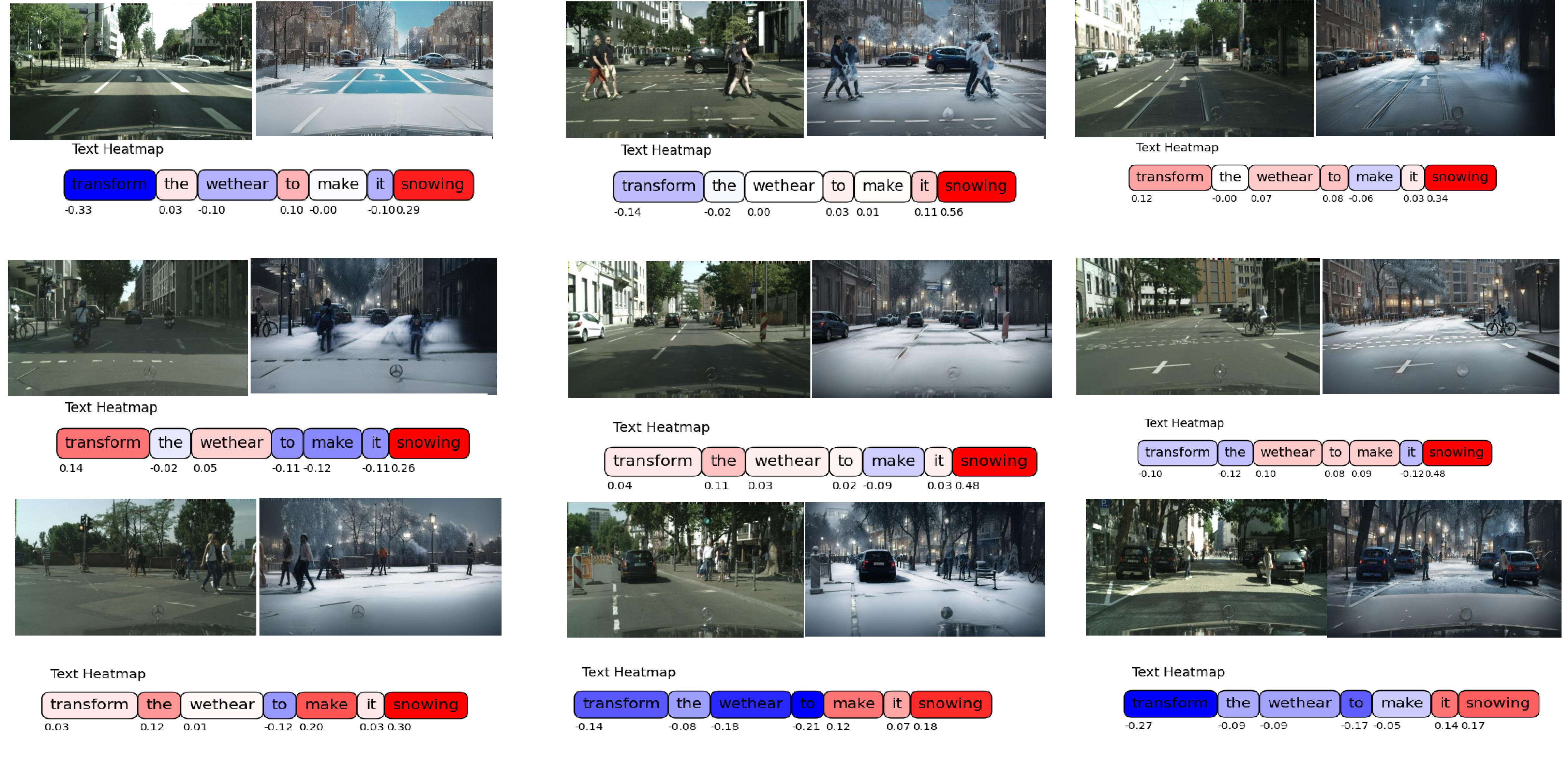}
    \caption{ Heatmap analysis of a single prompt applied across different images. Each heatmap shows the weight of each word in the prompt, indicating how the I2I-Turbo model maintains consistent word influence across varied scenes to achieve the desired transformations.}
    \label{fig:diff_images_turbo}
\end{figure}

Figs.~\ref{fig:diff_images} and~\ref{fig:diff_images_turbo} illustrates the consistency of word influence in a single prompt applied across different images. This demonstrates the robustness of the models in maintaining predictable word-driven transformations across varying scenarios. The heatmaps display the weights assigned to each word in the prompt, with darker colors signifying higher influence. Such visual representation helps in understanding how individual words dominate the transformation process. The keyword “snowing” has a strong impact, resulting in reliable visual changes that align with the modification. This highlights the effectiveness of the keyword in directing the model’s focus toward the desired weather condition.

This analysis shows how the Instruct-Pix2Pix and I2I-Turbo models interpret the same prompt across various contexts.


\section{Conclusion}

This work extended the \textbf{gSMILE} framework to instruction-based image editing and examined how textual structure, input image content, and perturbation strategies influence both model behaviour and explainability. Focusing on three state-of-the-art diffusion-based editing models, Instruct-Pix2Pix, Img2Img-Turbo, and Diffusers-Inpaint, we evaluated the effectiveness of gSMILE using a broad set of qualitative and quantitative metrics, including attention-based accuracy, stability, consistency, fidelity, and computational cost.

Qualitative evaluations grounded in the Operational Design Domain (ODD) demonstrated that gSMILE can precisely capture the relationship between textual tokens and visual changes in edited outputs. Diverse scenarios were generated, including environmental conditions (rain, fog, snow), temporal variations (day/night), and human- or object-centric factors. Heatmaps and t-SNE analyses revealed that keywords such as "snowing" consistently exert the most decisive influence on the resulting transformations, with the presence or absence of these words directly determining the nature of the generated edits. The clustering behaviour observed in perturbed prompt embeddings further confirmed that models produce coherent and interpretable changes in response to token-level modifications.

Results also showed that linguistic formulation significantly affects interpretability. Variations in prompt structure (nouns, adjectives, gerunds, and directive forms such as ``make'' and ``must'') revealed that simple descriptive forms generally yield more stable and predictable outputs. In contrast, more complex grammatical constructions introduce higher variability. Heatmap analyses across multiple prompts and images further confirmed that only a small set of keywords typically dominate the attribution space, underscoring the importance of careful prompt engineering for controlled and interpretable editing.

Quantitatively, gSMILE demonstrated strong performance as a model-agnostic local explainer for instruction-based image editing:

\begin{itemize}
    \item \textbf{Instruct-Pix2Pix} achieved the highest accuracy and attribution fidelity. In the ``different images'' setting, it achieved ATT ACC = 0.9571, ATT F1 = 0.9166, and ATT AUROC = 1.0, with strong fidelity metrics (low WMSE and high $R^2_\omega$), indicating that the weighted linear surrogate closely approximates the model’s behaviour.
    \item \textbf{Img2Img-Turbo} yielded competitive accuracy and, importantly, \emph{superior computational consistency}, achieving the lowest variance and standard deviation across repeated runs, demonstrating stable and reproducible attributions.
    \item \textbf{Diffusers-Inpaint} showed weaker performance in both fidelity and accuracy, revealing that certain architectures are intrinsically more challenging to approximate using text-driven perturbations and may require more specialised perturbation strategies.
\end{itemize}

Overall stability was high across all models (average Jaccard index $\approx 0.85$), and fidelity remained robust across multiple perturbation levels. Comparisons with LIME and Bay-LIME showed that the combination of Wasserstein distance and weighted linear regression generally produced the most faithful approximations of model behaviour, particularly for text--image alignment. However, these advantages come at a computational cost: the use of Wasserstein distance on high-dimensional textual embeddings significantly increased execution time over LIME and Bay-LIME, indicating an inherent trade-off between interpretative depth and computational efficiency.

Taken together, these findings demonstrate that \textbf{gSMILE offers accurate, stable, and visually intuitive explanations for instruction-based image editing models}, thereby complementing its prior effectiveness on large language models. gSMILE demonstrated that even without access to internal parameters, it is possible to recover meaningful relationships between textual inputs and visual outputs. This work directly answers \textbf{Research Question 3 (RQ3)} by showing that model-agnostic, perturbation-based explainability can be successfully and reliably extended to multimodal generative systems, enabling practical transparency in instruction-controlled image editing.

At the same time, several promising directions emerge:

\begin{enumerate}
    \item \textbf{Adaptive perturbation strategies:} Future work may replace uniform random perturbations with targeted linguistic perturbations focused exclusively on influential tokens. This would reduce computational cost while preserving fidelity, especially for long or complex prompts.

    \item \textbf{Causal verification for text-image alignment:} To mitigate spurious or hallucination-driven attributions, causal analysis (e.g., altering individual tokens while controlling image context) could be integrated to verify whether a word genuinely drives a visual change, improving reliability in safety-critical applications.

    \item \textbf{Extension to multimodal and temporal generative models:} While this work focused on static image editing, extending gSMILE to video diffusion models or higher-order multimodal systems (text--image--video) would allow evaluation of temporal consistency and broaden the generalisability of the framework.
\end{enumerate}

By combining rigorous evaluation with interpretable visual explanations, gSMILE increases transparency in instruction-based image editing without requiring privileged access to internal model mechanisms. This provides both researchers and practitioners with a practical and trustworthy tool for analysing and controlling modern generative visual systems.

\chapter{Conclusion}
\label{chap:Conclusion}

Interpretability is increasingly recognised as a fundamental requirement for the safe, transparent, and accountable deployment of modern machine learning and deep learning systems. As generative AI models grow in scale, complexity, and real-world impact, understanding how they behave and why they produce particular outputs has become essential for fostering user trust and ensuring responsible use. This thesis has demonstrated that meaningful human interaction with AI systems depends on the availability of clear and faithful explanations. Without transparency in their decision-making processes, users may hesitate to adopt generative models, particularly in high-stakes domains such as healthcare, law, and autonomous driving.

To address these challenges, this research introduced \textbf{gSMILE (Statistical Model-agnostic Interpretability with Local Explanations for Generative Models)}, a unified and model-agnostic interpretability framework for both textual and multimodal generative systems. Unlike many existing explainability tools that depend on internal attention maps or access to model layers, gSMILE operates as a purely \textit{black-box} method, requiring no access to gradients, parameters, or architectural details. This makes it suitable for explaining proprietary or API-based models, which now dominate the landscape of generative AI. By integrating statistical distance measures with perturbation-based sampling and weighted local surrogate models, gSMILE directly addresses key limitations of traditional post-hoc interpretability techniques, such as instability, vulnerability to input perturbations, and difficulties in handling multimodal output spaces. Its effectiveness was validated through rigorous quantitative and qualitative evaluations, using attribution metrics such as accuracy, fidelity, consistency, and stability.

This thesis successfully addressed the three research questions posed at the outset:

\textbf{RQ1} was answered by developing and applying a comprehensive suite of interpretability metrics capable of quantitatively evaluating explanation quality across both text and image domains (Chapter~\ref{chap:gSMILE Methodology}). These metrics formed the foundation for systematically assessing the reliability of gSMILE across diverse generative models.

\textbf{RQ2} was addressed by demonstrating how gSMILE provides fine-grained, token-level explanations for large language models, revealing how individual prompt components influence generated text and enabling transparent, human-aligned reasoning (Chapter~\ref{chap:llm_explainability}).

\textbf{RQ3} was addressed by adapting gSMILE to instruction-based image editing tasks, where the generated heatmaps served as clear, human-interpretable explanations. These visual attributions illustrated precisely how each word in the instruction influenced the corresponding image modifications and which regions of the image were affected by specific textual components (Chapter~\ref{chap:image_editing_explainability}). This enabled an intuitive, visual understanding of model behaviour for end-users.

Overall, this work provides a unified and practical pathway for interpreting generative models across modalities, contributing both methodological innovation and empirical insights into the behaviour of modern AI systems.

\section{Summary of Contributions}

This thesis makes several key contributions to the field of interpretability for generative AI systems:

\begin{itemize}

    \item \textbf{Introducing a Unified Interpretability Framework:}  
    We present gSMILE, a novel model-agnostic framework that generates human-interpretable explanations for both textual and multimodal generative models. By visualising the influence of individual input elements, such as words in prompts, on model outputs, gSMILE provides clear and actionable insights into the behaviour of large language models (LLMs) and instruction-based image editing systems.

    \item \textbf{Advancing Transparency Through Visual and Textual Explanations:}  
    gSMILE explicitly links user instructions to model behaviour, producing token-level attribution maps for text generation and heatmaps that visually connect words to image modifications. These explanations enhance transparency, promote user trust, and support the safe deployment of systems in sensitive domains, such as healthcare, education, autonomous driving, and legal decision-making.

    \item \textbf{A Rigorous Evaluation Protocol for Interpretability and Robustness:}  
    This work introduces a comprehensive evaluation methodology that assesses interpretability using stability, fidelity, accuracy, and consistency metrics across multiple generative models. Through extensive experiments, we demonstrate that gSMILE offers reliable, repeatable, and high-fidelity explanations, establishing it as an effective tool for auditing, understanding, and improving the behaviour of modern generative AI systems.

\end{itemize}

\section{Limitations}

While gSMILE provides a flexible, model-agnostic framework for interpreting both large language models (LLMs) and instruction-based image editing systems, several limitations arise from its methodological design choices and the inherent complexity of generative models.

First, the perturbation strategy used in this work relies primarily on binary removal–retention (0/1) of tokens. Although effective for analysing local sensitivity, this approach does not capture more nuanced linguistic transformations. Removing tokens can distort sentence meaning, violate syntactic structure, or cause the model to operate outside its natural range. More expressive perturbations, such as synonym substitution, paraphrasing, or Gaussian noise applied to the embedding space, were not explored and may reveal richer attribution patterns.

Second, token-level perturbations can produce disproportionately large shifts in output, even when the semantic content of the prompt is mostly preserved. This sensitivity undermines the assumption of local smoothness required by regression-based surrogate models, which can reduce the stability of explanations and interpretive coherence.

Third, generative models, particularly LLMs, are prone to hallucination when provided with ambiguous, incomplete, or degraded prompts. These hallucinated outputs introduce noise into the attribution process, potentially compromising the reliability of explanations generated using local perturbations, especially for models that are weakly aligned or highly sensitive to input variations.

Fourth, the framework’s computational requirements are substantial. gSMILE generates and evaluates a large number of perturbations, and the use of Wasserstein distance for weighting samples further increases computational cost. For API-based models, this results in higher inference costs and longer overall runtimes. For locally deployed models, particularly those involving vision–language pipelines, significant GPU resources are required, limiting accessibility for users without high-end hardware.

Fifth, the surrogate models examined in this thesis were limited to weighted linear regression and Bayesian ridge regression. While effective under local linearity assumptions, these surrogates may be insufficient for highly non-linear generative systems such as diffusion models. More expressive surrogate models were not considered.

Finally, practical deployment is constrained by model accessibility. Some LLMs and multimodal systems do not provide API access and require manual installation or local fine-tuning, increasing the operational complexity of reproducing experiments or scaling gSMILE to broader applications.

Despite these limitations, gSMILE represents a promising direction for improving the interpretability of generative models. Addressing these challenges through more advanced perturbation strategies, improved surrogate models, and optimised distance computations would further enhance its robustness, scalability, and applicability.

\section{Future Works}

Building on the limitations identified in this thesis, several promising directions can extend and strengthen the gSMILE framework across generative AI domains:

\begin{enumerate}

    \item \textbf{Developing more expressive and semantically aware perturbation strategies:}
    Since the current approach relies on binary removal, future work can explore richer perturbations, such as synonym substitution, paraphrasing, context-preserving edits, or adding Gaussian noise in the embedding space. These strategies can reduce semantic distortion and yield more informative attribution patterns.

    \item \textbf{Improving robustness to significant output shifts caused by token-level perturbations:}
    Future research may investigate perturbation schemes that preserve semantic coherence, including context-aware masking or constrained paraphrasing, along with regularisation techniques to stabilise surrogate models in locally non-smooth regions.

    \item \textbf{Mitigating hallucination-induced noise in generative explanations:}
    Integrating gSMILE with hallucination detection, confidence calibration, or causal masking techniques could reduce the influence of unreliable model outputs and improve explanation fidelity for models sensitive to degraded or ambiguous inputs.

    \item \textbf{Enhancing computational efficiency through optimised sampling and faster distance measures:}
    To reduce the computational cost of large perturbation sets and Wasserstein-based weighting, future work may incorporate optimisation-based sampling strategies (e.g., OptiLIME), adaptive perturbation selection, or faster optimal transport variants such as Sinkhorn divergence.

    \item \textbf{Exploring more powerful surrogate models for highly non-linear generative systems:}
    The current surrogate models (weighted linear regression and Bayesian ridge regression) may be insufficient for non-linear behaviours. Future extensions could utilise kernel regressors, neural local approximators, random-feature models, or tree-based surrogates to enhance fidelity in complex generative architectures, such as diffusion models.

    \item \textbf{Improving scalability and accessibility for large or non-API models:}
    Developing lightweight or approximate versions of gSMILE could reduce resource requirements, while the use of standardised interfaces may simplify experiments on locally deployed or open-source multimodal models.

    \item \textbf{Extending the framework to additional generative modalities:}
    Beyond text and image editing, future work may apply gSMILE to video generation, audio synthesis, and broader multimodal architectures. These settings offer opportunities to evaluate interpretability under temporal and cross-modal constraints.

    \item \textbf{Integrating gSMILE with attention-based explainability mechanisms:}
    Combining perturbation-based explanations with attention visualisation methods, such as DAAM, could produce joint token-attention-pixel explanations, providing a richer understanding of how textual inputs and internal attention patterns shape generated outputs.

\end{enumerate}



\begin{spacing}{0.9}


\bibliographystyle{unsrt} 
\cleardoublepage
\bibliography{reference} 



\end{spacing}


\begin{appendices} 



\end{appendices}

\printthesisindex 

\end{document}